\documentclass[11pt]{article}
\pdfoutput=1
\usepackage[margin=0.95in]{geometry} 
\usepackage{listings}
\usepackage{mathrsfs} 
\usepackage{kpfonts,comment}   
\usepackage{graphicx}
\usepackage{dialogue}
\usepackage{verbatim}
\usepackage{tabularx}
\usepackage[hybrid]{markdown}
\usepackage[table]{xcolor} %
\usepackage[most]{tcolorbox}
\usepackage{soul} %

\usepackage{lipsum}

\usepackage{booktabs}
\usepackage{makecell}

\usepackage{cpdef}
\usepackage{float}
\usepackage{mathrsfs}
\usepackage{epigraph}
\usepackage{fancyvrb}
\usepackage{authblk}
\usepackage{placeins}
\usepackage{booktabs}
\usepackage{enumitem}
\makeatletter
\def\thickhline{%
  \noalign{\ifnum0=`}\fi\hrule height 1.2pt
  \futurelet\@tempa\@xhline}
\makeatother
\RecustomVerbatimCommand{\VerbatimInput}{VerbatimInput}
{fontsize=\footnotesize,
 frame=single,  
 framesep=0.5em,
 labelposition=topline,
}

\newcommand{\mybluehyperlink}[2]{\hypersetup{linkcolor=blue}\hyperlink{#1}{#2}\hypersetup{linkcolor=red}}

 \newboolean{finalver}
\setboolean{finalver}{true} %
\newboolean{iclr}
\setboolean{iclr}{false}
\newboolean{arxiv}
\setboolean{arxiv}{true}
\ifthenelse{\boolean{finalver}}{
      \newcommand{\cp}[1]{\textcolor{purple}{\bf\small [#1 --CP]}}
    \newcommand{\kz}[1]{\textcolor{blue}{\bf\small [#1 --KZ]}}

\newcommand{\safevspace}[1]{\vspace{0mm}}
\newcommand{\zy}[1]{\textcolor{green}{\bf\small [#1 --ZY]}}

}{
  \newcommand{\cp}[1]{\textcolor{purple}{\bf\small}}
      \newcommand{\kz}[1]{}
\newcommand{\zy}[1]{\textcolor{green}{\bf\small }}

\newcommand{\safevspace}[1]{\vspace{0mm}}

}

\newcommand{\ours}[0]{\textbf{\textsc{Iterative Regret-Minimization Fine-Tuning}}}

\newcommand{\briefalgorithmexpand}[0]{\textbf{\textsc{Iterative RMFT}}}

\definecolor{armygreen}{rgb}{0.29, 0.33, 0.13}
\definecolor{cadmiumgreen}{rgb}{0.0, 0.42, 0.24}
\definecolor{darkgreen}{RGB}{0,100,0}
 
\title{
Post-Training LLMs as Better Decision-Making  Agents: \\A Regret-Minimization Approach}
 \author[1]{Chanwoo Park}
 \author[2]{~~~Ziyang Chen}
 \author[1]{~~~Asuman Ozdaglar}
 \author[2]{~~~Kaiqing Zhang}
 \affil[1]{Massachusetts Institute of Technology}
 \affil[2]{University of Maryland, College Park}
 
\date{November 7, 2025 (Updated: May 29, 2026)}

\begin{document}

\maketitle

\begin{abstract}
Large language models (LLMs) are increasingly deployed as ``agents'' for decision-making (DM) in interactive and dynamic environments. 
Yet, since they were not originally designed for DM, recent studies have shown that LLMs can struggle even in basic online DM problems, failing to exhibit desirable behaviors such as {achieving a} low regret value or {an effective exploration-exploitation tradeoff.} 
To address this issue, we introduce \ours{} (\briefalgorithmexpand{}), a {new} 
post-training 
procedure that repeatedly distills low-regret decision trajectories back into the base model. 
{At each iteration, the model rolls out several decision trajectories for a given online DM  task. \briefalgorithmexpand{} then selects the $k$-lowest regret trajectories, and  trains the model on those trajectories via supervised fine-tuning.} 
Unlike prior approaches that either (a) \emph{distill} action sequences from known online DM algorithms, and/or (b) rely on manually crafted chain-of-thought outputs structured around known and fixed algorithms, our approach  leverages the regret metric to automatically elicit the model's DM ability, and integrates 
the model's self-generated reasoning rationales. 
This reliance on model-generated reasoning avoids rigid output format engineering, and provides more flexible training signals in the natural language format.
Empirical results show that \briefalgorithmexpand{} can improve LLMs' DM performance across a spectrum of models—from Transformers with numerical input/output, to lightweight open-weight LLMs, and to the more advanced closed-weight LLM, GPT-4o mini.
Notably, the flexibility that \briefalgorithmexpand{} offers regarding output and reasoning formats allows the trained models to naturally exhibit generalization performance across tasks varying in time horizon, action space size, reward generation processes, and DM contexts/scenarios that are described by natural language.  
Finally, we provide a theoretical insight into how a single-layer {attention Transformer model} 
may lead to a no-regret learner under our training paradigm, in a simplified setting. 
Overall, we position our approach as an initial exploration, calling for more principled and novel post-training paradigms for LLMs when it comes to addressing DM tasks.
\end{abstract}

\section{Introduction}

Artificial Intelligence (AI) agents are increasingly deployed in many sectors to assist humans in decision-making (DM) tasks. These tasks necessitate AI agents' reasoning, adapting, and taking sequences of actions that unfold over time in multi-turn interactions.
Many real-world DM tasks are grounded in \emph{natural language}, where both the task descriptions and the input/output of the decision-maker are expressed in language, beyond the classical DM formulations with symbolic and numeric quantities, \emph{e.g.,} in multi-armed bandits (MABs) and online learning \citep{lattimore2020bandit,hazan2016introduction}. This makes LLMs, with their prominent language interface, a natural candidate as a \emph{DM agent} for language-grounded DM tasks \citep{yao2022react, hao2023reasoning, shinn2024reflexion, wang2023describe, SignificantGravitasAutoGPT,ahn2022can, wang2023voyager,li2024stride}.

However, since LLMs are not trained as decision-makers, it remains unclear why they would excel in DM. 
In fact, recent studies have shown that LLMs, when used \emph{off-the-shelf} (\textit{i.e.}, without fine-tuning or strong {inference-time intervention}), may perform poorly even in simple and canonical online DM tasks, exhibited through their inadequate exploration and exploitation behavior. Specifically, they have been observed to fail to robustly explore without heavy scaffolding \citep{krishnamurthy2024can}, suffer from  \emph{linear} growth of regret over  time 
{in non-stochastic  environments} \citep{park2024llm},  
fail to exhibit consistent exploitation {behaviors} \citep{xia2024beyond}, and fail to explore in non-stationary environments \citep{zhang2025comparing}. Hence, it is important to systematically enhance the DM ability of LLMs within their design, before blindly deploying them as DM agents at scale. 

To this end, a few recent studies have proposed \emph{training} methods to enhance the DM abilities of LLMs. \citet{nie2024evolve,schmied2025llms} proposed training approaches based on \emph{algorithm distillation} \citep{laskin2022context} and \emph{reinforcement learning}, respectively. However, LLMs trained by these approaches have limited generalization abilities to language-grounded DM tasks with rich scenarios, real-world contexts, and variable problem structures. Furthermore, it remains unclear how to generalize the approaches when the underlying DM environments do not follow the canonical ones (\emph{e.g.,} MABs), and/or when the (optimal) expert algorithms (to be {distilled from}) for the underlying DM tasks are \emph{unknown/unclear}, as is common in real-world applications. More importantly, most of these methods did not leverage nor enhance the rich \emph{reasoning}  process of LLMs, but just predict/improve their \emph{final (action) output}. A more detailed discussion on these most related contemporaneous studies can be found in \Cref{sec:related_work_detailed}.

{
In this work, we provide a different  framework for post-training LLMs for language-grounded DM, based on the common metric of \emph{regret} in
online DM \citep{shalev2012online,hazan2016introduction,lattimore2020bandit}. Regret measures how much worse an agent's sequence of decisions is compared to the best possible decision (sequence) in hindsight. 
Regret provides a quantitative framework for evaluating an agent’s performance in interactive, online environments. Minimizing regret typically requires a careful balance between exploration and exploitation, as well as robustness to arbitrary, or sometimes even adversarial, environments. 
Moreover, regret also  offers a way of  modeling and analyzing strategic behaviors in \emph{multi-agent interactive}  environments, with experimental evidence \citep{erev1998predicting,nekipelov2015econometrics}: the long-run interaction of no-regret learners leads to certain
equilibrium when they repeatedly play games \citep{cesa2006prediction}.  Indeed, regret has served as a \emph{unified} metric to be optimized across various (numerical) online DM environments. It is thus natural to ask: \emph{Can we leverage {regret-minimization as a principle} to post-train LLMs as better decision-makers?} Along this line, for Transformer models with numeric input/output,  \citet{park2024llm} recently developed a novel training loss, regret-loss, to promote the  no-regret behaviors of the model, with promising experimental results {for Transformers with numeric input/output, by directly minimizing the regret value through  backpropagation.} 
{However, extending such regret-based training from numeric to language-grounded decision-making is nontrivial, since while regret can be computed once decisions are extracted, it remains challenging to 
align this feedback with the token-level, autoregressive  sampling-based generation process of language input/output.}

Specifically, we develop a self-improving post-training approach for eliciting LLMs' online DM abilities, \ours{} (\briefalgorithmexpand{}), by iteratively  performing  supervised fine-tuning (SFT) on self-generated {decision-making and reasoning} trajectories with \emph{low regret}. \briefalgorithmexpand{}
is flexible enough to apply across various DM environments, thanks to the 
{universality of regret as a metric for online DM.} 
{Because \briefalgorithmexpand{} naturally operates in the language space—{it samples scenarios expressed in natural language, selects trajectories based on enhanced reasoning, ranks them by regret, and directly fine-tunes on those trajectories}—the approach does not need to be translated to the \emph{pre-specified} {numeric DM {environment}}, \emph{e.g.,} the action space size, reward generation processes, and time horizon.} More importantly, the iterative SFT process naturally leverages and further enhances the chain-of-thought (CoT) reasoning rationales  generated by LLMs, for improved online DM performance. {Our  SFT-based paradigm is also compatible with {all} existing post-training interfaces, even including the training API for proprietary \emph{closed-weight} models such as the GPT-series.} 
We evaluate the effectiveness of \briefalgorithmexpand{} on the canonical DM environments of \emph{full-information online learning (FOL)}, \emph{multi-armed bandits}, and \emph{non-stationary multi-armed
bandits (NS-MABs)}, when they are grounded in language.  
Beyond lowering regret values, \briefalgorithmexpand{} also  automatically elicits enhanced online DM behavior from LLMs (\emph{e.g.,} improved  \textit{exploration-exploitation (E-E) tradeoff}), {despite not being explicitly guided by or distilled from existing algorithmic heuristics from online DM.} 
{Without relying on pre-defined expert algorithms, \briefalgorithmexpand{} may be viewed as a way to empower LLMs to \emph{autonomously discover algorithms} for DM, as incentivized by the regret metric. 
In \citet{park2024llm}, the regret objective per se was directly optimized, yielding a known no-regret algorithm (Follow-the-Regularized-Leader (FTRL)) analytically. 
In contrast, our current framework trains the model to imitate its own best-performing decision trajectory as evaluated by the regret metric. In a simplified single-layer attention setting, this imitation-based process again converges to FTRL, suggesting that no-regret behaviors may naturally emerge {by this iterative self-imitation process}.}}

\paragraph{Roadmap.} 
The rest of the paper is organized as follows. {We first provide the meta algorithm \briefalgorithmexpand{} in \Cref{section:meta} and instantiate the meta algorithm for different tasks in the following sections.} In \Cref{sec:training-transformer}, we begin by training Transformers with numeric input/output, as a warm-up to better understand the feasibility/trainability of our \briefalgorithmexpand{} procedure, and provide {a theoretical analysis of our approach}.
In 
\Cref{sec:open-weight-training}, we present {the results of} 
applying \briefalgorithmexpand{} to open-weight but relatively weak LLMs (including Phi-3.5-mini-instruct \citep{abdin2024phi}, Gemma-2-9b-it \citep{team2024gemma}, and Qwen3-8B \citep{yang2025qwen3}). In \Cref{sec:training-gpt}, we demonstrate the successful training of the closed-weight LLM of GPT-4o mini \citep{openai2023gpt4},  
across a variety of language-grounded online DM tasks.

\subsection{Related Work}\label{sec:related_work_detailed}

\paragraph{LLM-Agents for Real-World Decision-Making.} The reasoning capabilities of LLMs have seen significant improvement due to the recent progress in pre-training \citep{openai2023gpt4,bubeck2023sparks}, post-training  \citep{guo2025deepseek,park2025maporl}, and 
prompting methods \citep{wei2022chain,yao2023tree}. Leveraging these substantial developments, there has been an emerging trend to position LLMs as the \emph{central controller} of  
autonomous agents for \emph{decision-making} 
\citep{yao2022react,shinn2024reflexion,sumers2024cognitive}. One major benefit of utilizing LLMs as controllers is their capacity to parse  arbitrary language inputs and generate natural-language-based outputs, allowing for a highly 
flexible and 
adaptable interaction interface for real-world applications. As an example, a notable strand of research investigated LLM agents through the lens of planning \citep{hao2023reasoning,valmeekam2023planbench,huang2022inner,shen2023hugginggpt}, an approach that has also been adopted in embodied-AI and robotic systems \citep{ahn2022can,wang2023describe,SignificantGravitasAutoGPT,wang2023voyager}. In parallel, another line of work concentrated on developing LLM agents tailored to specific domains, such as software engineering \citep{yang2024sweagent,wang2025openhands}, enterprise operations \citep{2024workarena,boisvert2024workarena}, healthcare \citep{kim2024adaptive}, and cybersecurity \citep{zhang2025cybench}.

\paragraph{Understanding LLM-Agents in Canonical Settings.}
The aforementioned progress on LLM-agents for decision-making primarily focused on tailoring LLMs to solve {complex but specific} real-world tasks. However, it remains unclear \emph{if} and \emph{why} LLMs are good decision-makers, especially given that they were not designed for DM. The high complexity in real-world tasks makes it challenging to diagnose these questions, inspiring a series of works on understanding  LLMs in basic, canonical DM settings.  \citet{wusmartplay,krishnamurthy2024can,park2024llm} first carefully studied the in-context online DM ability of  LLMs, with \citet{wusmartplay,krishnamurthy2024can} focusing on the exploration ability in stochastic MABs, and \citet{park2024llm} focusing on the online adversarial and game-theoretic settings. 
Subsequently, \citet{nie2024evolve} expanded the bandit settings considered in \citet{krishnamurthy2024can}, deepening the understanding of LLMs' (in)ability for efficient exploration. Further, \citet{zhang2025comparing} examined both the MAB and the NS-MAB environments, focusing on the comparison with human behaviors; \citet{xia2024beyond} studied the dueling bandit setting when explicit numerical rewards are not observable. 
Other works that also examined LLMs in canonical bandit environments are \citet{felicioni2024importance,rahn2024controlling,monea2024llms}. In particular, \citet{nie2024evolve,rahn2024controlling,monea2024llms, sun2025large} also developed \emph{training-free} interventions to enhance the DM ability of LLMs.

{
\paragraph{Post-Training  
LLMs for Decision-Making.} 
In light of the observations above, 
several studies have also proposed \emph{(post-)training}  methods to explicitly enhance 
LLMs' DM abilities. 
\citet{nie2024evolve,schmied2025llms} represent the recent efforts in post-training/fine-tuning LLMs as better decision-makers, aligning  with the motivation of the present paper. Specifically,  
\citet{nie2024evolve} developed a fine-tuning approach based on \emph{algorithm distillation} \citep{laskin2022context}, which requires \emph{pre-defined}  input/output formats, \emph{e.g.,} the action space size/reward vector   dimension and time horizon,  and thus has  limited generalizability to 
DM tasks with rich scenarios, real-world contexts, and variable problem structures. Moreover, it may not be applicable to scenarios when the optimal/expert algorithms are {unknown/unclear}, as is common in real-world applications. 
Finally, the approach did not incorporate nor enhance the \emph{CoT reasoning rationales} of the LLMs in DM. 
Very recently, 
\citet{schmied2025llms} introduced {reinforcement learning fine-tuning} (RLFT) on self-generated CoT paths and further compared it against {algorithm distillation}. RLFT is closely related to \briefalgorithmexpand{} in that both can incorporate self-generated reasoning rationales in the training data. 
In comparison, \briefalgorithmexpand{} uses the {cumulative regret} of the entire trajectory as the training signal, while RLFT used the {rewards} per se. As (cumulative) reward  maximization is not identical to regret minimization in general, it is unclear if such an approach applies to 
the online learning (with adversaries) or non-stationary bandit settings  we consider.   The tasks focused on are also different: RLFT addressed MABs with two specific DM scenarios from \citet{nie2024evolve}, 
also with extensions to contextual bandits (CBs) and tic-tac-toe; \briefalgorithmexpand{} 
focused on tasks with the MAB environment with more generic DM scenarios, as well as those with online learning and non-stationary bandit environments. 
Thanks to the flexibility and unification offered by the regret metric, 
\briefalgorithmexpand{} trained LLMs also exhibit generalizability in  various problem structures. 
Finally, \citet{schmied2025llms} also demonstrated the promise of SFT-based training on CoT rationales (as \briefalgorithmexpand{} does), although the rationales were manually designed to mimic the UCB algorithm \citep{auer2002finite}, rather than being organically and automatically generated by the LLMs.}

\paragraph{Post-Training LLMs for \emph{Decision-Making} vs. for \emph{Reward Maximization}.}
Our focus in this paper 
shall not be confused with the recent line of \emph{RL-based post-training} paradigms for LLMs, for tasks such as  
mathematical reasoning \citep{jaech2024openai,guo2025deepseek,shao2024deepseekmath} and alignment \citep{ouyang2022training,rafailov2023direct}. In those paradigms, LLMs are trained to \emph{maximize  certain reward functions} (under regularization) associated with the specific task, which reflect the correctness and/or human-preference of the responses, using RL methods. Notably, the reward maximization  objective per se cannot elicit \emph{efficient exploration} automatically, which is precisely the inspiration of many variants that explicitly incorporated exploration mechanisms in such RL-based \emph{training process}, see \emph{e.g.,} \citet{xieexploratory,du2024exploration,dwaracherla2024efficient}. In contrast, our focus is on post-training LLMs \emph{for decision-making} as an \emph{ability}, so that they may be able to address novel rewards/tasks at \emph{inference time} efficiently. To this end, the post-training should provide signals to elicit LLMs to \emph{learn to explore} at inference time, ideally with \emph{generalizability} across task specifications like the action (sets), rewards, horizons, etc. Our approach resorts to the regret metric as a possible source of such signals.

\section{Preliminaries}\label{sec:preliminary}
\subsection{Decision-Making Environments}
\label{ssec:sdm}

We focus on the following canonical online DM environments throughout the paper: \emph{full-information online learning}, \emph{multi-armed bandits}, and \emph{non-stationary multi-armed bandits}. 
\ifthenelse{\boolean{iclr}}
{Due to the standardness of these DM  environments,  
we defer a detailed introduction of them to \Cref{appendix:ssec_sdmenv}, and summarize the specifications of these environments in \Cref{tab:sequential_comparison}, together with the notation to be used throughout the paper.
}{}

\ifthenelse{\boolean{arxiv}}
{
\paragraph{Full-Information Online Learning.} 
In the FOL environment, an agent interacts with the  environment over $T$ rounds by sequentially making decisions based on the feedback received from prior rounds. At each round $t \in [T]$, the agent selects a decision policy $\pi_t \in \Pi$, where $\Pi$ denotes a bounded decision {policy} space. After the agent commits to $\pi_t$, the environment reveals a bounded reward function $f_t: \Pi \to [-B, B]$ for some constant $B > 0$, possibly chosen in an adversarial manner. The agent then {receives} {some feedback about $f_t(\pi_t)$ (\textit{e.g.}, $\nabla f_t(\pi_t)$), and updates the policy to $\pi_{t+1}$ based on the observed feedback so far \citep{cover1966behavior, vovk1990aggregating, littlestone1994weighted, hazan2016introduction}.}
A common instantiation of this framework is when $\Pi = \Delta(\cA)$, the probability simplex over a finite action set $\cA$, and the reward function is linear in the policy: $f_t(\pi_t) = \langle R_t, \pi_t \rangle$, where $R_t \in \mathbb{R}^d$ is a reward vector. This case is also referred to as the \emph{Experts Problem}  \citep{littlestone1994weighted}, which may also be used to model  agents' learning in repeated games \citep{cesa2006prediction}.   Alternatively, $\Pi$ may be an $\ell_2$-Euclidean-ball $B(\pmb{0}_d, R_\Pi, \norm{\cdot}_2)$ { for some $R_\Pi>0$}, and the reward takes the same linear form. In either case, we assume the agent receives the full reward vector $R_t$ at each round $t$. 

\paragraph{Multi-Armed Bandits.}
In the MAB environment, an agent interacts with the environment over $T$ rounds, sequentially selecting actions and updating its strategy based on partial feedback \citep{lattimore2020bandit}. The agent has access to an \emph{action set} $\mathcal{A}$ with $d:= |\cA|$, where each element corresponds to a distinct ``arm.'' At each round $t \in [T]$, a reward vector $R_t \in \RR^{d}$ {is sampled from some underlying reward generation distribution with mean $r \in [-B, B]^d$.} The agent chooses an action $a_t \in \mathcal{A}$, and upon committing to this action, the environment reveals only the {sampled} reward associated with that arm, $R_t(a_t)$. Thus, unlike the FOL environment, the agent receives only \emph{bandit feedback}—a single scalar reward $R_t(a_t)$—rather than the entire reward vector $R_t$.

\paragraph{Non-Stationary Multi-Armed Bandits.} 
In the NS-MAB environment, an agent operates with the same action set $\cA$, but the mean reward of each arm is allowed to evolve over time. Formally, at each round $t \in [T]$, the environment generates a reward vector $R_t$ with mean $r_t \in [-B, B]^d$, where the sequence $(r_t)_{t \in [T]}$ may drift. A common assumption is that this drift is controlled by a variation budget $V_T := \sum_{t=2}^T \norm{r_t - r_{t-1}}_\infty$, which captures the total amount of non-stationarity \citep{besbes2014stochastic}. The agent still chooses an action $a_t \in \cA$ and only observes the realized reward $R_t(a_t)$. The objective is therefore to adapt to the changing reward landscape while learning from bandit feedback.

{Both MABs and NS-MABs belong to, and will be referred to as, online DM problems with \emph{bandit/partial} feedback, in contrast to the \emph{full-information} feedback in the FOL environment.} 
 
}{}

\paragraph{Language-Grounded Decision-Making.} {These classical online DM environments are specified in a \emph{non-linguistic} form: the actions $a\in\cA$ are given as symbolic or numeric values, and the reward functions $r$ take numeric values. More importantly, both the \emph{input} and \emph{output} of the decision-making agent are \emph{numeric}. For example, for tasks with the MAB environment, the agent takes the history of numeric rewards received $\{R_\tau(a_\tau)\}_{\tau\leq t-1}$ and actions taken $\{a_\tau\}_{\tau\leq t-1}$ as input, and outputs an action $a_{t}$ (or a probability distribution over the action space $\pi_{t}\in\Delta(\cA)$). In contrast, thanks to their language ability, LLM agents may address online DM tasks beyond these classical, purely numeric ones, where the task description, input, and output can all be language-grounded. We refer to each \emph{language-based  description} of a task as a \textit{scenario} (\emph{e.g.,} from healthcare, resource allocation, and marketing applications, see \Cref{ssec:setting-sec5,ssec:setting-sec6} and \Cref{appendix:prompt-open-weight,appendix:prompt_scene_generation} for more examples). 
{The LLM agent is then asked to make a decision, based on the  language-based  description of the interaction history so far in a \emph{dialogue form},}  
receives feedback from the environment, and repeats. The decision output from LLM agent is also in the natural language format, from which a policy $\pi_t\in\Delta(\cA)$ {or an action $a_t\in\cA$} is extracted. 
Additionally, in contrast to the numeric online DM algorithms that \emph{directly} map the numeric inputs to the output policies, LLM agents may   also \emph{reason}, in the language format, about the decision before outputting the policies. We refer to such an interaction protocol between the LLM agent and  the environment as \emph{language-grounded DM}, and our \emph{dialogue-form} interaction protocols 
resemble how LLMs are used as agents in real-world applications, in which the reasoning rationales of the agents can also be naturally integrated.}

\subsection{Performance Metric: Regret}
\label{ssec:regret}

In online DM environments, a fundamental measure of performance is the \emph{regret}, a widely used metric in online learning \citep{hazan2016introduction, shalev2012online}, reinforcement learning \citep{szepesvari2022algorithms}, and learning in games \citep{cesa2006prediction}. Given an algorithm $\mathscr{A}$ designed for such environments, the regret metric quantifies how much worse the agent's decisions are, {during the course of learning,} 
compared to those of the best possible policy. We denote this metric by  $\text{Regret}_{\mathscr{A}}$ (or $\text{Regret}_{(\pi_{\mathscr{A}, t})_{t \in [T]}}$), \ifthenelse{\boolean{iclr}}{
and formally define regret for the MAB environment as below. We defer the standard regret definitions for other online DM environments to \Cref{appendix:ssec_regret} due to space constraints. 

\paragraph{Multi-Armed Bandits.}  
In the MAB environment, the \emph{expected regret} of algorithm $\mathscr{A}$, which generates an action sequence $(a_{\mathscr{A},t})_{t=1}^T$ of length $T$, is defined as
\begin{equation*}
    \text{Regret}_{\mathscr{A}}\left(r, T \right) := \mathbb{E} \left[ T\cdot \max_{a \in \cA} r(a)  - \sum_{t=1}^T r(a_{\mathscr{A},t}) \right]
\end{equation*}
\normalsize 
where the expectation is taken over the randomness of the algorithm. The regret notion  quantifies the difference between the cumulative expected reward that would have been obtained by selecting the best arm, and the expected reward accumulated by $\mathscr{A}$ through its chosen actions over $T$ rounds. 

}{ 
and formally define it for each online DM environment below.

\paragraph{Full-Information Online Learning.} In the FOL environment with reward vectors $(R_t)_{t \in [T]}$ and decision space $\Pi$, the regret of a DM algorithm $\mathscr{A}$—which generates a sequence of policies $\pi_{\mathscr{A}, t} \in \Pi$—is defined as
\begin{equation*}
\text{Regret}_{\mathscr{A}}\left((R_t)_{t \in [T]}, T\right) := \max_{\pi \in \Pi} \sum_{t=1}^T \langle \pi, R_t \rangle - \sum_{t=1}^T \langle \pi_{\mathscr{A},t}, R_t \rangle.
\end{equation*}
This expression compares the cumulative reward of the best fixed decision in hindsight against that of the algorithm's chosen decisions over time.

\paragraph{Multi-Armed Bandits.}  
In the MAB environment, the \emph{expected regret} of algorithm $\mathscr{A}$, which generates an action sequence $(a_{\mathscr{A},t})_{t=1}^T$ of length $T$, is defined as
\begin{equation*}
    \text{Regret}_{\mathscr{A}}\left(r, T \right) := \mathbb{E} \left[ T\cdot \max_{a \in \cA} r(a)  - \sum_{t=1}^T r(a_{\mathscr{A},t}) \right]
\end{equation*}
\normalsize 
where the expectation is taken over the randomness of the algorithm. The regret notion  quantifies the difference between the cumulative expected reward that would have been obtained by selecting the best arm, and the expected reward accumulated by $\mathscr{A}$ through its chosen actions over $T$ rounds.

\paragraph{Non-Stationary Multi-Armed Bandits.}  
In the NS-MAB environment, the common metric is \emph{dynamic regret} \citep{besbes2014stochastic}, which,   
for an algorithm $\mathscr{A}$ that generates an action sequence $(a_{\mathscr{A},t})_{t=1}^T$, is defined as 
\begin{equation*}
    \text{D-Regret}_{\mathscr{A}}\left((r_t)_{t \in [T]}, T, V_T \right) := \mathbb{E} \left[ \sum_{t=1}^T \max_{a \in \cA} r_t(a) - \sum_{t=1}^T r_t(a_{\mathscr{A},t}) \right],
\end{equation*}
where the expectation is taken over the randomness of the algorithm and the observed rewards. This notion compares~$\mathscr{A}$ against the sequence of best arms of each round.

{For convenience, we may write the regret above in shorthand notation as $
\text{(D-)Regret}\left(T\right)$, when the specifications of $(R_t)_{t\in[T]},r,(r_t)_{t\in[T]},\mathscr{A}$ are clear from the context.} 
{An algorithm $\mathscr{A}$ is referred to as being \emph{no-(dynamic-)regret} if the (dynamic-)regret notion defined above 
grows \emph{sublinearly}  with respect to the time horizon $T$, \textit{i.e.}, 
$
\text{(D-)Regret}\left(\cdot\right) = o(T)$.
This implies that, on average, the algorithm performs comparably to the appropriate benchmark in hindsight as $T$ increases.} Baseline algorithms for these canonical online DM environments will be introduced in \Cref{appendix_baseline}. 

}

\paragraph{Regression-Based Validation of No-Regret Behaviors.} 
We may validate the sublinearity of regret by regression, which was proposed in  \citet[Section 3]{park2024llm}. Specifically, we perform a linear regression on the data $\left\{ \left(t, \log \text{(D-)Regret}({t}) \right) \right\}_{t \in [T]}$ by fitting a function $g(t) = \beta \log t + \alpha$. The estimated coefficient $\hat{\beta} < 1$ may be used to indicate the sublinear growth of the regret over time. We denote the $p$-value of the coefficient estimate $\hat{\beta}$ as $p_{\text{reg}}$, and report the pair $(\hat{\beta}, p_{\text{reg}})$ as an indicator for assessing the sublinear regret growth, where smaller values of $\hat{\beta}$ accompanied by low $p_{\text{reg}}$ provide stronger evidence of the no-regret behavior.

\paragraph{Metrics for Exploration Efficiency.} 
To be qualified as a good decision-maker, the agent needs to balance \emph{exploitation}  with \emph{exploration} in learning, especially in environments with only \emph{bandit} feedback \citep{lattimore2020bandit}. To evaluate the \emph{exploration efficiency} of LLM agents {in MABs}, other than regret, we also adopt the metrics proposed by \citet{krishnamurthy2024can}: \texttt{SuffFailFreq$(t)$} and \texttt{MinFrac$(t)$}. Specifically,  \texttt{SuffFailFreq}$(t)$ measures the proportion of runs in which the best action (\textit{i.e.}, the action with the highest expected reward) is never selected from round $t$ to $T$. On the other hand, \texttt{MinFrac}$(t)$ captures how uniformly the available actions are explored. We provide more details of these metrics in \Cref{ssec:metricee}. 
For non-stationary MABs, we only report the results of \texttt{SuffFailFreq}$(t)$ as \texttt{MinFrac}$(t)$ does not provide a meaningful insight in this case (see \Cref{ssec:metricee} for more details). 

{In our experiments later, other than these metrics, we will also compare the performance of our trained models with that of \emph{known no-regret}  learning algorithms in online DM as baselines.}

\section{Meta Algorithm: \ours{}}
\label{section:meta}

Our goal is to develop a post-training paradigm for the decision-making ability of LLMs that are compatible {with the interaction protocol with language description of the tasks, as well as language-grounded input/output.}  
To this end, we propose to fine-tune LLMs based on \emph{self-generated} language data. To promote the decision-making capability, we then propose to leverage the regret notions introduced in \Cref{ssec:regret} as the criterion to select the self-generated trajectory data for training, which will 
then be used to fine-tune the model via supervised fine-tuning,  reinforcing the low-regret behavior.  
Thanks to the universal applicability of the regret metric in online DM, this process can be viewed as a meta-algorithm generally applicable to environments considered in \Cref{tab:sequential_comparison}.
In the remainder of this paper, we focus on the instantiations of the meta-algorithm for the FOL, MAB, and NS-MAB environments. We term such a meta-algorithm as \ours{} (\briefalgorithmexpand{}), as tabulated in \Cref{alg:ssft-meta-algorithm}.
\ifthenelse{\boolean{arxiv}}{\begin{algorithm}[!h]
\caption{Meta Algorithm: \ours{}}\label{alg:ssft-meta-algorithm}
\begin{algorithmic}[1]
\State \textbf{Input:} A DM environment  (\textit{e.g.}, FOL, MAB, NS-MAB); %
an initial model ($\mathscr{A}$) for DM
\For{iteration $= 0, 1, 2, \dots$}
    \State $\cD = \emptyset$
    \For{scenario index $i = 1, 2, \dots, M$}
        \State Sample $L$ trajectories $C_1, \dots, C_L$ from the model under scenario$_i$ \label{line:L-traj-meta-alg}
        \State Compute the regret (\Cref{ssec:regret}) of each trajectory and select the $k$ trajectories with the lowest regret: $C_{(1), i}, \dots, C_{(k), i}$ 
        \State Update the training dataset: $\cD = \cD \cup \{\{\text{scenario}_i, C_{(1), i}, \dots, C_{(k),i}\}\}$
    \EndFor
    \State Fine-tune the model on the dataset $\cD$ via supervised fine-tuning \label{line:sft}
\EndFor
\end{algorithmic}
\end{algorithm}
}{}

\noindent

Specifically,   
in each iteration of \briefalgorithmexpand{}, we iterate over $M$ different \emph{scenarios}. Here, a \emph{scenario} refers to a specifically generated instance of a DM environment {grounded in language}. 
Then, for each scenario, $L$ trajectories are sampled from the current model $\mathscr{A}$,  which is intentionally denoted by the same notation as an   \emph{algorithm} in \Cref{sec:preliminary}, as the LLM agent now is intended to \emph{act  as a decision-making algorithm.}  
 These $L$ trajectories are then  evaluated based on the regret metric (cf. \Cref{ssec:regret}). Note that the regret can be computed \emph{during training}, since the environments (\emph{e.g.,} rewards and transitions) are generated by and thus known to the trainer, and the $\max$ operator in the regret definition can thus also be evaluated. Importantly, we highlight that, such privileged information is not required \emph{during inference time}, and does not contradict our overall goal: the post-training procedure is not to teach LLM agents to solve DM tasks tied with a \emph{particular}  reward (seen during training), but instead to elicit their \emph{DM capability} when facing novel DM tasks at the {inference} time.  See \Cref{sec:remark_optimal_label} in the appendix for a more detailed discussion. 
 
Taking the MAB environment as an example, at each round $t$, we query a policy $\pi_{\mathscr{A},t}\in\Delta(\cA)$ from the model $\cA$, with only bandit feedback $\{R_\tau(a_\tau)\}_{\tau=1}^{t-1}$, where $R_\tau$ is the sampled reward vector at round $\tau$ (with the mean vector being $r$). The expected regret is then evaluated as $T\cdot \max_{a}r(a)-\sum_{t=1}^T \mathbb{E}_{a \sim \pi_{\mathscr{A},t}}[r(a)]$. Note that in the experiments, one may also use $\sum_{t=1}^T \mathbb{E}_{a \sim \pi_{\mathscr{A},t}}[R_t(a)]$ as a surrogate for the second term in the regret. We analogously compute regret for the FOL and NS-MAB environments discussed in this paper (\Cref{regret-instantitaion}). The top-$k$ trajectories with the \emph{lowest regret} are then selected and added to the dataset $\cD$. At the end of each iteration, the model is fine-tuned using the trajectories in $\cD$ via supervised fine-tuning. This self-improving loop allows the model to iteratively refine its decision-making capability by learning from its own best-performing behaviors so far, in terms of the regret metric.

{In the next sections, we will instantiate the meta-algorithm, \Cref{alg:ssft-meta-algorithm}, in several online DM tasks with different environments and input-output modalities. Specifically, {as summarized in \Cref{tab:modality},  as a sanity check on its trainability,} 
in \Cref{sec:training-transformer}, we first train \emph{Transformers with numerical input{/output}}  for \emph{numerical DM}; in \Cref{sec:open-weight-training}, we fine-tune \emph{open-weight LLMs} for \emph{language-grounded numerical DM} 
tasks; in \Cref{sec:training-gpt}, we fine-tune the \emph{closed-weight LLM,} GPT-4o mini for \emph{language-grounded DM with real-world contexts} tasks}. More details on the definitions of these tasks are provided in each section separately.  
\ifthenelse{\boolean{arxiv}}{

{
\begin{table}[ht]
\centering
\footnotesize
\renewcommand{\arraystretch}{1.5}
\begin{tabular}{@{}c c c c c@{}}
\toprule
 & \makecell[c]{\textbf{Transformers} \\ \textbf{with Numerical Input/Output} \\ (\Cref{sec:training-transformer})} & \makecell[c]{\textbf{Open-Weight} \\ \textbf{LLMs} \\ (\Cref{sec:open-weight-training})} & \makecell[c]{\textbf{Closed-Weight} \\ \textbf{LLMs} \\ (\Cref{sec:training-gpt})} \\
\specialrule{1.2pt}{1pt}{1pt}
\makecell[c]{\textbf{DM} \\ \textbf{Environments}} 
& FOL \& MAB 
& FOL \& MAB 
& FOL \& MAB \& NS-MAB \\
\midrule
\makecell[c]{\textbf{Category of} \\ \textbf{DM Tasks}} 
& Numerical DM 
& \makecell[c]{Language-Grounded \\  Numerical DM}
& \makecell[c]{Language-Grounded DM \\ with Real-world Contexts} \\
\midrule
\makecell[c]{\textbf{Output} \\ \textbf{with Reasoning}}
& N/A
& Yes
& Yes \\
\midrule
\textbf{Decision Output}
& $\Delta(\cA)$ or $B(\pmb{0}_d, R_{\Pi}, \norm{\cdot}_2)$
& \makecell[c]{Extract \makecell[c]{$\pi_t \in \Delta(\cA)$ for FOL \\$a_t \in \cA$ for MAB}\\ from the output}
&\makecell[c]{Extract $\pi_t \in \Delta(\cA)$ \\ from the output}\\
\midrule
\makecell[c]{\textbf{Sampling} \\ \textbf{Mechanism}} 
& \makecell[c]{Gaussian Noise \\ Perturbation}
& Stochastic Decoding 
& Stochastic Decoding \\
\bottomrule
\end{tabular}
\caption{Summary of {how our meta-algorithm is instantiated for different 
tasks studied in this paper:} 
Transformers with numerical input/output,  open-weight LLMs, and closed-weight LLMs.} 
\label{tab:modality}
\end{table}
}

}{We provide \Cref{tab:modality} to summarize the instantiations of \Cref{alg:ssft-meta-algorithm} considered in this paper: Transformers with numeric input/output, open/closed-weight LLMs with language input/output.}

\section{Training Transformers 
with \briefalgorithmexpand{}}  
\label{sec:training-transformer}

In this section, we instantiate the \ours{}  framework for training Transformer models to solve numerical online DM tasks. We will focus on the most basic online DM environments--FOL and MAB--throughout this section. 
{Our goal is to 
assess the feasibility of our new post-training paradigm, by focusing on}  Transformers—the architecture behind most LLMs.  
The analyses in this section aim to provide a quantitative and more controlled understanding of the potential of \briefalgorithmexpand{}. 

\subsection{Experimental Setup}
\label{ssec:setting-sec4}

\paragraph{Setup. } 
For the FOL environment, at each round $t$, the input to the Transformer is the history of numerical rewards $(R_1, R_2, \dots, R_{t-1})$, and the output is a policy $\pi_{t}\in\Pi$ for the next {round}, where $\Pi = \Delta(\cA)$ or $\Pi = B(\pmb{0}_d, R_{\Pi}, \norm{\cdot}_2)$. For the MAB environment, at each {round} $t$, the input is the history of the observed  rewards $(\hat{R}_1, \dots, \hat{R}_{t-1})$, where $\hat{R}_\tau\in\RR^{|\cA|}$ with $\hat{R}_\tau(a):= \pmb{1}(a = a_\tau) R_\tau(a)$,  and $a_\tau$ is the action selected at round $\tau$.   The Transformer then outputs a policy $\pi_{t}\in\Pi= \Delta(\cA)$, from which the next action $a_{t}\sim \pi_{t}$ is sampled. \ifthenelse{\boolean{arxiv}}{
This setup differs from the ones described in \Cref{sec:open-weight-training} and \Cref{sec:training-gpt}, where both the input and output use language descriptions. In those settings, the model generates the policy autoregressively through a sequence of language tokens. In contrast, in the numerical setting studied here, the Transformer is applied once per round to directly predict the policy. 
}{}
We provide more details on the reward generation processes used for training and evaluation in this section  
in \Cref{appendix:reward_dist}.  
\ifthenelse{\boolean{iclr}}{In \Cref{appendix:sec:defexplanation_tf}, we provide additional remarks on the input and asks, comparing them with {those in} \Cref{sec:open-weight-training,sec:training-gpt}.}{}

\paragraph{Model.} 
We adopt a single-layer linear attention architecture equipped with an output operator that maps the output into the policy space {$\Pi$}.  This represents one of the simplest forms of Transformer architectures, and has been studied in the literature {for the theoretical understanding of Transformers}   \citep{ahn2023transformers, mahankali2023one, park2024llm}. The model output at round $t$ is as follows:
{
\begin{align}
    g((R_1, \dots, R_{t-1}, \pmb{1}_d); V, K, Q, v_c, k_c, q_c) = \texttt{Operator} \left( \sum_{\tau=1}^{t-1} (V R_\tau + v_c) \left( (K R_\tau + k_c)^\intercal (Q \pmb{1}_d + q_c) \right) \right), \label{eqn:single-linear-transformer}
\end{align}}
\vspace{-10pt}

\noindent which is parameterized by $V, K, Q \in \mathbb{R}^{d \times d}$ and $v_c, k_c, q_c \in \mathbb{R}^d$, corresponding to the \emph{value}, \emph{key}, and \emph{query} matrices and their respective bias vectors, respectively. We denote the full set of parameters as $\theta = (V, K, Q, v_c, k_c, q_c)$. 
\ifthenelse{\boolean{arxiv}}{The attention {scores} are computed over the reward history, and the resulting vector is then passed through an {\texttt{Operator}}.
}{} The choice of \texttt{Operator} depends on the policy space $\Pi$: for the {Experts Problem where $\Pi$ is the probability simplex (\textit{i.e.}, $\Pi = \Delta(\cA)$), \texttt{Operator} corresponds to the  \texttt{Softmax} operator; for the case where $\Pi$ is an $\ell_2$-ball (\textit{i.e.}, $\Pi = B(\pmb{0}_d, R_{\Pi}, \norm{\cdot}_2)$), 
\texttt{Operator} corresponds to the projection operation onto the $\ell_2$-ball.} 
\ifthenelse{\boolean{arxiv}}{
\subsection{Instantiating \briefalgorithmexpand{} for Numerical Decision-Making with Transformers}
\label{ssec:instantiation_for_tf}

\begin{algorithm}[!h]
\caption{\ours{} for Numerical DM with Transformers}\label{alg:ssft-numerical-model}
\begin{algorithmic}[1]
\State \textbf{Input:} A DM environment (\textit{e.g.}, FOL, MAB); Transformer model parameterized by $\theta_0$; number of perturbations $L$
\For{iteration $= 0, 1, 2, \dots$}
    \State $\cD = \emptyset$
    \For{scenario index $i = 1, 2, \dots, M$}
    \State Sample $L$ trajectories $C_{1, i} = (\pi_{1, i, t})_{t \in [T]}, \dots, C_{L, i} = (\pi_{L, i, t})_{t \in [T]}$ {by applying perturbations at each round $t$ to the Transformer's output policy $\pi_{\theta_{\text{iteration}}}(\text{reward history}_{<t}\text{ of scenario}_i)$ by \Cref{eqn:perturbation}}
    \State Compute the regret (as in \Cref{ssec:regret}) of each trajectory under scenario$_i$ and select the $k$ trajectories with the lowest regret: $C_{(1), i}, \dots, C_{(k), i}$
        \State Update the training dataset: $\cD = \cD \cup \{\{\text{scenario}_i, C_{(1), i}, \dots, C_{(k),i}\}\}$
    \EndFor
\State Update model parameters to $\theta_{\text{iteration} + 1}$ by minimizing the loss \Cref{eqn:dist} starting from $\theta = \theta_{\text{iteration}}$
\EndFor
\end{algorithmic}
\end{algorithm}

{We instantiate the meta algorithm of \Cref{alg:ssft-meta-algorithm} for numerical DM tasks (cf. \Cref{ssec:setting-sec4}) as 
\Cref{alg:ssft-numerical-model}. In \Cref{line:L-traj-meta-alg} of \Cref{alg:ssft-meta-algorithm}, 
we need to sample multiple trajectories per scenario. However, in this numerical setting, a single Transformer model $\mathscr{A}$ \emph{deterministically} maps numerical reward histories to policies, without randomness nor sampling. This deterministic nature of the output differs from the sampling process of actual LLMs, to which feeding the same reward history yields different trajectories due to the stochastic nature of token sampling. In particular, the decision-making policy from LLMs is implicitly defined through the autoregressive generation process, where each sampled token affects subsequent ones, and the final output text encodes the policy.}

To fill this gap and generate diverse trajectories from the deterministic Transformer model $\mathscr{A}$, we introduce stochastic perturbations (\Cref{eqn:perturbation}) to its output at each round in \Cref{alg:ssft-numerical-model}. For each $\text{scenario}_i$, we sample $L$ trajectories $C_{1,i}, \dots, C_{L,i}$, where each trajectory $C_{\ell,i} = (\pi_{\ell, i, t})_{t \in [T]}$ is a sequence of \emph{perturbed}  policies. {Specifically, 
at each round $t$, the model outputs the following policy:}
\begin{align}
\pi_{\ell,i,t} = \texttt{Operator}\left( \pi_{\theta}(\text{reward history}_{<t}\text{ of scenario}_i) + \epsilon_{\ell,i,t} \right),   \label{eqn:perturbation} 
\end{align}
where $\epsilon_{\ell,i,t} \sim \mathcal{N}(\pmb{0}_d, \sigma^2 I)$ is a Gaussian noise. This procedure induces stochasticity while ensuring that each resulting policy $\pi_{\ell,i,t}$ is a valid policy.

{After perturbation,} $k$ trajectories with the \emph{lowest regret} are selected and stored in the dataset $\cD$ alongside their corresponding scenarios. The model is then updated by minimizing the following loss {with respect to $\theta$}:  
\begin{equation}
\sum_{\substack{(\text{scenario}_i,\, C_{\ell, i} = (\pi_{\ell, i, t})_{t=1}^T) \in \mathcal{D}}} \sum_{t=1}^T \texttt{dist}\left(\pi_{\theta}(\text{reward history}_{<t}\text{ of scenario}_i),\, \pi_{\ell, i, t} \right), \label{eqn:dist}
\end{equation}
where $\pi_{\theta}(\text{reward history}_{<t}\text{ of scenario}_i)$ denotes the model's output at round $t$ {when parameterized by $\theta$}, conditioned on the reward history up to round $t{-}1$, and  \texttt{dist} can be any suitable divergence metric (\textit{e.g.}, the $\ell_2$ distance, the cross-entropy loss, or the KL divergence).

}{}
\subsection{Experimental Results}
\label{ssec:experiment-results-sec4}
We consider two policy spaces for training the Transformer: (1) the $\ell_2$-ball with  $\Pi = B(\pmb{0}_d, R_{\Pi}, \norm{\cdot}_2)$ with radius $R_\Pi = 1$; (2) the probability simplex with  $\Pi = \Delta(\cA)$ (where actions are sampled by applying the softmax function to the Transformer's output vector). For convenience, we refer to each environment as \textbf{Env I} and \textbf{Env II}, respectively. During training, we adopt the $\ell_2$-distance as a \texttt{dist} function in \Cref{eqn:dist}. The Transformer is trained with $d = 3$ and time horizon $T = 25$, using reward vectors generated from the \mybluehyperlink{gaussianmu}{Gaussian} reward (see \Cref{appendix:reward_dist} for more details). 
Full details on the  training hyperparameters are provided in \Cref{app:exp-setup}. 
{}

\subsubsection{Full-Information Online Learning}\label{sec:FOL_transformer}

Interestingly, as illustrated below, for both \textbf{Env I} and \textbf{Env II} of FOL, the Transformer's output empirically converges to that of the known online learning algorithm, Follow-the-Regularized-Leader (FTRL), with various regularizers (see \Cref{appendix_baseline} for a formal introduction). 
Specifically, to  analyze this phenomenon, we first express the linear attention architecture from \Cref{eqn:single-linear-transformer} as follows:
\begin{align}\label{equ:rewrite_linear_att}
    g ((R_1, \dots, R_{t-1}, \pmb{1}_d);  \mathbf{A}, \mathbf{b}, \mathbf{C}, \mathbf{d})= \texttt{Operator}\left( \sum_{\tau=1}^{t-1} \mathbf{A} R_\tau R_\tau^\intercal \mathbf{b} + \mathbf{C} R_\tau + \mathbf{d} \right),
\end{align}
where 
$\mathbf{A} := V$, $\mathbf{b} := K^\intercal(Q \pmb{1}_d + q_c)$, $\mathbf{C} := k_c^\intercal(Q \pmb{1}_d + q_c)V + v_c \mathbf{b}^\intercal$, and $\mathbf{d} := k_c^\intercal (Q \pmb{1}_d + q_c) v_c$. 
We then track the evolution of the following quantities during training: 
\[
\norm{\mathbf{A}}_F \norm{\mathbf{b}}_2, \, \norm{\mathbf{C} - \text{mean}(\mathbf{C}) \cdot I_{d \times d} }_F, \quad \norm{\mathbf{d}}_2 ~\text{(for \textbf{Env I}) or } \norm{\mathbf{d}- \text{mean}(\mathbf{d}) \cdot \pmb{1}_d}_2 ~\text{(for \textbf{Env II})},
\]
if these values converge to 0, then the architecture converges to known {online learning} algorithms. Specifically (using \Cref{eqn:ftrl-def}):
\begin{itemize}
    \item When the \texttt{Operator} is \textit{projection onto the $\ell_2$-ball}, it recovers \textbf{FTRL with $\ell_2$-regularization},
    \item When the \texttt{Operator} is \texttt{Softmax} and $c = 0$, the architecture recovers the \textbf{Hedge algorithm} (\textit{i.e.}, \textbf{FTRL with entropy regularization});
\end{itemize}
since 
\begin{align*}
    \texttt{Operator}\left( \sum_{\tau=1}^{t-1} \left( \mathbf{A} R_\tau R_\tau^\intercal \mathbf{b} + \mathbf{C} R_\tau + \mathbf{d} \right) \right) = \texttt{Operator}\left(c' \sum_{\tau=1}^{t-1} R_\tau\right). 
\end{align*}

\begin{figure}[!t]
    \centering
     \includegraphics[width=\linewidth]{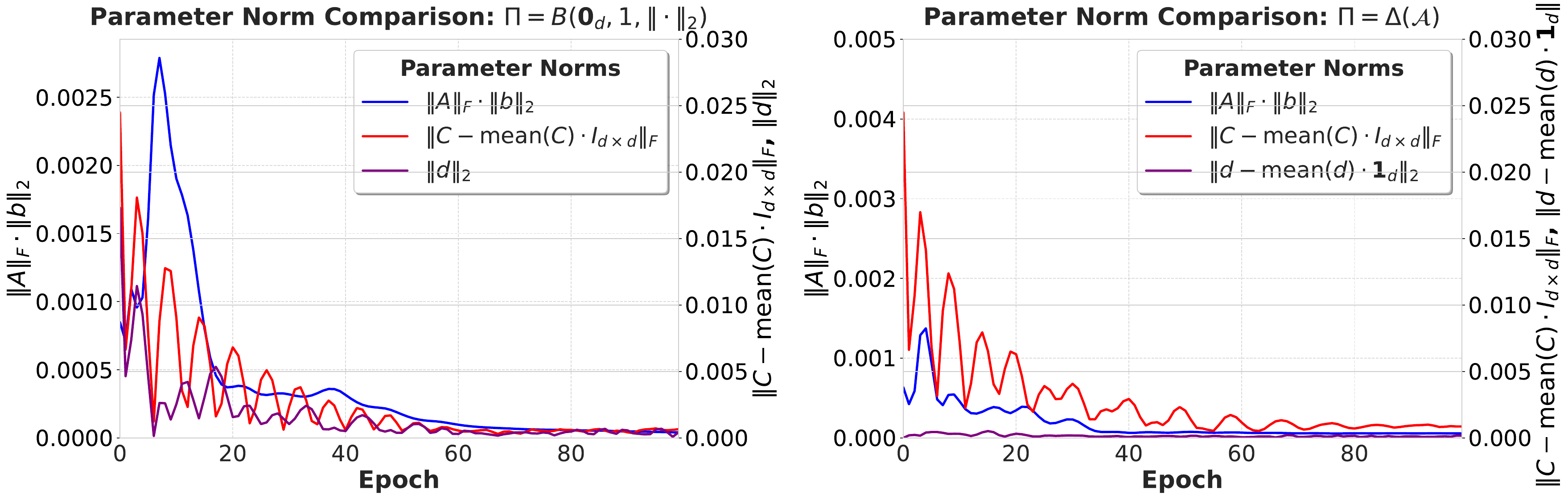}
   \caption{Evolution of the Transformer parameters with different policy spaces. \textbf{Left:} $\ell_2$-ball policy space $\Pi = B(\mathbf{0}_d, 1, \|\cdot\|_2)$ (\textbf{Env I}).  \textbf{Right:} Simplex policy space $\Pi = \Delta(\mathcal{A})$ (\textbf{Env II}).  
   } 
           \label{fig:parameter}
 \end{figure}
 We visualize the convergence behaviors of these quantities in \Cref{fig:parameter} along the training iterations of our \briefalgorithmexpand{}.  For \textbf{Env I}, we observe that $\norm{\mathbf{A}}_F \norm{\mathbf{b}}_2$, $\norm{\mathbf{C} - \text{mean}(\mathbf{C}) \cdot I_{d \times d}}_F$, and $\norm{\mathbf{d}}_2$ all converge to zero. This indicates that the Transformer's output asymptotically approximates that of FTRL  with $\ell_2$-regularization. For \textbf{Env II}, we similarly observe that $\norm{\mathbf{A}}_F \norm{\mathbf{b}}_2$, $\norm{\mathbf{C} - \text{mean}(\mathbf{C}) \cdot I_{d \times d}}_F$, and $\norm{\mathbf{d} - \text{mean}(\mathbf{d}) \cdot \pmb{1}_d }_2$ all approach approximately zero, suggesting that the Transformer's output behavior emerges to approximate as that of the Hedge algorithm.

\ifthenelse{\boolean{iclr}}
{
We defer the detailed regret analysis on our trained model to \Cref{appendix:ssec:regret_tranformer}. The model {trained by our \briefalgorithmexpand{}} exhibits both \texttt{Reward Generalization}, meaning that it achieves sublinear regret across diverse reward generation processes, and \texttt{Horizon Generalization}, meaning that it maintains sublinear regret on tasks with horizon $T=100$ despite being trained only on $T=25$ (\Cref{fig:simplex_tf_all,fig:ball_tf_all}). }
{

\begin{figure}[!h]
   \centering
    \includegraphics[width=\linewidth]{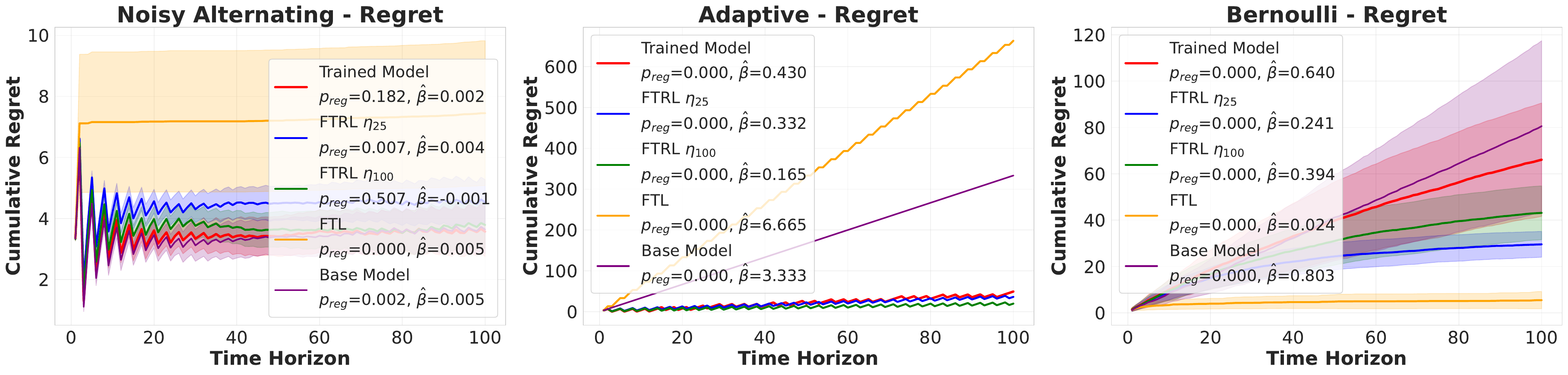}
\caption{\textbf{Evaluation of the trained Transformers with numerical input/output} for the FOL environment under \texttt{Horizon Generalization}[$T=25 \rightarrow T=100$] and \texttt{Reward Generalization} [a mixture of the \protect\mybluehyperlink{gaussianmu}{Gaussian}, \protect\mybluehyperlink{uniform}{Uniform} and \protect\mybluehyperlink{sine}{Sine-trend} rewards $\rightarrow$ \texttt{\protect\mybluehyperlink{noisy}{Noisy Alternating}}, \texttt{\protect\mybluehyperlink{adaptive}{Adaptive}}, and \texttt{\protect\mybluehyperlink{bernoulli}{Bernoulli}} rewards] under the simplex constraint $\Pi = \Delta(\mathcal{A})$ (\textbf{Env II}). The Transformer consistently demonstrates sublinear regret growth and competitive performance compared to the FTRL baseline. Here, $\eta_t = \sqrt{\frac{2 \log(d)}{t}}$, which corresponds to the entropy regularization parameter (or stepsize) for FTRL, and FTL refers to the Follow-the-Leader algorithm \citep{cesa2006prediction} (see \Cref{appendix_baseline} for a detailed introduction).}
\label{fig:regret-dynamics-noisy}
\end{figure}

To evaluate the sublinearity of the regret over time, we test our trained Transformer models across a diverse set of rewards, including previously unseen ones (\textit{i.e.},  \texttt{Reward Generalization}). To further assess the generalization capability of our trained Transformer {as an algorithm}, we {also} conduct evaluations on a longer horizon $T = 100$, while the model trains on horizon $T = 25$ (\textit{i.e.}, \texttt{Horizon Generalization}). 
We use the following rewards for testing: the \mybluehyperlink{gaussianmu}{Gaussian}, \mybluehyperlink{uniform}{Uniform}, \mybluehyperlink{bernoulli}{Bernoulli}, \mybluehyperlink{sine}{Sine-trend}, \mybluehyperlink{alternating}{Alternating}, \mybluehyperlink{noisy}{Noisy Alternating}, and \mybluehyperlink{adaptive}{Adaptive} rewards. 
We report the \texttt{Reward Generalization} and \texttt{Horizon Generalization} results in \Cref{fig:regret-dynamics-noisy}, which shows the regret over time and the final regret distribution for \textbf{Env II} under the \mybluehyperlink{noisy}{Noisy Alternating} reward. Our trained model consistently demonstrates sublinear regret growth across all tested rewards, often matching the performance of the FTRL baseline. {Note that, what the regret-over-time subfigure plots is $(t,\text{Regret}(t))$, with $\text{Regret}(t)$ being computed by the comparator at \emph{each around $t$}, instead of that at the \emph{final round $T$}. This way, the plot  by definition  shows \texttt{Horizon Generalization} results over 
varying horizons (\emph{i.e.,} for $T=26,\dots,100$).} 

We defer the complete set of regret plots for each reward to \Cref{appendix:sssec:full_result_sec4}. Additionally, \Cref{tab:simplex_table,tab:l2ball_table} present a statistical analysis of the regret over time, reporting slope coefficients ($\hat{\beta}$) and $p$-values (see \Cref{ssec:regret}), confirming the sublinear nature of regret exhibited by our trained Transformer with \texttt{Reward Generalization} and \texttt{Horizon Generalization}. 
For \textbf{Env I}, we observe similar sublinear regret {and good generalization behaviors}, with detailed results being deferred to \Cref{appendix:sssec:full_result_sec4}. 
}

\subsubsection{Multi-Armed Bandits}
\begin{figure}[!t]
    \centering
    \includegraphics[width=\linewidth]{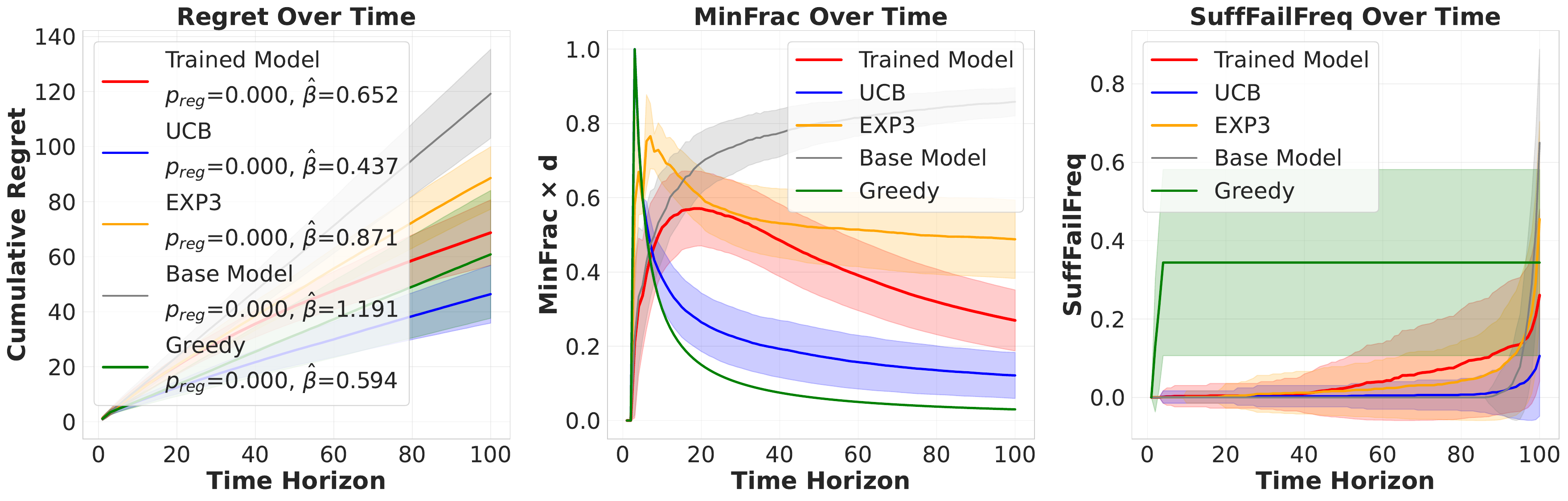}
    \caption{
        \textbf{Evaluation of the trained Transformers with numerical input/output for the MAB environment under \texttt{Horizon Generalization}[$T=25 \rightarrow T=100$]}, trained and evaluated with the  \protect\mybluehyperlink{gaussian}{Gaussian} reward.
        The trained model exhibits sublinear regret {growth as validated by our validation framework and comparison with baseline algorithms.} It also shows   
         {$\texttt{MinFrac}(t)$ also exhibits a trend that reflects a proper \textit{E-E tradeoff}: the metric}  
        first increases (active exploration in early rounds) and later decreases (progressive exploitation of optimal actions).  Meanwhile, $\texttt{SuffFailFreq}(t)$ maintains a consistently lower value than the Greedy and  the Base Model near the end of the horizon, indicating convergence toward more optimal action choices. Overall, these patterns demonstrate an improved \textit{E-E tradeoff} compared to baselines.
        }
    \label{fig:tfbandit}
\end{figure}

For the MAB environment, using \Cref{alg:ssft-numerical-model}, we train a single-layer linear attention model. 
{As shown in \Cref{fig:tfbandit},} 
our trained model exhibits sublinear regret in longer horizon (\emph{i.e.,} with  \texttt{Horizon Generalization}). Notably, the \texttt{MinFrac$(t)$} metric shows an \textit{E-E} trend: it first increases, reflecting active exploration in early rounds, and later decreases as the model progressively exploits the best actions. This improves upon the base model before training and aligns with the behavior of other baseline algorithms. The consistently lower \texttt{SuffFailFreq$(t)$} of our model compared to the greedy policy near the end of the horizon, indicates convergence toward more optimal action choices while preserving adequate exploration. We defer the result for the \mybluehyperlink{gamma}{Gamma} reward to \Cref{appendix:ssec-mab-results-tf}, where the same patterns can also be observed. 
It is worth noting that although we train our model with rewards from the \mybluehyperlink{gaussianmu}{Gaussian} reward and horizon $T=25$, it generalizes to \emph{different horizon lengths} and \emph{reward types}, demonstrating good  \texttt{Reward Generalization} and \texttt{Horizon Generalization} performance in such MAB environments. 

\subsection{Theoretical Insight: {From An Imitator to A Decision-Maker}}
\label{sec:theory}

{Motivated by the empirical findings in \Cref{sec:FOL_transformer}, where we observe that the single-layer linear attention architecture 
empirically recovers known online learning algorithms (\textit{e.g.}, Hedge, FTRL) under our post-training paradigm, 
we now provide a formal analysis to better understand this emergent behavior. Specifically, we show that in a simplified setting of 
\briefalgorithmexpand{}, our training paradigm can indeed provably recover FTRL.} 
{This result offers a theoretical support for \briefalgorithmexpand{}, 
by  
showing that the \emph{imitation}  of low-regret behaviors may  yield a no-regret learning \emph{algorithm}, 
which justifies the improved and generalizable no-regret behaviors observed empirically in  \Cref{ssec:experiment-results-sec4}.}

For analytical convenience, we study a simple yet canonical subclass of \Cref{eqn:single-linear-transformer} obtained by removing \texttt{Operator}. For any history length $t\ge1$, we define
\begin{align}
g_t\!\big((R_1,\dots,R_{t-1}, \pmb{1}_d);\, V,K,Q,v_c,k_c,q_c\big)
:= \sum_{\tau=1}^{t-1} \big(VR_\tau + v_c\big)\,\big((KR_\tau + k_c)^\top (Q\pmb{1}_d + q_c)\big),
\label{eqn:original-param}
\end{align}
where each $R_\tau\in\mathbb{R}^d$ denotes a reward vector in the FOL environment.  
We adopt the empty-sum convention, so $g_1((\pmb{1}_d);V,K,Q,v_c,k_c,q_c)=0$.  
Throughout the analysis, we assume $\{R_\tau\}_{\tau=1}^T$ are i.i.d. draws from $\mathcal{N}(\pmb{0}_d, I_{d\times d})$. We aim to learn model parameters that minimize the expected squared distance between the model’s output and the \emph{best-in-hindsight} policy. Specifically, we minimize the loss: 
\begin{align}
f(V, K, Q, v_c, k_c, q_c) 
:= \mathbb{E} \left[\sum_{t=1}^{T} \left\| g((R_1, \dots, R_{t-1}, \pmb{1}_d) ; V, K, Q, v_c, k_c, q_c) - \pi^\star(R_1, \dots, R_T) \right\|_2^2 \right], \label{eqn:loss-for-theory}
\end{align}
where the expectation is taken over the randomness of  $R_i$, and 
$
\pi^\star(R_1, \dots, R_t) {\in}   \arg\max_{\pi\in\Pi} \left\langle \pi, \sum_{i=1}^t R_i \right\rangle,$ 
which is a policy 
that maximizes the cumulative reward in hindsight. 

Note that, minimizing the above loss function corresponds to an  idealized regime of \Cref{alg:ssft-meta-algorithm}, when an infinite number of trajectories \( L \to \infty \) are sampled per scenario, and only the \emph{lowest-regret}  trajectories are {selected}. In this limit, the dataset used for supervised fine-tuning consists entirely of trajectories that are globally optimal (in terms of regret). We show that post-training the model on such a best-in-hindsight policy leads the model parameters to converge to those that implement FTRL. Thus, \briefalgorithmexpand{} in this regime does not merely imitate a good policy—it \textit{provably emerges} {as an \emph{algorithm} for online DM.}  
This result highlights {the insight} that \briefalgorithmexpand{}, when applied with sufficient samples and a well-structured inductive bias, may go beyond behavioral cloning/imitation learning, and {recover principled algorithms for online DM.} We defer the proof of \Cref{thm:single-layer-regret-minimizer} to \Cref{appendix:thm1pf}. 

\begin{restatable}{theorem}{linear}
\label{thm:single-layer-regret-minimizer} 
{Consider the policy space \( \Pi = B(\pmb{0}_d, R_{\Pi}, \norm{\cdot}_2)\) for some $R_{\Pi} > 0$, and consider the minimization of \Cref{eqn:loss-for-theory}, which corresponds to \Cref{alg:ssft-meta-algorithm} with infinite  data and $k=1$. Then, plugging in any  
global  minimizer of  \Cref{eqn:loss-for-theory} 
within the model class parameterized by \Cref{eqn:original-param}  
and projecting the resulting output using \( \texttt{Proj}_{\Pi, \norm{\cdot}} \)  {yield an output} 
{from} 
running FTRL with an $\ell_2$-regularizer and a stepsize of order $\Theta\left(\frac{1}{\sqrt{Td}}\right)$.}
\end{restatable}

\section{Training Open-Weight LLMs   
with \briefalgorithmexpand{}} \label{sec:open-weight-training}

In this section, we apply our \ours{} framework to train open-weight LLMs in {tasks with both FOL and MAB environments}. Compared to \Cref{sec:training-transformer}, which focused on Transformers with numerical input/output, we now explore more language-grounded numerical DM tasks, where we provide a language-based prompt to describe the numerical DM environment, and both input and output of the model  are also language-based. 
{Similar settings have also been considered recently in}  \citet{park2024llm, krishnamurthy2024can, nie2024evolve}{, to understand the in-context  
decision-making capabilities of LLM agents}.

We here focus on smaller open-weight LLMs to examine their \textit{trainability}—whether they can be effectively trained under our \briefalgorithmexpand{} paradigm to function as decision-makers. To do so, we begin with language-grounded numerical DM examples that capture essential decision-making structures, before contrasting these results with those in more complex, real-world contexts later in \Cref{sec:training-gpt}. The instantiation of \Cref{alg:ssft-meta-algorithm} for training such models in language-grounded numerical DM tasks is presented in  Algorithm~\ref{alg:ssft-language-model}, with a detailed explanation in \Cref{appendix:sec:openweight-explanation}. 

\subsection{Experimental Setup}
\label{ssec:setting-sec5}

At each round $t$, the model receives a natural language description of a DM problem together with the reward history $(R_1, R_2, \dots, R_{t-1})$ or the history of the observed rewards $(\hat{R}_1, \hat{R}_2, \dots, \hat{R}_{t-1})$. The model is then expected to produce a natural language response specifying the policy $\pi_{t}$ to be applied at round $t$. We refer to this task as \textit{language-grounded numerical DM}. 
We use the same numerical reward generation processes as described in \Cref{ssec:setting-sec4}.

\ifthenelse{\boolean{iclr}}
{We defer the prompt design and its ablation in \Cref{appendix:prompt-open-weight}}{}

\ifthenelse{\boolean{arxiv}}
{

In designing the {\color{orange}interaction protocol}, we adopt a 
\fcolorbox{orange!30}{orange!30}{\textit{dialogue-type} interaction}, in contrast to prior work that relied on \fcolorbox{orange!30}{orange!30}{\textit{summary-type} interaction} \citep{park2024llm, krishnamurthy2024can, nie2024evolve}. {At each round, a}  \fcolorbox{orange!30}{orange!30}{summary-type interaction} condenses the entire past interactions  into a \emph{single prompt} {that summarizes the past} \emph{actions} or \emph{policies} and their \emph{rewards}. In contrast, a \fcolorbox{orange!30}{orange!30}{dialogue-type} \fcolorbox{orange!30}{orange!30}{interaction} unfolds {the past interactions} {\emph{round by round}}, {including the prompts and responses of the models, oftentimes also with their \emph{reasoning} rationales, at each round. Such dialogue-type interaction protocols better resemble how LLMs are used as agents in real-world agentic applications \citep{yao2022react,shinn2024reflexion}, in which the reasoning rationales of the agents can be naturally integrated. Moreover, we also observe empirically that, \fcolorbox{orange!30}{orange!30}{summary-type protocols} are more prone to overfitting to summarized outputs after early iteration in the experiments, whereas \fcolorbox{orange!30}{orange!30}{dialogue-type protocols} can mitigate this issue and further offer additional advantages, possibly due to the incorporation of the reasoning process. We provide the experimental evidence in \Cref{appendix:summary-results}.}

For the {\color{blue}output type}, we focus on \fcolorbox{blue!30}{blue!30}{\textit{action-based outputs}}, where the model directly provides an action as its final {decision/action. This is in contrast to the  \fcolorbox{blue!30}{blue!30}{\textit{distributional outputs}}, where the model outputs a \emph{probability distribution} over the action space. 
Both types of outputs have been used in the literature \citep{park2024llm,krishnamurthy2024can,nie2024evolve}, with the \fcolorbox{blue!30}{blue!30}{\textit{action-based outputs}} being used often in MAB/bandit-feedback environments, while the \fcolorbox{blue!30}{blue!30}{\textit{distributional outputs}} being used in FOL environments, where randomization is necessary in the face of adversaries \citep{hazan2016introduction}. 
We focus on the former as we observe experimentally that some open-weight LLMs struggled to generate meaningful \fcolorbox{blue!30}{blue!30}{\textit{distributional outputs}} that reliably lie in the simplex (see a formal diagnosis in \Cref{appendix:distributional-results}). 
To make \fcolorbox{blue!30}{blue!30}{\textit{action-based outputs}} compatible with the FOL environments that require randomized policies, {we extract the top-5 token probabilities at the action output position, and re-normalize the probabilities over the subset of tokens corresponding to feasible actions.} 
}
Finally, for the {\color{red}output format}, we use the \fcolorbox{red!30}{red!30}{\textit{policy-with-reasoning}} output, which {includes the reasoning rationales in the responses, improving} 
the interpretability of LLM's output compared to the \fcolorbox{red!30}{red!30}{\textit{policy-only}} output{, as used in the literature \citep{krishnamurthy2024can,park2024llm,nie2024evolve}}. 
More detailed explanations on these settings and prompt designs can be found in \Cref{appendix:prompt-open-weight}. 
}{}

Each experiment trains the model for $T=25$ rounds with $d=3$ actions. For FOL environments, the training reward generation process is defined as a mixture of the \mybluehyperlink{gaussianmu}{Gaussian}, \mybluehyperlink{uniform}{Uniform}, and \mybluehyperlink{sine}{Sine-trend} rewards, whereas for MAB environments, we use the \mybluehyperlink{gaussianmu}{Gaussian} reward. 
We train three open-weight LLMs -- Phi-3.5-mini-instruct, Gemma-2-9b-it, and Qwen3-8B. Although these models are smaller and generally weaker than the proprietary closed-weight models, they remain among the most capable in their size category, exhibiting competitive performance in language understanding. All models are pre-trained and instruction-tuned, and are expected to effectively interpret natural language descriptions of the DM environments. 
\ifthenelse{\boolean{arxiv}}{

\subsection{Instantiating \briefalgorithmexpand{}  for Language-Grounded DM}
    \label{appendix:sec:openweight-explanation}
\begin{algorithm}[!h]
\caption{\ours{} for Language-Grounded DM (\Cref{ssec:setting-sec5}, \Cref{ssec:setting-sec6})}
\label{alg:ssft-language-model}
\begin{algorithmic}[1]
\State \textbf{Input:} A DM task (\textit{e.g.,} FOL, MAB, NS-MAB), LLM parameterized by $\theta_{0}$ %
\For{iteration $= 0, 1, 2, \dots$}
    \State $\cD = \emptyset$
    \For{scenario index $i = 1, 2, \dots, M$}
        \State Sample $L$ trajectories $C_{1, i}, \dots, C_{L, i}$ from the model parameterized by $\theta_{\text{iteration}}$ under scenario$_i$ using temperature $\tau$ for sampling. Each trajectory $C_{j, i} = (u_{j,i,0}, m_{j,i,1}, u_{j,i,1}, m_{j,i,2}, \dots, u_{j,i,T-1}, m_{j,i,T}, u_{j,i, T})$, where $u_{j,i,0}$ is the input prompt, $u_{j,i,t}$  for $t \geq 1$ is feedback for round $t$, and $m_{j,i,t}$ is the output from the model parameterized by $\theta_{\text{iteration}}$ at round $t$ for trajectory $j$ under scenario$_i$.
        \State For each $j \in [L]$, parse policies from $(m_{j, i, t})_{t \in [T]}$ and compute regret (\Cref{ssec:regret}) under scenario$_i$. Then, select the $k$ trajectories with the lowest regret: $C_{(1),i}, \dots, C_{(k),i}$
        \State Update dataset: $\cD \gets \cD \cup \{\{\text{scenario}_i, C_{(1),i }, \dots, C_{(k),i} \}\}$
    \EndFor
    \State  Update model parameters to $\theta_{\text{iteration}+1}$ by maximizing the log-likelihood starting from $\theta = \theta_{\text{iteration}}$. The log-likelihood is defined as 
        \[
\sum_{(\text{scenario}_i, C_{(j),i}) \in \cD} \sum_{t=1}^T \sum_{s = 1}^{\text{len}(m_{(j),i,t})} \log \text{model}_{\theta}\left(m_{(j),i,t,s} \mid u_{(j), i, 0}, (m_{(j),i,r}, u_{(j),i,r})_{r \in [t-1]}, m_{(j),i,t,1:(s-1)}\right)
        \]       
    \text{where} $m_{j,i,t} = (m_{j,i,t,1}, \dots, m_{j,i,t,\text{len}(m_{j,i,t})})$ denotes the token-level decomposition of $m_{j,i,t}$, and $\text{model}_\theta$ outputs the logits for next-token prediction.
\EndFor
\end{algorithmic}
\end{algorithm}

{To instantiate \Cref{alg:ssft-meta-algorithm} in the language-grounded setting, we generate multiple trajectories per scenario by repeatedly prompting the LLMs with stochastic decoding. We sample the output from the model using temperature $\tau = 1.0$ to encourage the diversity of the outputs, which will be used as training data. Note that this diversity affects the exploratory property of the training data, and is distinct from the exploration to be performed by the agent at inference time. 
The complete instantiation of \Cref{alg:ssft-meta-algorithm} is detailed in Algorithm~\ref{alg:ssft-language-model}. Unlike the numerical setting described in \Cref{ssec:setting-sec4}, where a Transformer directly outputs numerical policies, the language-grounded trajectories consist of text outputs. These outputs implicitly define a policy, which is then parsed and executed in the environment to produce feedback.  
Post-training is then performed via supervised fine-tuning: minimizing the cross-entropy loss between the model’s predicted token distributions and those from the top-$k$ low-regret trajectories. This training objective encourages the model to imitate high-performing behaviors and reasoning rationales (in terms of regret) encoded in the natural language form.}

}{}

\subsection{Experimental Results}
\label{ssec:exp-results-openweight}

{Across all reported results in this section and its corresponding appendices, we assess performance using $50$ trajectory samples for open-weight LLMs. We report the maximum and average values of the cumulative regret at round $T$, {denoted by} \emph{max(LR)} and \emph{avg(LR)}, as well as  $\hat\beta$, the estimate of the \emph{growth rate}  $\beta$ of the regret over time in \Cref{ssec:regret}, to validate the regret performance of the trained models/DM algorithms.} The \fcolorbox{yellow!30}{yellow!30}{highlighted (yellow) cells} indicate the lowest values of {max(LR)}, {avg(LR)}, and $\hat\beta$ of the base model and the trained model. We also provide figures of the regret over time, and the final regret distributions from the trained and base models, with the Kolmogorov–Smirnov (KS) test results to compare them. A detailed explanation of the KS test is provided in \Cref{appendix:ssec:ks_explanation}. A lower $p$-value of the KS test indicates stronger statistical evidence that the trained model yields lower regret. We further report 3-replicate robustness checks with per-setting error bars in \Cref{appendix:significance-tests}. {For the MAB environment, we additionally provide the exploration metric \texttt{SuffFailFreq$(t)$} {before and after our training}. As we adopt \fcolorbox{blue!30}{blue!30}{\textit{action-based outputs}} throughout this section, we argue that the exploitation metric \texttt{MinFrac$(t)$} is already adequately low for the base model (see \Cref{fig:open-MAB-minfrac}), in line with the observations of \citet{krishnamurthy2024can}, and therefore we omit this result here. 

As baselines, we also include the performance of FTRL for the FOL environment and UCB for the MAB environment. {Importantly, note that this comparison is in favor of these baseline algorithms, as they directly take the \emph{numeric} descriptions of the DM environments (\emph{i.e.,} reward vectors/values) as input, for which they are known to be asymptotically optimal in terms of regret minimization.} 
Hence, these algorithms should be viewed as \emph{reference points}  when the numeric descriptions of the problem are perfectly accessible, without distraction or ambiguity due to language descriptions, instead of some baselines to be \emph{beaten} by our language-grounded, post-trained LLM agents.} 
{Finally, in terms of the generalizability of the trained LLM-agents,  
our evaluation will focus on two axes: \texttt{Reward Generalization} 
and \texttt{Horizon Generalization}.}

\paragraph{Full-Information Online Learning.} 

We evaluate the \texttt{Horizon Generalization} and \texttt{Reward Generalization} performance of our trained model with $d=3$ actions and a time horizon of $T=50$, across multiple different reward generation processes on Gemma-2-9b-it. We first provide the result using the same reward generation process as the training one in \Cref{tab:open-FOL-indist} (which corresponds to \texttt{Horizon Generalization}). The regret over time and the final regret distribution can also be found in \Cref{fig:open-FOL-indist}. The trained model exhibits lower regrets than the base model in all three reward generation processes, which demonstrates the effectiveness of \briefalgorithmexpand{}. We also evaluate the trained Gemma-2-9b-it model on the \mybluehyperlink{alternating}{Alternating}, \mybluehyperlink{bernoulli}{Bernoulli}, \mybluehyperlink{noisy}{Noisy Alternating} rewards in \Cref{fig:open-FOL-alternating,fig:open-FOL-bernoulli,fig:open-FOL-noisy-alternating} (which corresponds to both \texttt{Horizon Generalization} and \texttt{Reward Generalization}). We observe lower regret values for the trained model for all the reward generation processes. 

{
\begin{table}[h!]
\centering
\resizebox{\textwidth}{!}{
\begin{tabular}{l|ccc||ccc||ccc}
\thickhline
\multicolumn{1}{c|}{} & \multicolumn{3}{c||}{\textbf{\mybluehyperlink{gaussianmu}{Gaussian}}} & \multicolumn{3}{c||}{\textbf{\mybluehyperlink{uniform}{Uniform}}} & \multicolumn{3}{c}{\textbf{\mybluehyperlink{sine}{Sine-trend}}} \\ \cline{2-10}
\multicolumn{1}{c|}{} & max(LR) & avg(LR) & $\hat{\beta}$ & max(LR) & avg(LR) & $\hat{\beta}$ & max(LR) & avg(LR) & $\hat{\beta}$ \\ \hline
FTRL & 49.70 & 27.99 & 0.81 & 55.04 & 38.90 & 0.75 & 54.33 & 12.32 & 0.31 \\
\cline{1-10}
Gemma-2-9b-it & 137.19 & 20.62 & 0.87 & 227.48 & 28.47 & 0.81 & 186.81 & 18.00 & 0.48 \\
Trained Gemma-2-9b-it & \cellcolor{yellow!30}93.08 & \cellcolor{yellow!30}20.45 & \cellcolor{yellow!30}0.80 & \cellcolor{yellow!30}122.45 & \cellcolor{yellow!30}12.05 & \cellcolor{yellow!30}0.57 & \cellcolor{yellow!30}139.79 & \cellcolor{yellow!30}13.28 & \cellcolor{yellow!30}0.38 \\ \thickhline
\end{tabular}
}
\caption{\textbf{{Summary of the regret values for the FOL environment evaluated with \texttt{Horizon Generalization} [$T=25\rightarrow T=50$]  on the Gemma-2-9b-it} model}, trained and tested on a mixture of the \protect\mybluehyperlink{gaussianmu}{Gaussian}, \protect\mybluehyperlink{uniform}{Uniform} and \protect\mybluehyperlink{sine}{Sine-trend} rewards. The results with the \protect\mybluehyperlink{uniform}{Uniform} and \protect\mybluehyperlink{sine}{Sine-trend} rewards both demonstrate a lower regret value and sublinear regret growth after \briefalgorithmexpand{}, while that of the \protect\mybluehyperlink{gaussianmu}{Gaussian} reward shows a better maximum regret value and sublinear regret growth, but does not significantly lower the average regret value. }
\label{tab:open-FOL-indist}
\end{table}
}

\begin{figure}[H]
   \centering
   \includegraphics[width=0.9\linewidth]{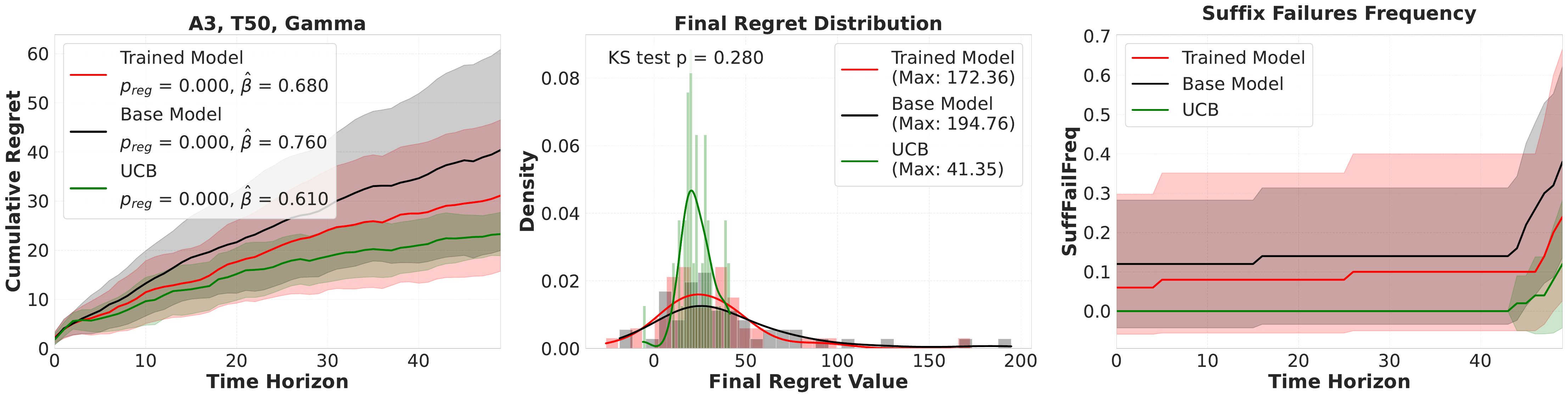}
   \caption{\textbf{The regret over time, the final regret distribution, and the exploration metric \texttt{SuffFailFreq$(t)$} for the MAB environment, under both \texttt{Horizon Generalization} [$T=25 \rightarrow T=50$] and \texttt{Reward Generalization} [\protect\mybluehyperlink{gaussianmu}{Gaussian} $\rightarrow$ \protect\mybluehyperlink{gamma}{Gamma}] on the Gemma-2-9b-it model}, which shows a lower regret value, sublinear regret growth, and improved exploration after \briefalgorithmexpand{}.}
   \label{fig:open-MAB-gamma-T50-re}
\end{figure}

\begin{figure}[H]
   \centering
   \includegraphics[width=0.9\linewidth]{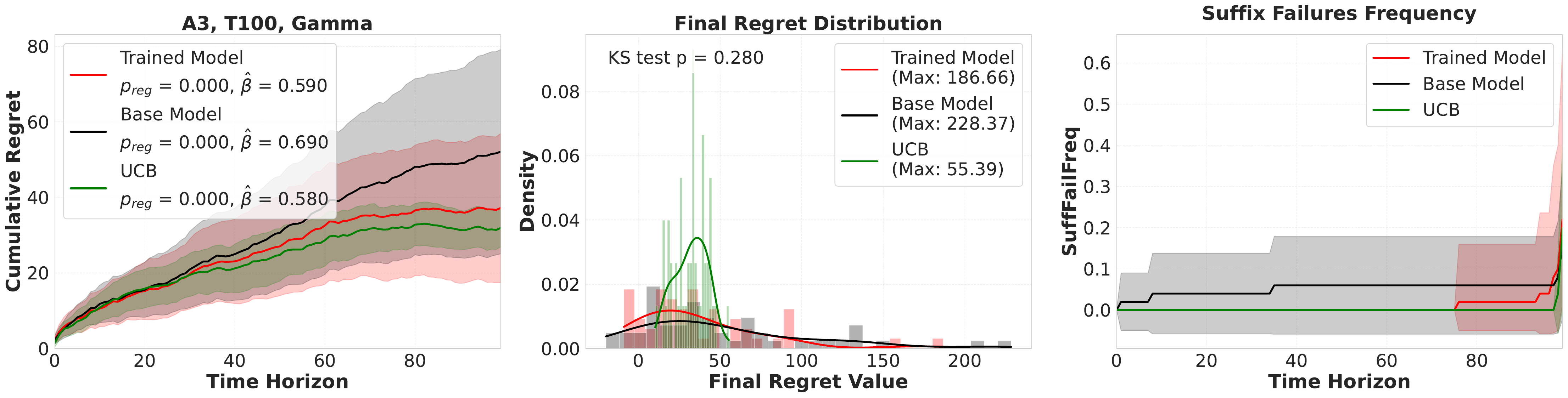}
   \caption{\textbf{The regret over time, the final regret distribution, and the exploration metric \texttt{SuffFailFreq$(t)$} for the MAB environment under both \texttt{Horizon Generalization}[$T=25 \rightarrow T=100$] and \texttt{Reward Generalization}[\protect\mybluehyperlink{gaussianmu}{Gaussian} $\rightarrow$ \protect\mybluehyperlink{gamma}{Gamma}] on Qwen3-8B}, which shows a lower regret value, sublinear regret, and improved exploration after \briefalgorithmexpand{}. }
   \label{fig:qwen3-gamma-T50-re}
\end{figure}

\paragraph{Multi-Armed Bandits.}
Similar to the FOL environment, we evaluate both \texttt{Horizon Generalization} and \texttt{Reward Generalization} [\mybluehyperlink{gaussianmu}{Gaussian} $\rightarrow$ \mybluehyperlink{gamma}{Gamma}] performance with $d=3$ actions on both Gemma-2-9b-it and Qwen3-8B models. Because the two LLMs differ in their context length limits, we set the time horizon to $T=50$ for Gemma-2-9b-it and $T=100$ for Qwen3-8B. The Gemma-2-9b-it model is limited to a context length of 8192 tokens, therefore we cannot run experiments at $T=100$. We provide the experimental results of the  \mybluehyperlink{gamma}{Gamma} reward in \Cref{fig:open-MAB-gamma-T50-re} (Gemma-2-9b-it) and \Cref{fig:qwen3-gamma-T50-re} (Qwen3-8B). We find that the model achieves lower regret values and slower regret growth rates, while also demonstrating improved exploration, as evidenced by the reduced \texttt{SuffFailFreq$(t)$} at smaller time steps $t$, and also exhibits good \texttt{Horizon Generalization} performance. We also provide results for the \protect\mybluehyperlink{gaussianmu}{Gaussian} reward on both Gemma-2-9b-it and Qwen3-8B models in \Cref{fig:open-MAB-gaussian-T50-gemma,fig:open-MAB-gaussian-T50-qwen}. 

\paragraph{AD and GRPO Baselines.}
We further compare \briefalgorithmexpand{} against algorithm distillation (AD) \citep{laskin2022context,nie2024evolve} and GRPO-style baselines \citep{shao2024deepseekmath,guo2025deepseek} on the Qwen3-8B MAB experiment with a UCB teacher \citep{auer2002finite}, where all methods are trained with horizon $T=25$ and evaluated under \texttt{Horizon Generalization} at $T=100$. Let $I$ denote the number of iterations, $M$ the number of instances per iteration, $L$ the number of samples per instance, and $k$ the number of selected trajectories. Thus, \briefalgorithmexpand{} uses $IMk$ trajectories for SFT after generating $IML$ trajectories.

\begin{table}[h!]
\centering
\resizebox{\textwidth}{!}{
\begin{tabular}{l|ccccccc}
\thickhline
\textbf{Method} & \textbf{Base} & \textbf{AD$_{IMk}$} & \textbf{AD$_{IML}$} & \textbf{AD$_{2IML}$} & \textbf{GRPO$_{\text{step}}$} & \textbf{GRPO$_{\text{regret}}$} & \textbf{Ours} \\ \hline
Avg Regret ($T=100$) & 52.10 & 58.72 & 33.44 & 32.97 & 35.88 & 40.09 & \cellcolor{yellow!30}\textbf{22.37} \\
\thickhline
\end{tabular}
}
\caption{
\textbf{Comparison with algorithm distillation and GRPO-style baselines on Qwen3-8B in the MAB environment.}
All methods are trained with $T=25$ and evaluated at $T=100$ using the UCB teacher setup. \briefalgorithmexpand{} substantially outperforms AD with the matched selected-data budget ($IMk$), AD with the matched generation budget ($IML$), AD with twice the generated budget ($2IML$), and both GRPO variants. The $GRPO_{\text{regret}}$ run uses the $IMk$ budget due to time constraints.
}
\label{tab:qwen-ad-grpo-comparison}
\end{table}

This comparison suggests that the gains of \briefalgorithmexpand{} are not explained solely by the generation budget: although \briefalgorithmexpand{} generates more trajectories than $AD_{IMk}$ in order to perform selection, it outperforms $AD_{IML}$ with a matched generation budget and $AD_{2IML}$ with twice the generation budget. The efficiency likely comes from two factors: selective SFT on low-regret trajectories induces minimal policy drift, while AD trained to imitate UCB at $T=25$ suffers from a horizon mismatch when evaluated at $T=100$, since UCB's exploration schedule is coupled with the history length. We provide additional parameter-sensitivity ablations in \Cref{appendix:param-sensitivity}.


\begin{figure}[!h]
   \centering
   \includegraphics[width=\linewidth]{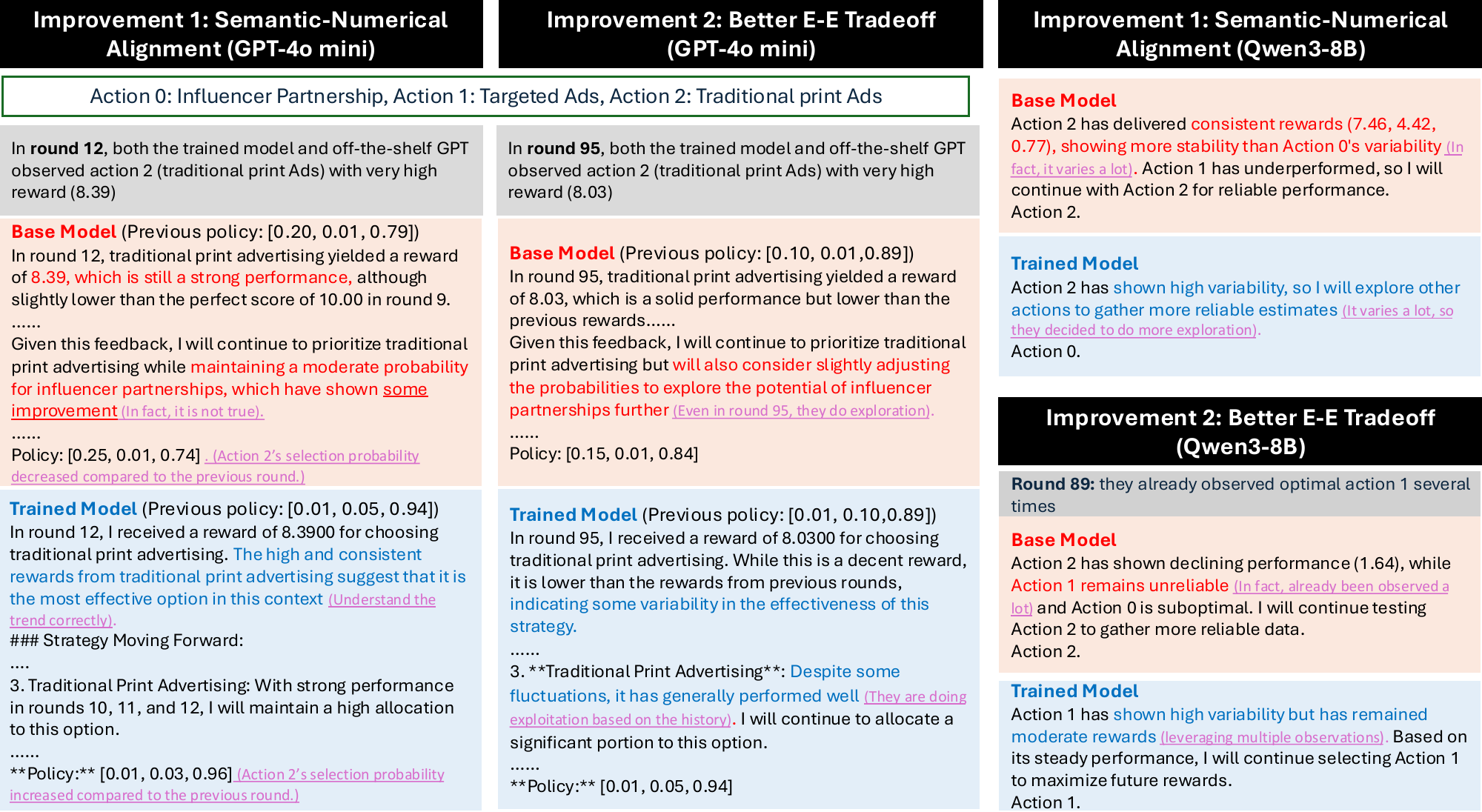}
   \caption{\textbf{Illustration of the reasoning rationales generated by the base model and the trained model in the MAB environment for both \textbf{GPT-4o mini} and \textbf{Qwen3-8B}.} 
   The figure highlights two major improvements observed after regret-based post-training: (1) enhanced \textit{semantic--numerical alignment}, and (2) improved \textit{E-E tradeoff}. 
   \textcolor{red}{Red text} indicates incorrect or inconsistent reasoning by the base model, \textcolor{blue}{blue text} denotes correct and reward-aligned reasoning by the trained model, and \textcolor{purple}{purple text} provides explanatory annotations clarifying why each response is good or bad.}
   
   \label{fig:reasoning-comparison}
\end{figure}

\paragraph{Improved Reasoning Rationales.} 
Lastly, we visualize representative reasoning rationales produced by the base and trained models of Qwen3-8B in \Cref{fig:reasoning-comparison}. The color-coded examples illustrate how our post-training paradigm improves the reasoning of the models in terms of 
both the \textit{semantic–numerical alignment} and the \textit{E–E tradeoff}. Here, \textit{semantic–numerical alignment} reflects the model's capability to accurately interpret numerical concepts from  language descriptions, while the \textit{E–E tradeoff} captures the model's decision-making ability in online DM tasks. We further present a quantitative analysis demonstrating the model’s enhanced robustness when dealing with unstable rewards, as detailed in \Cref{appendix:reasoning}. 

\section{Training the Closed-Weight LLM with \briefalgorithmexpand{}}
\label{sec:training-gpt}

{In this section, we apply the \ours{} framework to post-train the closed-weight LLM GPT-4o mini, which does not yet expose its full model weights but supports \emph{supervised fine-tuning} via API access. Notably, our \briefalgorithmexpand{} is inherently compatible with such a supervised fine-tuning paradigm (see \Cref{line:sft} in \Cref{alg:ssft-meta-algorithm}), making it possible for us to post-train such proprietary, but arguably stronger, closed-weight  models for DM.}

Unlike the training for open-weight LLMs in \Cref{sec:open-weight-training}, which focused on \textit{language-grounded numerical DM} (\Cref{ssec:setting-sec5}), {thanks to the stronger language ability of the closed-weight LLMs,} we now consider more complex and context-rich DM scenarios expressed in natural language, which we term \textit{language-grounded DM with real-world contexts}.

{Although closed-weight LLMs have demonstrated strong general-purpose language capabilities, prior studies \citet{park2024llm, krishnamurthy2024can,nie2024evolve} have shown that such models may struggle even in simple language-grounded DM tasks with specific contexts. 
Indeed,  these studies have largely relied on simple and \emph{fixed}   examples of contexts  (\textit{e.g.}, button-clicking \citep{krishnamurthy2024can}) that lack the diversity and complexity needed for the generalization to more  real-world DM tasks. 
In contrast, our goal is to fine-tune these models, via our \briefalgorithmexpand{}, to be more generalizable to a wider range of language-based DM tasks 
with diverse real-world contexts. To this end, we first need to curate new language-grounded DM datasets, the systematic generation of which may be a contribution of independent interest.}

\subsection{Experimental Setup: Language-grounded DM with Real-World Contexts}
\label{ssec:setting-sec6}
\begin{figure}[!h]
    \centering
    \includegraphics[width=\linewidth]{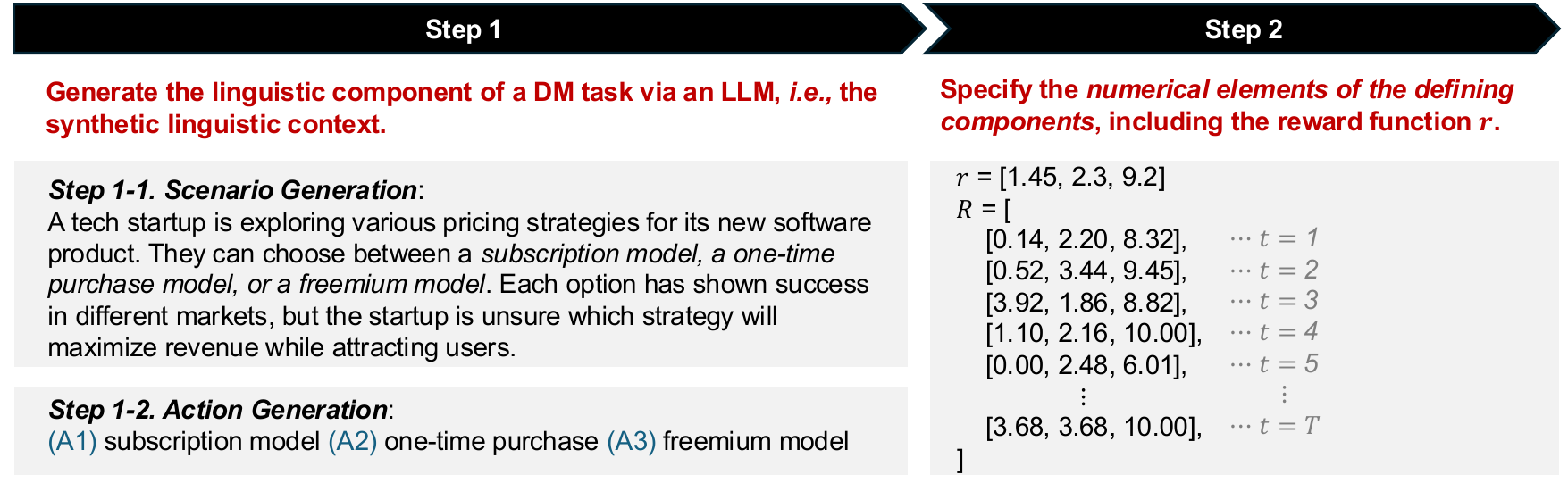}
    \caption{
    Illustrative example of the \emph{language-grounded DM with real-world contexts} 
    used for training GPT-4o mini using   \briefalgorithmexpand{}. 
    }
    \label{fig:sdm-example}
\end{figure}

We propose the following two steps to systematically create diverse language-grounded DM tasks with real-world contexts:   
\begin{enumerate}
  \item[{\color{red} \textbf{Step 1}}] \textbf{Linguistic Context Generation.} Generate the {\emph{linguistic component} of a DM task via an LLM, \emph{i.e.,} the synthetic \emph{linguistic context}  
  that specifies the {scenario} and the defining components of the task.} The defining components include the action space {$\cA$}, 
  but does not specify the \textit{numerical elements} of the DM task such as the reward functions.%
  \item[{\color{red} \textbf{Step 2}}] \textbf{Reward Generation Processes.} Specify the \emph{numerical elements of the defining components}, including the reward functions, 
  which are systematically instantiated from predefined random distributions.
  \end{enumerate}
\ifthenelse{\boolean{arxiv}}{In the remainder of this subsection, we detail the procedures {of the two steps above.}

\ifthenelse{\boolean{iclr}}{\subsection{Full-Information and Bandit Tasks}}{}
\label{sssec:ol-gen}
\paragraph{{Linguistic Context} Generation.}
We adopt a unified procedure to construct linguistic contexts for both the full-information online learning (FOL) and (non-)stationary multi-armed bandit (MAB/NS-MAB) environments. Using the GPT-4o mini API, we synthesize scenario descriptions that specify the common action set shared by the environments. Manually designed prompt templates span domains such as healthcare, business strategy, and science fiction, and they explicitly instruct GPT-4o mini to describe multiple plausible actions so that no candidate is obviously optimal. We further randomize the narrative style—ranging from concise reports to analytical dilemmas—to encourage linguistic and structural diversity. These prompts apply verbatim across FOL and MAB/NS-MAB tasks; the only difference is how the generated scenarios are paired with subsequent reward signals. Complete prompt templates appear in \Cref{appendix:prompt_scene_generation}, and we validate that this richer prompt induces substantially more diverse linguistic contexts than a simple prompt baseline in \Cref{appendix:prompt-diversity-ablation}. 

\paragraph{Reward Generation Processes.}
To keep the linguistic description decoupled from the numerical feedback, we instantiate rewards independently of the generated narratives, drawing from the distributions in \Cref{appendix:reward_dist}. This separation discourages the model from exploiting the superficial textual cues when predicting rewards.

}
{We defer the detailed {introduction} to the dataset generation process for each DM {environment} in \Cref{appendix:ssec:datagen}. \cp{not setup... but will change }\kz{yes pls change..}} We provide illustrative examples of the language-grounded DM scenarios used in our experiments in \Cref{fig:sdm-example}. 

\subsection{Experimental Results}
\label{ssec:sec6results}

Similar to \Cref{ssec:exp-results-openweight}, we report \emph{max(LR)}, \emph{avg(LR)}, the \emph{growth rate estimate} of regret $\hat\beta$, and for tasks with MAB environments, we additionally report \texttt{SuffFailFreq$(t)$} and \texttt{MinFrac$(t)$} (\Cref{ssec:regret}). Further training details are provided in \Cref{appendix:sec6-training-details}. The evaluation focuses on: (a) \texttt{In-Distribution} performance, to confirm whether effective training has occurred, and (b) four axes of generalization: \texttt{Reward Generalization} and  \texttt{Horizon Generalization} as in \Cref{sec:open-weight-training}, as well as \texttt{Linguistic Context Generalization} (\emph{i.e.,} testing under diverse real-world description distributions), and  \texttt{Action Space Size  Generalization}. Notably, the last two generalization dimensions are unique to language-grounded DM with real-world contexts and do not typically arise in numerical DM settings. {As noted before,  our intention is not to claim superior performance of the trained model over the  
classical algorithms such as UCB or EXP3. 
Instead, our main goal is to compare the performance of the LLMs \emph{before} and \emph{after} \briefalgorithmexpand{}, using these classical algorithms as references (the \emph{favorable} cases), as 
they directly take \emph{numeric} values as input and do not need to parse language descriptions.}    

\begin{figure}[!h]
   \centering
   \includegraphics[width=\linewidth]{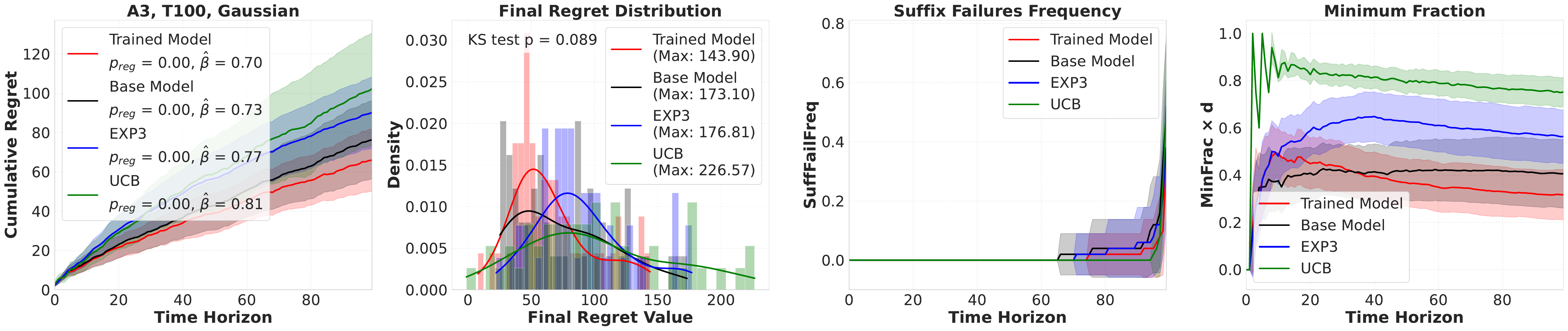}
   \caption{
   \textbf{The regret over time, the final regret distribution, and the exploration and exploitation metrics for tasks with the MAB environment under \texttt{Horizon Generalization}[$T=25 \rightarrow T=100$] on GPT-4o mini}, trained and evaluated with the \protect\mybluehyperlink{gaussianmu}{Gaussian} reward, which shows a lower regret and sublinear regret behavior for the trained model. The trained GPT-4o mini model sustains effective exploration and adapts its exploitation strategy over time.}
   \label{fig:Bandit-ood-gaussian}
\end{figure}

\paragraph{Full-Information Online Learning.}
The model is trained with $d=3$ actions and a time horizon of $T=15$ using a mixture of the \mybluehyperlink{gaussianmu}{Gaussian}, \mybluehyperlink{uniform}{Uniform}, and \mybluehyperlink{sine}{Sine-trend} rewards, with policy space $\Pi= \Delta(\cA)$. 
We evaluate \texttt{Reward Generalization} by altering the reward generation processes to the \mybluehyperlink{bernoulli}{Bernoulli} and \mybluehyperlink{alternating}{Alternating} rewards, evaluate \texttt{Horizon Generalization} by extending the time horizon to 25, evaluate  \texttt{Linguistic Context Generalization} by generating linguistic contexts by Gemini 2.0-Flash \citep{team2024gemini} rather than GPT-4o mini, and evaluate \texttt{Action Space Size Generalization} by increasing the number of actions to 4. Overall generalization results are summarized in \Cref{tab:FOL-ood-horizon,tab:FOL-ood-gemini,tab:FOL-ood-action}. Across these diverse and unseen conditions, the trained model consistently demonstrates 
competitive performance.
\begin{table}[h!]
   \centering
   \resizebox{\textwidth}{!}{%
   \begin{tabular}{l|ccc||ccc||ccc}
   \thickhline
   \multicolumn{1}{c|}{} & \multicolumn{3}{c||}{\textbf{\mybluehyperlink{gaussianmu}{Gaussian}}} & \multicolumn{3}{c||}{\textbf{\mybluehyperlink{uniform}{Uniform}}} & \multicolumn{3}{c}{\textbf{\mybluehyperlink{sine}{Sine-trend}}} \\ \cline{2-10}
   \multicolumn{1}{c|}{} & max(LR) & avg(LR) & $\hat{\beta}$ & max(LR) & avg(LR) & $\hat{\beta}$ & max(LR) & avg(LR) & $\hat{\beta}$ \\ \hline
   FTRL & 39.59 & 27.21 & 0.64 & 39.15 & 24.16 & 0.75 & 38.09 & 11.32 & 0.43 \\
   GPT-4o mini & 57.36 & 27.42 & 0.67 & 70.82 & 22.28 & 0.74 & 40.62 & 11.24 & 0.43 \\
   Trained GPT-4o mini & \cellcolor{yellow!30}53.29 & \cellcolor{yellow!30}25.63 & \cellcolor{yellow!30}0.62 & \cellcolor{yellow!30}39.65 & \cellcolor{yellow!30}17.09 & \cellcolor{yellow!30}0.65 & \cellcolor{yellow!30}39.64 & \cellcolor{yellow!30}9.83 & \cellcolor{yellow!30}0.38 \\
   \thickhline
   \end{tabular}%
   }
   \caption{
   \textbf{Summary of the regret value for the FOL environment under \texttt{Horizon Generalization}[$T=15 \rightarrow T=25$] on GPT-4o mini,} trained and evaluated on a mixture of the \protect\mybluehyperlink{gaussianmu}{Gaussian}, \protect\mybluehyperlink{uniform}{Uniform}, and \protect\mybluehyperlink{sine}{Sine-trend} rewards. The trained model consistently shows improved regret behavior in a longer horizon. 
   }
   \label{tab:FOL-ood-horizon}
   \end{table}
\paragraph{Multi-Armed Bandits.} 
We train our model using $d=3$ actions and a short horizon $T=25$ using the \mybluehyperlink{gaussianmu}{Gaussian} reward. 
First, we evaluate the \texttt{Horizon Generalization}[$T = 25 \to T = 100$] performance and report the results in \Cref{fig:Bandit-ood-gaussian}, which presents $\texttt{SuffFailFreq}(t)$ and $\texttt{MinFrac}(t)$.  
In particular, $\texttt{MinFrac}(t)$ reveals a key difference in exploitation strategies: the base GPT-4o mini model exhibits \textit{uniform-like failures} \citep{krishnamurthy2024can}, \emph{i.e.,} {the agent continues to select all arms at approximately equal rates without effectively eliminating the suboptimal choices}. In contrast, the trained GPT-4o mini model initially explores broadly (higher $\texttt{MinFrac}(t)$) but later concentrates its selections, as indicated by the decline in $\texttt{MinFrac}(t)$, reflecting a transition from exploration to exploitation, similar to the behaviors of known online DM algorithms of EXP3 and UCB. In addition, as shown in \Cref{fig:Bandit-ood-gaussian}, the trained GPT-4o mini model continues to exhibit favorable regret performance, achieving a lower maximum regret. {Then, we evaluate  \texttt{Reward Generalization} by altering the reward generation processes to the \mybluehyperlink{bernoulli}{Bernoulli} one, and evaluate \texttt{Linguistic Context Generalization} by generating linguistic contexts by Gemini 2.0-Flash. We also increase the action space to 4 and 5 with longer horizons  to further evaluate \texttt{Action Space Size Generalization} (see \Cref{fig:Bandit-ood-bernoulli,fig:Bandit-ood-gemini,fig:Bandit-ood-action}}).

\paragraph{Non-Stationary Multi-Armed Bandits.}
We train our model using $d=3$ actions and a short horizon $T=25$ with the \mybluehyperlink{gradual}{Gradual Variation} reward generation process.
The trained GPT-4o mini model sustains effective exploration and adapts its exploitation strategy over time.
Our result shows an improved regret growth rate (from 0.96 to 0.91) compared to the base model {with} \texttt{Horizon Generalization}.
Note that the \protect\mybluehyperlink{gradual}{Gradual Variation} reward used here yields a regret lower bound of $\Omega(T^{5/6})$, obtained by substituting $V_T=\Theta(T^{1/2})$ (see \Cref{appendix:reward_dist}) into the lower bound in \citet{besbes2014stochastic}.
As such, the trained model sustains competitive performance under this non-stationary environment.
Furthermore, we report $\texttt{SuffFailFreq}(0.98T)$ values, indicating that the model more reliably identifies the optimal arm near the end of the horizon compared to the base model.
The quantitative results are summarized in \Cref{tab:nsmab}, and for completeness, corresponding plots are provided in \Cref{fig:nonstationary-ood-gradual,fig:nonstationary-ood-gemini,fig:nonstationary-ood-action}.

\begin{table}[h!]
   \centering
   \resizebox{\textwidth}{!}{%
   \begin{tabular}{l|ccc||ccc||ccc}
   \thickhline
   \multicolumn{1}{c|}{} & \multicolumn{3}{c||}{\textbf{LC=Gemini, T=50, A=3}} & \multicolumn{3}{c||}{\textbf{LC=GPT, T=100, A=3}} & \multicolumn{3}{c}{\textbf{LC=GPT, T=100, A=5}} \\ \cline{2-10}
   \multicolumn{1}{c|}{} & max(LR) & $\hat{\beta}$ & \texttt{SuffFailFreq}&  max(LR) & $\hat{\beta}$ & \texttt{SuffFailFreq}&  max(LR) & $\hat{\beta}$ & \texttt{SuffFailFreq} \\ \hline
   Rexp3 & 214.10 & 0.89 & 0.32 & 403.45 & 0.86 & 0.26 & 560.95 & 0.88 & 0.46 \\
   GPT-4o mini & 277.74 & 0.88 & 0.22 & 427.91 & 0.96 & \cellcolor{yellow!30}0.18 & 453.72 & 0.90 & 0.32 \\
   Trained GPT-4o mini & \cellcolor{yellow!30}204.74 & \cellcolor{yellow!30}0.86 & \cellcolor{yellow!30}0.22 & \cellcolor{yellow!30}370.02 & \cellcolor{yellow!30}0.91 & 0.22 & \cellcolor{yellow!30}418.87 & \cellcolor{yellow!30}0.89 & \cellcolor{yellow!30}0.18 \\
   \thickhline
   \end{tabular}%
   }
   \caption{
   \textbf{Summary of max(LR), $\hat{\beta}$, and $\texttt{SuffFailFreq}(0.98T)$ for the NS-MAB environment,} trained on the \protect\mybluehyperlink{gradual}{Gradual Variation} reward with horizon $T=25$, evaluated on various settings with different  linguistic contexts  ($LC$), time horizon ($T$), and action space size ($A$). \emph{LC=Gemini} means the linguistic context is generated by Gemini 2.0-Flash, \emph{LC=GPT} means the linguistic context is generated by GPT-4o mini. The trained model improves max regret and $\hat{\beta}$ in all three settings; for $\texttt{SuffFailFreq}$, it improves in one ($LC=GPT, T=100, A=5$), ties in one ($LC=Gemini, T=50, A=3$), and is slightly higher than the base model in one ($LC=GPT, T=100, A=3$).
   }
   \label{tab:nsmab}
   \end{table}

\paragraph{Improved Reasoning Rationales.} 
Lastly, we visualize representative reasoning rationales produced by the base and trained models of GPT-4o mini in \Cref{fig:reasoning-comparison}. The color-coded examples highlight how our training framework enhances both \textit{semantic–numerical alignment} and the \textit{E-E tradeoff}. It illustrates that our post-training can indeed enhance the reasoning rationales of the models in these examples, via both a better understanding of the numerical values and the online DM principle of properly balancing the  exploration and exploitation.

\section{Conclusion, Limitations, and Future Directions}

In this paper, we introduced \ours{}, a regret-driven post-training paradigm that turned self-generated trajectories into a supervised fine-tuning signal for training better decision-making LLM agents. Empirically, \briefalgorithmexpand{} consistently lowered the  regret values across three model classes: Transformers with numerical input/output, open-weight LLMs (Qwen3-8B and Gemma-2-9b-it) operating in language-grounded numerical tasks, and a closed-weight LLM (GPT-4o mini) fine-tuned in linguistically rich, real-world DM scenarios. Across these settings, we observed that the post-trained LLMs can achieve lower regret growth over time, a better exploration-exploitation tradeoff, and  generalizations to longer horizons, new reward distributions, and novel language descriptions. 

This study was constrained by several practical considerations. 
The training was limited to relatively short horizons (although the trained model, as an \emph{algorithm}, exhibited  good performance on longer horizons at inference time), due to 
API budget and context-length limits. Moreover,  our tasks remained mostly synthetic or procedurally generated, leaving the  performance evaluation on real human-in-the-loop applications as an important future work. Other future directions  include   
training \briefalgorithmexpand{} on genuinely long-horizon interactions, generalizing  
to other DM environments with richer structures such as the contextual bandits and Markov decision processes, augmenting the \briefalgorithmexpand{} trained models with agentic workflows and other inference-time interventions to further enhance their DM ability, and evaluating the trained LLM agents on more real-world DM applications such as tool-use, web browsing, and software engineering. Overall, we position our approach as an initial exploration, calling for more principled and novel post-training paradigms for LLM agents when it
comes to addressing decision-making tasks.

\section*{Acknowledgement}
 
The authors would like to thank Jisu Jang for the feedback on the Figures. C.P. acknowledges the
support from the Amazon AI PhD Fellowship and Korea Foundation for Advanced Studies Scholarship.  A.O. acknowledges this work was supported by the Department of the Navy, Office of Naval Research, under grant 036388-00002. Z.C. and K.Z. acknowledge the support from the Army Research Office (ARO) grant W911NF-24-1-0085, the NSF CAREER Award 2443704, a Coefficient Giving AI Safety Research Award, and the University of Maryland supercomputing resources (\href{http://hpcc.umd.edu}{http://hpcc.umd.edu}) made available for conducting the research reported in this paper. K.Z. additionally  acknowledges the support from the AFOSR YIP Award
FA9550-25-1-0258, a Cisco Faculty Research Award, and a JP Morgan Faculty Research Award.  

\bibliographystyle{ims}
\bibliography{main} 
\clearpage 

\newpage
\tableofcontents
\newpage
\appendix

\ifthenelse{\boolean{iclr}}
{

}{}
\section{Additional Notation}
We use $\mathbb{R}$ to denote the set of real numbers, $\RR^+$ to denote the set of non-negative real numbers, $\mathbb{N}$ for the set of non-negative integers, and $\mathbb{N}^+$ for the set of positive integers. For any finite set $\mathcal{S}$, let $\Delta(\mathcal{S})$ denote the probability simplex over $\mathcal{S}$. Given a positive integer $d \in \mathbb{N}^+$, we define the shorthand notation $[d] := \{1, 2, \dots, d\}$. For vectors $x, y \in \mathbb{R}^d$, we write $\langle x, y \rangle$ to denote their standard Euclidean inner product. Let $\pmb{0}_d$ and $\pmb{1}_d$ represent the $d$-dimensional all-zero and all-one vectors, respectively. We also let $\pmb{O}_{d \times d}$ and $I_{d \times d}$ denote the $d \times d$ zero matrix and identity matrix, respectively. When the dimension $d$ is clear from context, we omit the subscript. The vector $e_i$ denotes the $i$-th standard basis (unit) vector in $\mathbb{R}^d$. Given a vector $p \in \mathbb{R}^d$, a radius $R > 0$, and a norm $\norm{\cdot}$, the closed ball centered at $p$ with radius $R$ is denoted by $B(p, R, \norm{\cdot}) := \{x \in \mathbb{R}^d \mid \norm{x - p} \leq R\}$. For a convex set $C \subseteq \mathbb{R}^d$, we define the projection of $p$ onto $C$ under norm $\norm{\cdot}$ as $\texttt{Proj}_{C, \norm{\cdot}}(p) := \arg\min_{x \in C} \norm{x - p}$, which is well-defined due to the convexity of $C$. The softmax function is defined by $\texttt{Softmax}(x) := \left(\frac{e^{x_i}}{\sum_{j \in [d]} e^{x_j}}\right)_{i \in [d]}$. For a matrix $A \in \mathbb{R}^{m \times n}$ with columns $A_i\in\RR^m$, we define the Frobenius norm $\norm{A}_F:= \sqrt{\tr(A^\intercal A)}$. We use $A_{-1} := A_n$ to denote the last column of $A$. For any set $\Pi \subseteq \mathbb{R}^d$, we define its diameter under a norm $\norm{\cdot}$ as $\text{diam}(\Pi, \norm{\cdot}) := \sup_{\pi_1, \pi_2 \in \Pi} \norm{\pi_1 - \pi_2}$. The indicator function $\mathbbm{1}(\mathcal{E})$ equals  $1$ if the event $\mathcal{E}$ is true, and $0$ otherwise. For functions $f, g: \mathbb{R} \to \mathbb{R}$, we write $g(x) = \mathcal{O}(f(x))$ if there exist constants $x_0 \in \mathbb{R}$ and $M < \infty$ such that $|g(x)| \leq M |f(x)|$ for all $x > x_0$. 
For a sequence $(\ell_t)_{t \in [T]}$ with $T \in \mathbb{N}^+$, we define the sub-sequence $\ell_{a:b} := (\ell_a, \dots, \ell_b)$ for indices $1 \le a \le b \le T$. 
Let $\mathbb{O}(d)$ denote the orthogonal group of $d \times d$ real matrices, \textit{i.e.}, $\mathbb{O}(d) = \{ Q \in \mathbb{R}^{d \times d} \mid Q^\intercal Q = I_d \}$. We use $n!$ to denote the factorial of $n$, and $n!!$ to denote the double factorial, \textit{i.e.}, the product of all integers from $n$ down to 1 that have the same parity as $n$. The Gamma function $\Gamma(z)$ generalizes the factorial and is defined for complex $z$ with positive real part as $\Gamma(z) = \int_0^\infty t^{z-1} e^{-t} dt$.
\section{Deferred Preliminaries}
\ifthenelse{\boolean{iclr}}
{
\subsection{Decision-Making Environments}
\label{appendix:ssec_sdmenv}

\subsection{Performance Metric: Regret}
\label{appendix:ssec_regret}

}{}

\subsection{Kolmogorov–Smirnov Test for Regret Distributions}
\label{appendix:ssec:ks_explanation}
To statistically validate that the trained model achieves lower regret than the base model, we apply a one-sided two-sample Kolmogorov–Smirnov (KS) test. This test compares the distributions of final regrets obtained from repeated runs of the two models under identical scenarios. The one-sided version specifically checks whether the regret values from the trained model tend to be shifted toward smaller values relative to those from the base model. A lower $p$-value indicates stronger evidence against the null hypothesis—that the trained and base models have indistinguishable regret distributions—in favor of the alternative, namely that the trained model yields significantly lower regret.
\subsection{(Linear) Self-Attention}
\label{ssec:self-tf}
One key component in the Transformer architecture \citep{vaswani2017attention}, the backbone of modern LLMs, is the \emph{(self-)attention} mechanism. For simplicity, we here focus on introducing the \emph{single-layer} self-attention architecture.
The mechanism takes a sequence of  vectors \( Z = [z_1, \dots, z_t] \in \RR^{d \times t} \) as input, and  outputs some sequence of $[\hat{z}_1, \dots, \hat{z}_t] \in \RR^{d \times t}$. For each $i \in [t]$ where $i > 1$, the output is generated by  $\hat{z}_i = (Vz_{1:{i-1}}) \sigma((Kz_{1:i-1})^\intercal  (Qz_i))$, where $z_{1:{i-1}}$ denotes the $1$ to $i-1$ columns of $Z$, $\sigma$ is either the $\texttt{Softmax}$ or $\texttt{ReLU}$ activation function, and for the initial output, \( \hat{z}_1 = \pmb{0}_d \). Here, $V, Q, K \in \RR^{d \times d}$ are referred to as  the \emph{Value}, \emph{Query}, and \emph{Key} matrices,  respectively. Following the theoretical framework in \citet{von2023transformers, mahankali2023one}, we exclude the attention score for a token $z_i$ in relation to itself.
For theoretical analysis, we also consider the \emph{linear} self-attention  model, where $\hat{z}_i = (Vz_{1:{i-1}}) ((Kz_{1:i-1})^\intercal  (Qz_i))$.

\subsection{Metrics for Exploration Efficiency} 
\label{ssec:metricee}

To be qualified as a good decision-maker, the agent needs to balance \emph{exploitation} with \emph{exploration} in learning, especially in environments with only \emph{bandit} feedback \citep{szepesvari2022algorithms}. To evaluate the exploration efficiency of LLM agents, beyond regret, we adopt the metrics proposed by \citet{krishnamurthy2024can}: \texttt{SuffFailFreq$(t)$} and \texttt{MinFrac$(t)$}. 

Here, an \emph{experiment replicate} refers to a single independent run of the algorithm over $T$ rounds (\textit{e.g.}, with a different random seed). Then, \texttt{SuffFailFreq}$(t)$ measures the proportion of replicates in which the best action (\textit{i.e.}, the action with the highest expected reward) is never selected from round $t$ through $T$. Formally, for a replicate $R$, let $\texttt{SuffFail}(t, R) \in \{0, 1\}$ indicate whether the best action is missed entirely from round $t$ onward; then \texttt{SuffFailFreq}$(t)$ is the average of $\texttt{SuffFail}(t, R)$ across replicates. Lower values indicate better exploration efficiency. 

On the other hand, \texttt{MinFrac}$(t)$ captures how uniformly the available actions are explored. For each replicate and action $a \in \cA$, we compute $f_a(t, R)$, the fraction of times action $a$ is selected up to round $t$. \texttt{MinFrac}$(t)$ is then defined as the mean across replicates of the minimum $f_a(t, R)$ over all actions $a$. To normalize \texttt{MinFrac}$(t)$ to lie in $[0, 1]$, we plot $d \cdot \texttt{MinFrac}(t)$, where $d = |\cA|$ is the number of available actions. Higher values imply more balanced exploration across actions. Typically, a well-behaved no-regret algorithm exhibits a \texttt{MinFrac}$(t)$ pattern that increases during the early rounds—reflecting exploration—and gradually decreases in later rounds as the algorithm shifts toward exploitation. However, in the non-stationary setting, this trend becomes ill-defined since the underlying reward distribution evolves over round.

\subsection{Table of the Online Decision-Making Environments Considered}
\label{appendix:ssec:table_sdm}
\begin{table}[H]
\centering
\footnotesize
\renewcommand{\arraystretch}{1.5}
\begin{tabular}{@{}c c c c@{}}
\toprule
 & \textbf{FOL} & \textbf{MAB} & \textbf{NS-MAB} \\
\specialrule{1.2pt}{1pt}{1pt}
\textbf{\makecell{Defining \\ Components}} 
& $\cA$, $R_t$ 
& $\cA$, $r$ 
& $\cA$, $(r_t)_{t \in [T]}$, $V_T$  \\
\midrule
\textbf{Policy} 
& $\pi \in \Pi$ 
& $\pi \in \Delta(\cA)$ 
& $\pi \in \Delta(\cA)$ \\
\midrule
\textbf{Feedback} 
& Full 
& Bandit 
& Bandit \\
\midrule
\textbf{Regret Notion} 
& \makecell{v.s. Best Policy\\in Hindsight} 
& v.s. Best Arm 
& \makecell{v.s. Best Arm Sequence} \\
\midrule 
\textbf{\makecell{Assumptions on \\the Environment}} 
& Arbitrary{/Adversarial} 
& Stochastic 
& \makecell{Non-stationary\\$\sum_{t=2}^T \norm{r_t - r_{t-1}}_\infty \le V_T$} \\
\bottomrule
\end{tabular}
\caption{Summary of the notation for the language-grounded DM tasks studied in this paper:
\emph{FOL}, \emph{MABs}, and \emph{NS-MABs}. 
Here, $\Pi$ denotes a bounded policy class, $\cA$ is the \emph{action} set, and $V_T$ captures the total variation of the mean rewards in the non-stationary bandit setting. We may also use  $d=|\cA|$ to denote the number of actions.}

\label{tab:sequential_comparison}
\end{table}

\section{Baseline Algorithms}
\label{appendix_baseline}

Here we introduce the baseline algorithm used to compare with our method in several DM environments. All our baseline algorithms are iterative methods that updates the policy (or choose the action) based on the observed rewards, actions (or policies), and contexts (or states).
We apply all the baseline algorithms to normalized rewards
\[
    \bar R_t(a) = \frac{R_t(a)-R_{\min}}{R_{\max}-R_{\min}},
\]
where $[R_{\min},R_{\max}]$ denotes the known reward range of the environment (e.g., $[0,10]$ for the clipped reward processes in our experiments), while reported regret values are still computed on the original reward scale.
\paragraph{Follow-the-Leader (FTL).}

The FTL algorithm updates the policy that maximizes the sum of the past rewards. 
Mathematically, given a sequence of reward vectors $R_1,R_2,\cdots,R_t$, the FTL algorithm updates the policy $\pi$ at each round $t$ as follows: 

\begin{equation}
    \pi_{t+1} = \arg\max_{\pi \in \Pi} \sum_{\tau=1}^t \langle R_\tau, \pi \rangle.
\end{equation}

\paragraph{Follow-the-Regularized-Leader (FTRL).}
The FTRL algorithm \citep{shalev2007online} updates the policy based on observed data and a regularization term. The idea is to choose the next policy that maximizes the sum of the past rewards and a regularization term. 

Mathematically, given a sequence of reward vectors $R_1,R_2,\cdots,R_t$, the FTRL algorithm updates the policy $\pi$ at each round $t$ as follows: 

\begin{equation}
    \pi_{t+1} = \arg\max_{\pi \in \Pi} \left( \sum_{\tau=1}^t \langle R_\tau, \pi\rangle - \frac{1}{\eta}\cdot \texttt{Reg}(\pi) \right), \label{eqn:ftrl-def} 
\end{equation}
where $\texttt{Reg}(\pi)$ is a regularization term, and $\eta>0$ is the learning rate.

\paragraph{Hedge.}
The Hedge algorithm \citep{freund1997decision} (also known as the Multiplicative Weight Update algorithm \citep{arora2012multiplicative}) can also be derived from the FTRL algorithm, where $\Pi = \Delta(\cA)$ and the regularization term is the negative entropy $\texttt{Reg}(\pi) = \underset{a \in \cA}{\sum} \pi(a) \log \pi(a)$. Hence, at each round $t$, the policy is updated by:  
\begin{equation*}
    \pi_{t+1}(a) = \pi_{t}(a) \frac{\exp[\eta R_t(a)]}{\underset{a' \in \cA}{\sum}\pi_{t}(a') \exp[\eta R_t(a')]}. 
\end{equation*}

\paragraph{Upper Confidence Bound (UCB).}
The Upper Confidence Bound  algorithm \citep{auer2002finite} scores each arm by combining an empirical mean with an exploration bonus. For every arm $a$ we define
\begin{equation*}
    \text{UCB}_t(a) :=
    \begin{cases}
        +\infty, & N_t(a)=0, \\
        \hat R_t(a) + \sqrt{\frac{2\log t}{N_t(a)}}, & \text{otherwise},
    \end{cases}
\end{equation*}
where $\hat R_t(a) = \frac{\sum_{\tau=1}^t \mathbb{I}(a_\tau = a) R_\tau(a_\tau)}{N_t(a)}$ and $N_t(a)=\sum_{\tau=1}^t \mathbb{I}(a_\tau=a)$. At round $t$ the algorithm chooses $a_t = \arg\max_{a\in\cA} \text{UCB}_t(a)$, so rarely sampled arms automatically receive larger bonuses until their uncertainty diminishes.

\paragraph{EXP3.} 
The EXP3 algorithm \citep{auer2002nonstochastic} maintains a probability vector that mixes exploitation with uniform exploration. Let $\eta \in (0,1]$ be the learning rate. Initialise weights $w_1(a)=1$ for all $a \in \cA$. For each round $t=1,\dots,T$:
\begin{enumerate}[leftmargin=16pt]
    \item Form the sampling distribution
    \[
        p_t(a) = (1-\gamma)\frac{w_t(a)}{\sum_{a'} w_t(a')} + \frac{\gamma}{K},
    \]
    where $\gamma \in (0,1]$ is the exploration parameter.
    \item Sample arm $a_t \sim p_t$ and observe reward $R_t(a_t)$.
    \item Construct the unbiased estimate $\tilde{R}_t(a) = \frac{R_t(a_t)}{p_t(a_t)} \mathbb{I}\{a=a_t\}$.
    \item Update the weight $w_{t+1}(a) = w_t(a) \exp(\eta \tilde{R}_t(a))$ for all $a \in \cA$.
\end{enumerate}
With $\eta = \sqrt{\frac{2 \log K}{K T}}$ and $\gamma = \min\{1,\eta K/2\}$, EXP3 attains a regret bound $R_T = O\!\left(\sqrt{K T \log K}\right)$.
\paragraph{Rexp3.} 
For non-stationary stochastic bandits with a variation-budget $V_T$, we adopt the \textsc{Rexp3} algorithm from \citet{besbes2014stochastic}. The horizon is partitioned into batches of length
\[ \Delta_T = \Big\lceil \Big(\frac{K \log K}{V_T}\Big)^{1/3} T^{2/3} \Big\rceil, \]
and at the start of every batch we reset the EXP3 weights $w(a)=1$ for all $a \in \cA$. With exploration parameter $\gamma = \min\left\{1, \sqrt{\frac{K \log K}{(e-1)\Delta_T}} \right\}$, each round $t$ within the batch proceeds as follows:
\begin{enumerate}[leftmargin=16pt]
    \item Form the mixed distribution $p_t(a) = (1-\gamma) \frac{w_t(a)}{\sum_{a'} w_t(a')} + \frac{\gamma}{K}$.
    \item Sample arm $a_t \sim p_t$, observe reward $R_t(a_t)$, and define the importance-weighted estimate $\tilde{R}_t(a) = \frac{R_t(a_t)}{p_t(a_t)} \mathbb{I}\{a = a_t\}$.
    \item Update the weights via $w_{t+1}(a) = w_t(a) \exp\!\big( \frac{\gamma}{K} \tilde{R}_t(a) \big)$.
\end{enumerate}
After $\Delta_T$ rounds the weights are reset and the next batch begins.

\section{Remark for Meta Algorithm in \Cref{section:meta}}

Here, we purpose an addition remark for the Meta Algorithm purposed in \Cref{section:meta}.

\begin{remark}[{Optimal action labels in \briefalgorithmexpand{}}]\label{sec:remark_optimal_label} 
Similar to other supervised learning-based training/fine-tuning paradigms \citep{lee2023supervised,sinii2024context,beck2025tutorial}, optimal action labels are also needed in \briefalgorithmexpand{}  by the regret definition. However,  this is not an unrealistic requirement. First, note that such \emph{privileged information} is only needed for \emph{training} but not for \emph{testing/inference time}, and at inference time, new online DM tasks with novel reward generation processes will be addressed. Second, in non-adversarial and stochastic environments (\textit{i.e.}, MABs), such a requirement can be relaxed -- instead of comparing the $L$ trajectories in terms of regret, it suffices to directly compare their  \emph{accumulated reward}, as the optimal-action comparator remains \emph{identical} across the trajectories and can thus be omitted. Finally, for the adversarial but full-information environment we focused on (\textit{i.e.}, the FOL environment), since a \emph{fixed optimal}  action is not well-defined, computing such a \emph{best-in-hindsight} action is a natural benchmark  for performance evaluation  that may need to compute anyways. This can be obtained  by finding the best action for  each realized sequence of accumulated reward vectors. 
\end{remark}

\section{Reward Generation Processes}
\label{appendix:reward_dist}
We introduce a diverse set of reward generation processes used for training and evaluation throughout this paper. These reward generation processes generalize the reward generation processes of \citet{park2024llm}, which is sometimes restricted to the case of $d = 2$, enabling broader applicability across various FOL environments. Notably, when $d = 2$, the reward generation processes \mybluehyperlink{alternating}{VI}, \mybluehyperlink{noisy}{VII}, and \mybluehyperlink{adaptive}{VIII} reduce to the loss configuration studied in \citet{park2024llm}, rendering it a special case of the more general framework proposed here.

Throughout this appendix, we denote the observed reward vector at round $t$ by $R_t \in [0, 10]^d$. For both the MAB and NS-MAB environments, $r_t$ represents the mean reward vector of the underlying reward distribution at round $t$. The reward generation processes for the MAB environment are restricted to the  \mybluehyperlink{uniform}{Uniform}, \mybluehyperlink{gaussianmu}{Gaussian}, \mybluehyperlink{gamma}{Gamma}, and \mybluehyperlink{bernoulli}{Bernoulli} rewards, as these processes are stationary and stochastic. For the NS-MAB environment, we use the \mybluehyperlink{gradual}{Gradual Variation} reward.

\begin{enumerate}[label=\textcolor{blue}{\Roman*.}]
    \item \hypertarget{uniform}{\textbf{Uniform:}} For each $a \in \cA$, draw parameters $x_a, y_a \sim \mathrm{Unif}(0,10)$ and set reward independently at each round: 
    \[
    R_t(a) \sim \mathrm{Unif}([\min\{x_a,y_a\},\max\{x_a,y_a\}]).
    \]
    
    \item \hypertarget{gaussianmu}{\textbf{Gaussian:}} Each reward vector is drawn from a mixture of Gaussian distributions:  
    \[
    R_t \sim \tfrac{1}{3}\mathcal{N}(\mu, I_{d \times d}) + \tfrac{1}{3}\mathcal{N}(\mu, 3I_{d \times d}) + \tfrac{1}{3}\mathcal{N}(\mu, 10I_{d \times d}),
    \]  
    where $\mu \sim \mathcal{N}(\pmb{5}_d, I_{d \times d})$. The resulting vector is clipped element-wise to $[0,10]$.

    \item \hypertarget{gamma}{\textbf{Gamma:}} For each $a \in \cA$, sample shape and scale parameters $\alpha_a \sim \mathrm{Unif}(0,10)$ and $\theta_a \sim \mathrm{Unif}(0,2)$, then draw $R_t(a) \sim \mathrm{Gamma}([\alpha_a,\theta_a])$ independently across rounds. The resulting vector is clipped element-wise to $[0,10]$.
        
    \item \hypertarget{bernoulli}{\textbf{Bernoulli:}}
    Sample two values $x, y \sim \mathrm{Unif}(0,10)$ independently. 
    For each action $a \in \mathcal{A}$, sample a success probability $p_a \sim \mathrm{Unif}(0,1)$ independently.  
    At each round $t$, the mean reward for action $a$ is given by:
    \[
    R_t(a) = 
    \begin{cases}
    \max\{x, y\}, & \text{with probability } p_a \\
    \min\{x, y\}, & \text{with probability } 1 - p_a
    \end{cases}.
    \]
    \item \hypertarget{sine}{\textbf{Sine-trend:}} $R_t = 5(1 + \sin(xt + y))$, where $x, y \sim \mathrm{Unif}([0,10]^d)$.

    \item \hypertarget{alternating}{\textbf{Alternating:}} 
Sample a random shift $\tau \in \{0, 1, \dotsc, d-1\}$ uniformly at random. $R_t$ is defined as:
\[
R_t( a_{(t + \tau) \bmod d}) = 10,
\quad \text{and} \quad
R_t(a_s) = 0 \quad \text{for all } s \neq  (t + \tau) \bmod d.
\]

    \item \hypertarget{noisy}{\textbf{Noisy Alternating:}} Sample a random shift $\tau \in \{0, 1, \dotsc, d-1\}$ uniformly at random. $R_t$ is defined as follows:
\[
R_t(a_{(t + \tau) \bmod d}) = \min\left( \frac{25}{t+1}, 10 \right),
\quad \text{and} \quad
R_t(a_s) \sim \mathrm{Unif}([9,10]) \quad \text{for all } s \neq (t + \tau) \bmod d.
\]

    \item \hypertarget{adaptive}{\textbf{Adaptive:}} 
Let $\pi_t \in \Delta^{d-1}$ be the policy at round $t$ over these actions. $R_t$ is defined as follows:
\[
j^* = \arg\max_{j \in \{0, \dotsc, d-1\}} \pi_t(a_j),
\quad \text{then} \quad
R_t(a_j) = 
\begin{cases}
0 & \text{if } j = j^*, \\
10 & \text{otherwise}.
\end{cases}
\]

    \item \hypertarget{gradual}{\textbf{Gradual Variation (NS-MAB):}} 
    Sample an initial mean reward vector $r_1 = \mu \sim \mathrm{Unif}([0,10]^d)$. For each subsequent round $t \ge 1$, a perturbation vector $\Delta_t \sim \mathrm{Unif}([-1/\sqrt{t},1/\sqrt{t}]^d)$ is drawn independently, and the mean reward vector is updated as $r_{t+1} = r_t + \Delta_t$. Then, $R_t \sim \tfrac{1}{3}\mathcal{N}(r_t, I_{d \times d}) + \tfrac{1}{3}\mathcal{N}(r_t, 3I_{d \times d}) + \tfrac{1}{3}\mathcal{N}(r_t, 10I_{d \times d})$, and the resulting vector is clipped element-wise to $[0,10]$. Note that such a reward generation process yields a mean-reward variation rate of $V_T=\Theta(T^{1/2})$. 

\end{enumerate}

\ifthenelse{\boolean{iclr}}
{
\section{Meta Algorithm}
\begin{algorithm}[!h]
\caption{Meta Algorithm: \ours{}}\label{alg:ssft-meta-algorithm}
\begin{algorithmic}[1]
\State \textbf{Input:} A DM environment  (\textit{e.g.}, FOL, MAB, NS-MAB); %
an initial model ($\mathscr{A}$) for DM
\For{iteration $= 0, 1, 2, \dots$}
    \State $\cD = \emptyset$
    \For{scenario index $i = 1, 2, \dots, M$}
        \State Sample $L$ trajectories $C_1, \dots, C_L$ from the model under scenario$_i$ \label{line:L-traj-meta-alg}
        \State Compute the regret (\Cref{ssec:regret}) of each trajectory and select the $k$ trajectories with the lowest regret: $C_{(1), i}, \dots, C_{(k), i}$ 
        \State Update the training dataset: $\cD = \cD \cup \{\{\text{scenario}_i, C_{(1), i}, \dots, C_{(k),i}\}\}$
    \EndFor
    \State Fine-tune the model on the dataset $\cD$ via supervised fine-tuning \label{line:sft}
\EndFor
\end{algorithmic}
\end{algorithm}

\subsection{Summary of Instantiation of \Cref{alg:ssft-meta-algorithm}}

{
\begin{table}[ht]
\centering
\footnotesize
\renewcommand{\arraystretch}{1.5}
\begin{tabular}{@{}c c c c c@{}}
\toprule
 & \makecell[c]{\textbf{Transformers} \\ \textbf{with Numerical Input/Output} \\ (\Cref{sec:training-transformer})} & \makecell[c]{\textbf{Open-Weight} \\ \textbf{LLMs} \\ (\Cref{sec:open-weight-training})} & \makecell[c]{\textbf{Closed-Weight} \\ \textbf{LLMs} \\ (\Cref{sec:training-gpt})} \\
\specialrule{1.2pt}{1pt}{1pt}
\makecell[c]{\textbf{DM} \\ \textbf{Environments}} 
& FOL \& MAB 
& FOL \& MAB 
& FOL \& MAB \& NS-MAB \\
\midrule
\makecell[c]{\textbf{Category of} \\ \textbf{DM Tasks}} 
& Numerical DM 
& \makecell[c]{Language-Grounded \\  Numerical DM}
& \makecell[c]{Language-Grounded DM \\ with Real-world Contexts} \\
\midrule
\makecell[c]{\textbf{Output} \\ \textbf{with Reasoning}}
& N/A
& Yes
& Yes \\
\midrule
\textbf{Decision Output}
& $\Delta(\cA)$ or $B(\pmb{0}_d, R_{\Pi}, \norm{\cdot}_2)$
& \makecell[c]{Extract \makecell[c]{$\pi_t \in \Delta(\cA)$ for FOL \\$a_t \in \cA$ for MAB}\\ from the output}
&\makecell[c]{Extract $\pi_t \in \Delta(\cA)$ \\ from the output}\\
\midrule
\makecell[c]{\textbf{Sampling} \\ \textbf{Mechanism}} 
& \makecell[c]{Gaussian Noise \\ Perturbation}
& Stochastic Decoding 
& Stochastic Decoding \\
\bottomrule
\end{tabular}
\caption{Summary of {how our meta-algorithm is instantiated for different 
tasks studied in this paper:} 
Transformers with numerical input/output,  open-weight LLMs, and closed-weight LLMs.} 
\label{tab:modality}
\end{table}
}

}{}

\section{Training Details}
\label{app:exp-setup}

\subsection{Practical Computation of Regret in the MAB and NS-MAB Environments}
\label{regret-instantitaion}
During LLM training in the MAB and NS-MAB environments, the regret defined in \Cref{ssec:regret} involves taking an expectation over the algorithm’s stochasticity. For models employing the \textit{distributional} output—such as Transformers with numerical input/output or closed-weight LLMs—the probabilistic simplex itself serves as the algorithm’s output distribution, allowing direct computation of the regret in \Cref{ssec:regret}. In contrast, for open-weight LLMs that use the \textit{action-based} output, estimating this expectation would require sampling multiple trajectories. To circumvent this issue, we instead adopt the realized regret, defined as $\sum_{t=1}^T \max_{a \in \cA} r(a) - \sum_{t=1}^T R_t(a_{\mathscr{A},t})$, as the training signal, thereby avoiding the need for repeated sampling. 

\subsection{Significance Tests and Replicate Robustness}
\label{appendix:significance-tests}

In addition to the regression-based regret-growth validation in \Cref{ssec:regret}, one-sided KS tests on final regret distributions (see \Cref{appendix:ssec:ks_explanation}), and confidence intervals shown in the figures, we conduct 3-replicate experiments with per-setting error bars to assess robustness across independent runs. Individual settings remain underpowered for strong per-instance significance claims, but the replicate-level variability is small relative to the variability induced by task instances.

\begin{table}[h!]
\centering
\begin{tabular}{l|cc}
\thickhline
\textbf{Model} & \textbf{Mean Final Regret} & \textbf{Mean SD (Per-Setting)} \\ \hline
Base & 27.30 & 8.86 \\
Trained & 23.13 & 8.13 \\
\thickhline
\end{tabular}
\caption{
\textbf{Three-replicate robustness check for final regret.}
The per-setting standard deviation ($\approx 8$) is substantially below the task-level standard deviation ($\approx 25$), indicating that replicated experiments preserve the improvement trend despite limited power in individual settings.
}
\label{tab:replicate-robustness}
\end{table}

\subsection{Training Details for \Cref{ssec:experiment-results-sec4}}
We use a single-layer linear self-attention model to produce the results reported in \Cref{ssec:experiment-results-sec4}, corresponding to one of the simplest architectures. All models are trained for $1000$ iterations using the Adam optimizer with hyperparameters $\beta_1 = 0.9$, $\beta_2 = 0.999$, $\epsilon = 10^{-8}$, and a learning rate of $0.01$ for both the FOL and MAB environments. The batch size is set to 1000. Perturbations to the output of Transformers are added using a Gaussian noise: for the FOL environment, the noise follows $\cN(\pmb{0}_d, I_{d \times d})$, and for the MAB environment, it follows $\cN(\pmb{0}_d, 0.01\cdot  I_{d \times d})$. We sample $M = 100$ scenarios during each iteration and select $k = 1$ trajectory to the training dataset $\cD$ with the lowest regret from $L = 10$ trajectories per scenario.

\subsection{Training Details for \Cref{ssec:exp-results-openweight}}

All experiments use the AdamW optimizer with learning rate $5 \times 10^{-5}$ and batch size 4. Each model is trained on a single A100 GPU. We apply {high}-temperature sampling ($\tau = 1.0$) to promote diverse policy generation. We sample $M = 100$ scenarios during each iteration and select $k = 1$ trajectory to the training dataset $\cD$ with the lowest regret from $L = 10$ trajectories per scenario. For all training runs, we perform $3$ iterations of fine-tuning. Prompts are enhanced with numeric token highlighting, explicit constraints on policy validity, and formatting cues for short structured answers. 

\subsection{Parameter Sensitivity for Open-Weight LLM Training}
\label{appendix:param-sensitivity}

We conduct a parameter-sensitivity ablation on Qwen3-8B in the MAB environment, training with $T=25$ and evaluating with $T=100$. The ablation varies the number of sampled trajectories per instance $L$, the number of selected trajectories $k$, the number of fine-tuning iterations, and the sampling temperature $\tau$.

\begin{table}[h!]
\centering
\resizebox{0.7\textwidth}{!}{
\begin{tabular}{cccc|c}
\thickhline
\textbf{$L$} & \textbf{$k$} & \textbf{Iterations} & \textbf{Temperature} & \textbf{Final Regret} \\ \hline
\textit{default} & \textit{default} & \textit{default} & \textit{default} & $33.08 \pm 24.82$ \\
2 & 1 & 1 & 1.0 & $31.05 \pm 23.34$ \\
2 & 1 & 2 & 1.0 & $28.33 \pm 16.95$ \\
2 & 1 & 3 & 1.0 & $25.10 \pm 14.50$ \\
5 & 1 & 1 & 0.2 & $35.28 \pm 15.66$ \\
5 & 1 & 1 & 1.0 & $24.00 \pm 13.34$ \\
\thickhline
\end{tabular}
}
\caption{
\textbf{Parameter sensitivity for Qwen3-8B in the MAB environment.}
All runs train with $T=25$ and evaluate with $T=100$. The \textit{default} row denotes the default ablation configuration for this experiment, and should not be compared directly with base-model values from other evaluation protocols. Higher sampling temperature is important for preserving candidate diversity: with $L=5$, lowering $\tau$ from $1.0$ to $0.2$ increases final regret from $24.00$ to $35.28$. Increasing the number of iterations with $L=2$ and $\tau=1.0$ also improves final regret from $31.05$ to $28.33$ to $25.10$.
}
\label{tab:param-sensitivity-qwen}
\end{table}

These results suggest that candidate diversity and iterative refinement are important for \briefalgorithmexpand{}. Larger $L$ improves the quality of the selected low-regret trajectory, as seen by the decrease from $31.05$ at $L=2$ to $24.00$ at $L=5$ when $\tau=1.0$, but the benefit is expected to saturate as $L$ grows. Careful selection of $L$ and $\tau$ is therefore important in practice.

\subsection{Training Details for \Cref{ssec:sec6results}} \label{appendix:sec6-training-details}

For all training runs, we perform $5$ iterations of fine-tuning. In each iteration, we sample $L = 5$ trajectories per scenario across $M=200$ scenarios. For each scenario, we select $k=1$ trajectory with the lowest regret and include it in the training dataset $\cD$. We fine-tune the GPT-4o mini model (released on July 18, 2024) using the OpenAI API, with a learning rate multiplier set to 1.8. The training cost depends on token length and is approximately $ 0.5\times T^2$ dollars, where $T$ denotes the time horizon. This estimate includes both the cost of generating trajectories and the subsequent fine-tuning process.

\section{Prompts for Language-Grounded Numerical DM}
\label{appendix:prompt-open-weight}

\ifthenelse{\boolean{iclr}}
{}{}

\ifthenelse{\boolean{arxiv}}
{In \Cref{sec:open-weight-training}, we design a prompt for open-weight LLMs. Specifically, we distinguish two types of {\color{orange}interaction protocols}.}{}In this section, we will go deeper into how we convert numerical DM environments into language-grounded tasks using both types of {\color{orange}interaction protocols}. These protocols are broken down into individual steps, and we use distinct colors for each step to match the color-coded sentences in the examples below, illustrating how our prompts works for various settings. 

\begin{enumerate}
    \item[(a)] \fcolorbox{orange!30}{orange!30}{Summary-type Interaction}: At each turn $t+1$, the full interaction history is summarized into a single prompt, which includes either:  
(1) the sequence of past actions $(a_\tau)_{\tau \in [t]}$ and their corresponding scalar rewards $(R_\tau(a_\tau))_{\tau \in [t]}$, or   
(2) the sequence of past policies $(\pi_\tau)_{\tau \in [t]}$ and their corresponding reward vectors $(R_\tau)_{\tau \in [t]}$. 
    \item[(b)] \fcolorbox{orange!30}{orange!30}{Dialogue-type Interaction}: the interaction unfolds as a turn-by-turn dialogue. At each turn $t$, the model observes the full dialogue history up to that point and outputs a policy $\pi_t$ or an action $a_t$—either with or without accompanying reasoning. The task then provides feedback, in the form of either (1) a reward vector $R_t$, or (2) a sampled action $a_t$ if the model outputs a policy $\pi_t$ and its corresponding scalar reward $R_t(a_t)$. This process continues as an iterative sequence, with each turn incorporating the accumulated interaction history. 
\end{enumerate}

In prior studies \citet{park2024llm, krishnamurthy2024can, nie2024evolve}, DM interactions were typically of the \fcolorbox{orange!30}{orange!30}{\textit{summary-type}}, and often involved only aggregated statistics derived from the raw interaction history—for example, time-averaged rewards per action. In contrast, the \fcolorbox{orange!30}{orange!30}{\textit{dialogue-type}} interaction protocol adopted in our setting captures the interaction as a turn-by-turn conversational exchange, which more closely resembles how LLM agents are deployed in real-world DM tasks. This protocol preserves the full trajectory of past prompts and responses, supporting continuity across turns. When the model provides reasoning along with its policy, this reasoning becomes part of the dialogue context in future turns—\textbf{allowing the model to reflect on or revise earlier rationales}. Even in the absence of explicit reasoning, the model still conditions on its prior outputs and feedback, offering a flexible and context-sensitive framework for both learning and evaluation—an advantage not afforded by the summary-type setting. 

In the main body of the paper, we focus exclusively on \fcolorbox{orange!30}{orange!30}{\textit{dialogue-type}} interaction. Furthermore, \Cref{appendix:summary-results} presents an analysis of overfitting that may arise when applying \briefalgorithmexpand{} with \fcolorbox{orange!30}{orange!30}{\textit{summary-type}} interactions.

To convert numerical DM environments into language-grounded tasks, we apply the following prompting procedure: 

\begin{enumerate} 

\item[{\color{green!50!black} \textbf{Step 1}}] Provide the available action space $\cA$ the LLM will encounter. 

\item[{\color{blue} \textbf{Step 2}}] Specify that the model should return a policy in the policy space $\Pi = \Delta(\cA)$ or an action in the action space $\cA$. Give general instructions on the shape of valid policies (\textit{i.e.}, they must be distributions over actions). 

\item[{\color{purple} \textbf{Step 3}}] Show how the reward is returned, especially on how rewards are calculated (\textit{i.e.}, inner product $\langle \pi, r \rangle$). 

\item[{\color{red} \textbf{Step 4}}] Add setting-specific instructions, depending on the online DM environment. 

\end{enumerate}

We consider various settings for different steps of our prompting procedure: {\color{blue} \textbf{Step 2}}, {\color{purple} \textbf{Step 3}}, and {\color{red} \textbf{Step 4}}. 

In {\color{blue} \textbf{Step 2}}, we consider two different {\color{blue}output types}: 

\begin{enumerate}
    \item[(a)] \hyperlink{action-MAB-no-reasoning}{\color{black}\fcolorbox{blue!30}{blue!30}{Action-based output}}: the agent will be required to decide a specific action in the action space $\cA$. {Since the \textbf{\textit{full-information} feedback} requires a probability simplex over $\cA$, we extract the top-5 token probabilities at the action output position and re-normalize the probabilities over the subset of tokens corresponding to feasible actions. }
    \item[(b)] \hyperlink{distribution-FOL-reasoning}{\color{black}\fcolorbox{blue!30}{blue!30}{Distributional output}}: the agent will decide a policy from the policy space $\Pi = \Delta(\cA)$ and return. The task will also provide an instruction on what the policy should look like (\textit{i.e.}, a simplex). For the \textbf{\textit{bandit} feedback}, the agent will be informed that the environment will sample an action from the policy it provides.
\end{enumerate}

For open-weight LLMs , we adopt action-based outputs, whereas for the closed-weight LLM, we employ distributional outputs. A separate subsection analyzing output types is provided in \Cref{appendix:distributional-results}.

In {\color{purple} \textbf{Step 3}}, we consider two different types of feedback:

\begin{enumerate}
    \item[(a)] \hyperlink{distribution-FOL-reasoning}{\color{black}\fcolorbox{purple!30}{purple!30}{Full-information feedback}}: the task will reveal the full-information feedback $r$, and the task also informs the agent that the policy will be evaluated by inner product $\langle \pi, r \rangle$.
    \item[(b)] \hyperlink{action-MAB-no-reasoning}{\color{black}\fcolorbox{purple!30}{purple!30}{Bandit feedback}}: the task informs the agent that only the reward of the realized action will be revealed as feedback.
\end{enumerate}

In {\color{red} \textbf{Step 4}}, we vary the prompt to elicit different {\color{red}output formats}, introducing two distinct types of interactions:

\begin{enumerate}
    \item[(a)] \hyperlink{action-MAB-no-reasoning}{ \color{black}\fcolorbox{red!30}{red!30}{Policy-only format}}: the agent responds with a concise answer that includes only the policy, optionally omitting any explanation.
    \item[(b)] \hyperlink{distribution-FOL-reasoning}{\color{black}\fcolorbox{red!30}{red!30}{Policy-with-reasoning format}}: the agent provides a policy along with its reasoning process for outputting the policy. 
\end{enumerate}

Now we provide two examples of prompts for \fcolorbox{orange!30}{orange!30}{dialogue-type interaction} with different settings. The first is using \fcolorbox{blue!30}{blue!30}{distributional output} with \fcolorbox{purple!30}{purple!30}{full-information feedback} in \fcolorbox{red!30}{red!30}{policy-with-reasoning format}, while the second is using \fcolorbox{blue!30}{blue!30}{action-based output} with \fcolorbox{purple!30}{purple!30}{bandit feedback} in \fcolorbox{red!30}{red!30}{policy-only format}. The colored sentences in the example match the colored steps of the prompting procedure. It is worth noting that the prompt may be slightly adjusted across different open-weight LLMs to encourage the model to produce a roughly consistent number of tokens per round, ensuring that reasoning rationales are neither excessively long nor too brief. 

\begin{tcolorbox}[title=\hypertarget{distribution-FOL-reasoning}{Example Prompt Using \textbf{\textit{Distributional} Output} with \textbf{\textit{Full-Information} Feedback} in \textbf{\textit{Policy-with-Reasoning} Format}}, colback=white!95!gray, colframe=black!75, sharp corners=south, breakable]
\begin{dialogue}
\speak{Environment} 

\textcolor{green!50!black}{You are solving a decision-making problem for multiple rounds. There are 3 number of action (which is 0 to 2).}

\textcolor{blue}{At each round, you need to choose a policy. The policy specifies your probability of choosing each action. \\This policy should be 3-dimensional, and the sum of its components should equal 1. After that, you will be shown the reward vector for choosing each action.}

\textcolor{purple}{Remember that this reward vector is decided by the external system and can be potentially different for different rounds. \\It is not decided by what policies you have chosen. The reward vector is also 3-dimensional. \\It represents the reward of choosing action from 1 to 3. \\For example, a reward vector of [0.8, 3.2, 1.2] means the reward for action 1 is 0.8 and the reward for action 2 is 3.2 and the reward for action 3 is 1.2. \\Then your reward for this round will be calculated according to the reward of each action and your probability of choosing each action.
For example, if you choose the policy [0.2, 0.6, 0.2] and get the reward vector [0.8, 3.2, 1.2], then your expected reward is 0.2*0.8 + 0.6*3.2 + 0.2*1.2 =2.32.}

Your goal is to maximize your cumulative expected reward. You can adjust your policy based on the reward vectors for all previous rounds. 

\textcolor{red}{You're required to provide your policy with reasoning. Please explain briefly why you chose the policy based on the reward vectors from previous rounds, }

and provide your policy in numerical format in the last line. The last line of your reply should be in the format: "Policy: [your 3-dimensional policy]. " 

You are now on round 1 of the decision-making problem. Now, please provide \textcolor{red}{your reasoning and new policy.} The last line of your reply should be in the format: "Policy: [your 3-dimensional policy]. \\

\speak{Agent} \textcolor{red}{This is an exciting challenge! \\Since this is the first round, I don't have any past reward vectors to inform my decision. As a starting point, I'll adopt a uniform policy, assuming the rewards are initially unknown and evenly distributed.} 

\textcolor{blue}{Policy: [0.33, 0.33, 0.33]. \\}

\speak{Environment} You are now on round 2 of the decision-making problem. \textcolor{purple}{For the last round, you've received the following rewards: [7.64, 7.22, 1.56].} Now, please provide \textcolor{red}{your reasoning and new policy.} The last line of your reply should be in the format: "Policy: [your 3-dimensional policy]. \\ 

$\cdots\cdots$

\end{dialogue}
\end{tcolorbox}

\begin{tcolorbox}[title=\hypertarget{action-MAB-no-reasoning}{Example Prompt Using \textbf{\textit{Action-Based} Output} with \textbf{\textit{Bandit} Feedback} in \textbf{\textit{Policy-Only} Format}}, colback=white!95!gray, colframe=black!75, sharp corners=south, breakable]
\begin{dialogue}
\speak{Environment} \textcolor{green!50!black}{You are solving a decision-making problem for multiple rounds. There are 3 number of action (which is 1 to 3). }

\textcolor{blue}{At each round, you need to choose an action. After that, you will be shown the reward for choosing this action. }

\textcolor{purple}{Remember that this reward is decided by the external system and can be potentially different for different rounds, even if you choose the same action. \\Because you cannot see the reward corresponding to other actions that you did not choose, balancing exploration and exploitation is crucial in this decision-making process. }

Your goal is to maximize your cumulative reward. You can adjust your action based on the rewards you received for all previous rounds. 

You're required to only provide your action between 1 and 3. Your reply should be a single line: "Action: 'the number of your action'." Nothing else should be included. 

You are now on round 1 of the decision-making problem. Now, please provide your new action between 1 and 3. Your reply should be a single line: "Action: 'the number of your action'." Nothing else should be included. \\

\speak{Agent} \textcolor{blue}{Action: 1. }\\

\speak{Environment} You are now on round 2 of the decision-making problem. \textcolor{purple}{For the last round, you chose action 1, and you've received the following reward: 8.29.} Now, please provide your new action between 1 and 3. Your reply should be a single line: "Action: 'the number of your action'." Nothing else should be included.   \\ 

$\cdots\cdots$

\end{dialogue}
\end{tcolorbox}

\ifthenelse{\boolean{iclr}
}{
\section{Deferred Dataset Generation Details for \Cref{ssec:setting-sec6}}
\label{appendix:ssec:datagen}

\ifthenelse{\boolean{iclr}}{\subsection{Full-Information and Bandit Tasks}}{}
\label{sssec:ol-gen}
\paragraph{{Linguistic Context} Generation.}
We adopt a unified procedure to construct linguistic contexts for both the full-information online learning (FOL) and (non-)stationary multi-armed bandit (MAB/NS-MAB) environments. Using the GPT-4o mini API, we synthesize scenario descriptions that specify the common action set shared by the environments. Manually designed prompt templates span domains such as healthcare, business strategy, and science fiction, and they explicitly instruct GPT-4o mini to describe multiple plausible actions so that no candidate is obviously optimal. We further randomize the narrative style—ranging from concise reports to analytical dilemmas—to encourage linguistic and structural diversity. These prompts apply verbatim across FOL and MAB/NS-MAB tasks; the only difference is how the generated scenarios are paired with subsequent reward signals. Complete prompt templates appear in \Cref{appendix:prompt_scene_generation}, and we validate that this richer prompt induces substantially more diverse linguistic contexts than a simple prompt baseline in \Cref{appendix:prompt-diversity-ablation}. 

\paragraph{Reward Generation Processes.}
To keep the linguistic description decoupled from the numerical feedback, we instantiate rewards independently of the generated narratives, drawing from the distributions in \Cref{appendix:reward_dist}. This separation discourages the model from exploiting the superficial textual cues when predicting rewards.

}{}
\section{Prompts for Language-Grounded DM with Real-World Contexts}
\label{appendix:prompt_scene_generation}

In \Cref{sec:training-gpt}, we use the following prompts to generate the linguistic context for the MAB environment. 

\begin{tcolorbox}[title=Prompt for the Multi-Armed Bandit Linguistic Context Generation, colback=white!95!gray, colframe=black!75, sharp corners=south, breakable]

\begin{lstlisting}[breaklines=true,literate={\#}{{\char`\#}}1, columns=flexible, keepspaces=True]
**Task:**  

Generate `{batch_size}` distinct multi-armed bandit scenarios, ensuring diversity across domains, decision types, and levels of uncertainty. Each scenario should describe a realistic decision-making situation with `{action_num}` different choices/actions. The scenarios must follow these guidelines:  

### **1. Domain Diversity**  
- Each scenario should belong to a different domain, such as:
- **Business & Marketing** (e.g., advertising strategies, pricing models)  
- **Healthcare & Medicine** (e.g., treatment options, shift schedules)  
- **Education & Learning** (e.g., online learning methods, test strategies)  
- **Technology & AI** (e.g., recommendation algorithms, search ranking models)  
- **Entertainment & Gaming** (e.g., game design choices, streaming content selection)  
- **Public Policy & Governance** (e.g., tax incentives, urban planning)  
- **Sci-Fi & Futuristic** (e.g., AI-powered cities, space exploration decisions)  

### **2. Action Fairness & Trade-offs**  
- Ensure that **no action is obviously better** than the others.  
- Each choice should have **both benefits and drawbacks** to encourage exploration.  
- Avoid biased phrasing that suggests one option is superior.  

### **3. Decision Types & Framing**  
- Use **different decision-making structures**, such as:
- **Exploration vs. Exploitation** (e.g., investing in a new vs. proven strategy)  
- **Short-term vs. Long-term Trade-offs** (e.g., immediate profits vs. sustainable growth)  
- **Human vs. Algorithmic Decision-Making** (e.g., manual vs. automated hiring)  
- **Competitive vs. Cooperative Settings** (e.g., market competition vs. collaboration)  

### **4. Writing Variety**  
- Avoid repetitive phrasing; vary sentence structures and tones.  
- Use a mix of:
- **Problem statements** (e.g., "A company must decide...")  
- **Open-ended framing** (e.g., "What is the best way to...")  
- **Narrative-driven approaches** (e.g., "A researcher is testing different...")  

### **5. Output Format (JSONL)**  
Each scenario should be outputted in the following format:  
{{"scenario": "A hospital is testing different shift schedules for nurses to improve patient care and reduce burnout. There are four scheduling options: traditional 8-hour shifts, 12-hour shifts, rotating shifts, and flexible self-scheduled shifts. Each system has been successful in different hospitals, but it's unclear which will work best in this particular setting.", "action": ["8-hour shifts", "12-hour shifts", "rotating shifts", "flexible self-scheduled shifts"]}}
\end{lstlisting}

\end{tcolorbox}

\subsection{Prompt Diversity Ablation}
\label{appendix:prompt-diversity-ablation}

To validate that the prompt used in the paper induces diverse linguistic contexts, we compare it against a simpler earlier prompt, \texttt{generate bandit scenarios}, using $N=200$ generated scenarios for each prompt. The rich prompt is the scenario-generation prompt used in our experiments.

\begin{table}[h!]
\centering
\resizebox{\textwidth}{!}{
\begin{tabular}{lcc}
\thickhline
\textbf{Metric} & \textbf{Rich Prompt} & \textbf{Simple Prompt} \\ \hline
Vocabulary Size (higher is more diverse) & 1,481 & 634 \\
Yule's $K$ (lower is more diverse) & 83.70 & 331.64 \\
MSTTR (higher is more diverse) & 0.7771 & 0.4936 \\
Unique Jargon Terms (higher is more diverse) & 49 & 1 \\
\thickhline
\end{tabular}
}
\caption{
\textbf{Prompt diversity ablation for linguistic context generation.}
The rich prompt used in our experiments produces substantially greater linguistic diversity than the simple prompt baseline: it has roughly $4\times$ lower Yule's $K$, 57\% higher MSTTR, and 49 unique jargon terms across more than eight domains, compared with near-zero jargon diversity for the simple prompt.
}
\label{tab:prompt-diversity-ablation}
\end{table}

These results confirm that the rich scenario-generation prompt drives substantially greater linguistic diversity. Since the interaction protocol in our experiments is multi-turn by construction, chain-of-thought reasoning and few-shot-like context accumulation are structural properties of each generated instance rather than separate configurations to ablate.

\section{Deferred Experimental Results for \Cref{sec:training-transformer}}
\label{appendix:sec4results}

\ifthenelse{\boolean{iclr}}
{
\subsection{Remarks on \Cref{ssec:setting-sec4}}
\label{appendix:sec:defexplanation_tf}
\paragraph{Task comparison with \Cref{sec:open-weight-training} and \Cref{sec:training-gpt}.}
\cp{need to check tasks vs setup.} This setup differs from the ones described in \Cref{sec:open-weight-training} and \Cref{sec:training-gpt}, where both the input and output use language descriptions. In those settings, the model generates the policy autoregressively through a sequence of language tokens. In contrast, in the numerical setting studied here, the Transformer is applied once per round to directly predict the policy. 
\paragraph{Input Format for the Single-Layer Transformer}

The attention {scores} are computed over the reward history, and the resulting vector is then passed through an {\texttt{Operator}}.

}{}

\subsection{Full-Information Online Learning}
\label{appendix:ssec:regret_tranformer}

\ifthenelse{\boolean{iclr}}{}{}

\subsubsection{Complete Results for Full-Information Online Learning}
\label{appendix:sssec:full_result_sec4}
\begin{figure}[!h]
    \centering
    \includegraphics[width=\linewidth]{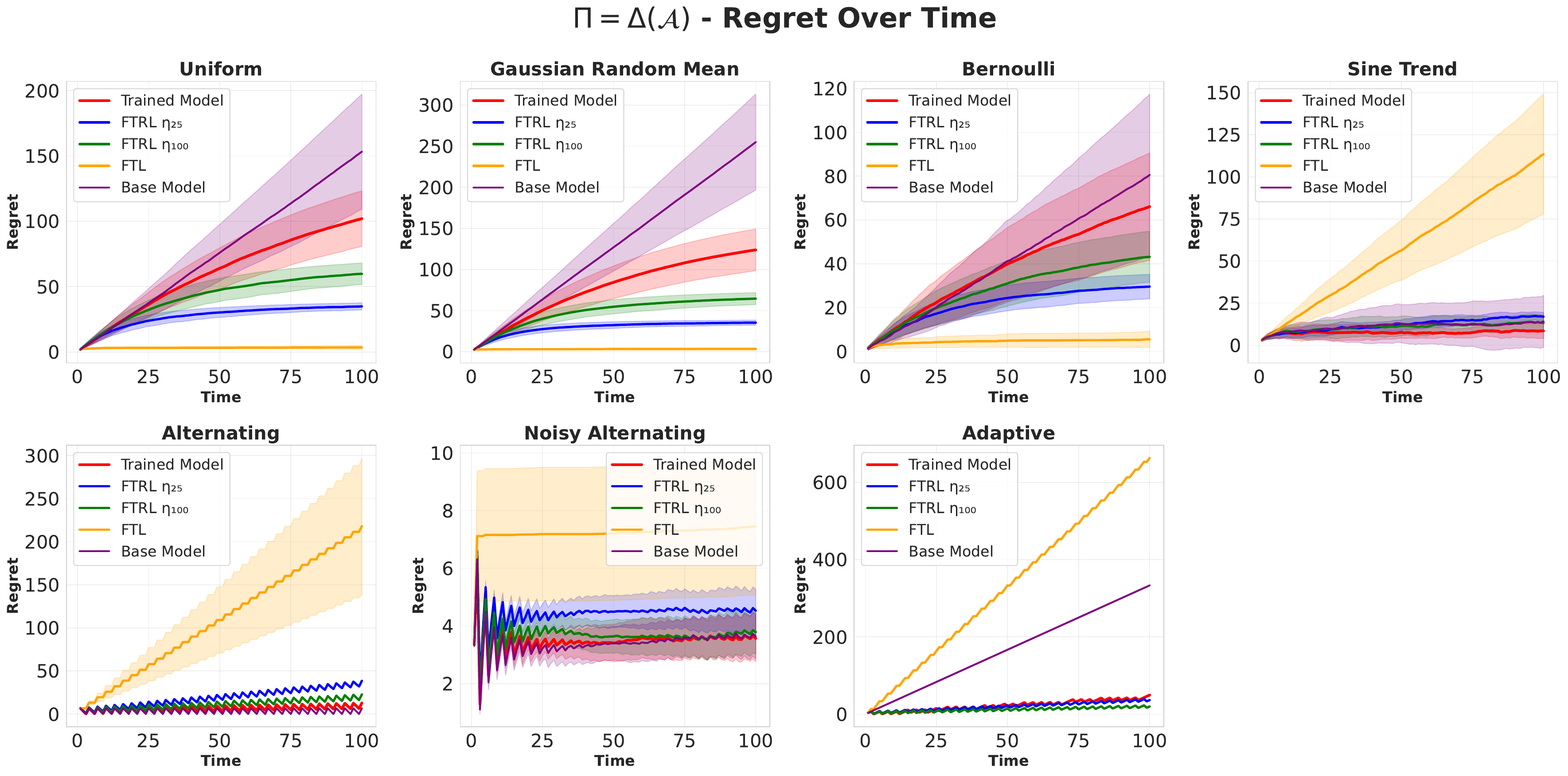}
    \caption{\textbf{The regret over time of the Transformers with numerical input/output for the FOL environment under \texttt{Horizon Generalization} [$T=25 \rightarrow T=100$] with $\Pi = \Delta(\mathcal{A})$.} The model is trained on the \protect\mybluehyperlink{gaussianmu}{Gaussian} reward. The evaluation consider multiple reward generation processes, including the training reward generation process and \texttt{Reward Generalization}[\protect\mybluehyperlink{gaussianmu}{Gaussian} $\rightarrow$ \protect\mybluehyperlink{uniform}{Uniform}, \protect\mybluehyperlink{bernoulli}{Bernoulli}, \protect\mybluehyperlink{sine}{Sine-trend}, \protect\mybluehyperlink{alternating}{Alternating}, \protect\mybluehyperlink{noisy}{Noisy Alternating}, \protect\mybluehyperlink{adaptive}{Adaptive}]. We compare the performance of our trained model (red) against classical baselines: FTRL with stepsize $\eta_{25}$ (blue) and $\eta_{100}$ (green), and FTL (orange). The shaded regions represent the standard deviation over 100 random instances. The results demonstrate that our model consistently achieves sublinear regret across diverse reward generation processes with $T=100$ despite being trained only for the \protect\mybluehyperlink{gaussianmu}{Gaussian} reward with $T=25$. }
    \label{fig:simplex_tf_all}
\end{figure}
\clearpage 
\begin{figure}[!h]
    \centering
    \includegraphics[width=\linewidth]{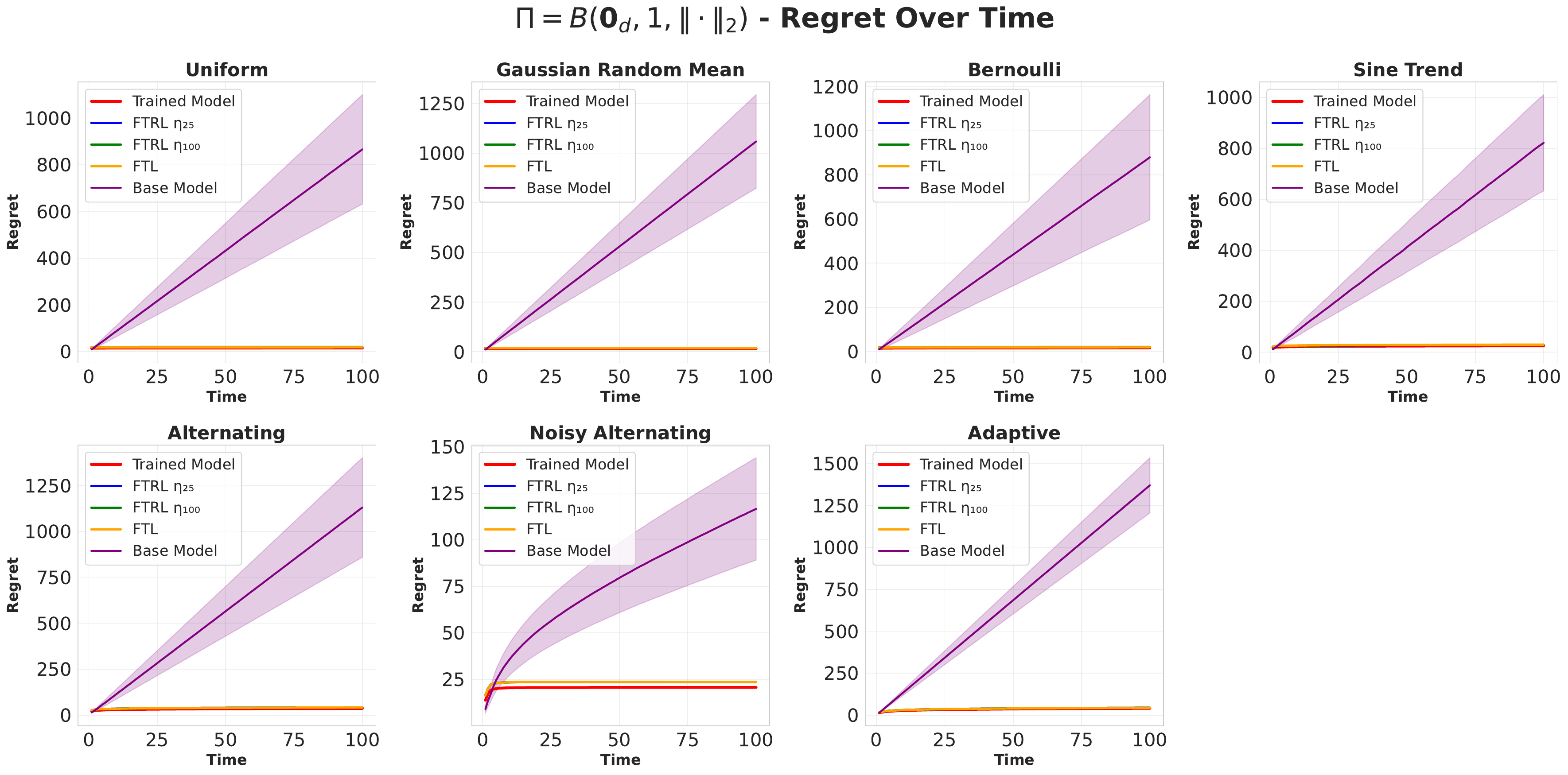}
\caption{\textbf{The regret over time of the Transformers with numerical input/output for the FOL environment under \texttt{Horizon Generalization}[$T=25 \rightarrow T=100$] with $\Pi = B(\pmb{0}_d, 1, \norm{\cdot}_2)$.} The model is trained on the \protect\mybluehyperlink{gaussianmu}{Gaussian} reward, and the evaluation considers multiple reward generation processes, including the training reward generation process and multiple kinds of \texttt{Reward Generalization}[\protect\mybluehyperlink{gaussianmu}{Gaussian} $\rightarrow$ \protect\mybluehyperlink{uniform}{Uniform}, \protect\mybluehyperlink{bernoulli}{Bernoulli}, \protect\mybluehyperlink{sine}{Sine-trend}, \protect\mybluehyperlink{alternating}{Alternating}, \protect\mybluehyperlink{noisy}{Noisy Alternating}, \protect\mybluehyperlink{adaptive}{Adaptive}]. We compare the performance of our trained model (red) against classical baselines: FTRL and FTL. The shaded areas indicate the standard deviation across 100 random instances. 
}
\label{fig:ball_tf_all}
\end{figure}

We report both $p_{\text{trend}}$ and $\hat{\beta}$ for each reward generation process to assess the generalization capability of our Transformer-based agent in \Cref{tab:simplex_table} and \ref{tab:l2ball_table}. Across all reward generation processes, we observe that $\hat{\beta} < 1$ with consistently low $p_{\text{trend}}$, indicating that the cumulative regret grows sublinearly with high statistical confidence. These findings are further supported by the regret over time presented in the accompanying figures. Notably, this generalization holds despite the model being trained under a fixed time horizon ($T = 25$) and a single distributional family—namely, the \mybluehyperlink{gaussianmu}{Gaussian} reward. These results suggest that our Transformer is capable of extrapolating to unseen temporal scales and diverse reward structures, highlighting its robustness in the FOL environment in both $\Pi = \Delta(\cA)$ and $\Pi = B(\pmb{0}_d, 1, \norm{\cdot}_2)$. We also provide the regret over time for each reward generation processes that we evaluate in \Cref{fig:simplex_tf_all} and \Cref{fig:ball_tf_all}. 

\begin{table*}[t]
  \begin{minipage}[t]{0.47\textwidth}
    \centering
    \caption{
    \textbf{Summary of $p_{\text{reg}}$ and fitted regret exponent ($\hat{\beta}$) with \texttt{Horizon Generalization}[$T=25 \rightarrow T=100$] for the FOL environment} with $\Pi = \Delta(\mathcal{A})$. We provide results on both the training distribution and multiple \texttt{Reward Generalization}. 
    }
    \label{tab:simplex_table}
    \begin{tabular}{llrr}
\toprule
Reward & Algorithm & $\hat{\beta}$ & $p_\text{reg}$ \\
\midrule
\multirow{4}{*}{\mybluehyperlink{adaptive}{Adaptive}} & Trained & 0.430 & $<$0.001 \\
& FTRL  $\eta_{25}$ & 0.332 & $<$0.001 \\
& FTRL  $\eta_{100}$ & 0.165 & $<$0.001 \\
& FTL  & 6.665 & $<$0.001 \\
\midrule
\multirow{4}{*}{\mybluehyperlink{alternating}{Alternating}} & Trained & 0.062 & $<$0.001 \\
& FTRL  $\eta_{25}$ & 0.322 & $<$0.001 \\
& FTRL  $\eta_{100}$ & 0.163 & $<$0.001 \\
& FTL  & 2.300 & $<$0.001 \\
\midrule
\multirow{4}{*}{\mybluehyperlink{bernoulli}{Bernoulli}} & Trained & 0.580 & $<$0.001 \\
& FTRL  $\eta_{25}$ & 0.223 & $<$0.001 \\
& FTRL  $\eta_{100}$ & 0.410 & $<$0.001 \\
& FTL  & 0.026 & $<$0.001 \\
\midrule
\multirow{4}{*}{\mybluehyperlink{gaussianmu}{Gaussian}} & Trained & 0.665 & $<$0.001 \\
& FTRL  $\eta_{25}$ & 0.212 & $<$0.001 \\
& FTRL  $\eta_{100}$ & 0.495 & $<$0.001 \\
& FTL  & 0.003 & $<$0.001 \\
\midrule
\multirow{4}{*}{\mybluehyperlink{noisy}{Noisy Alternating}} & Trained & 0.003 & 0.051 \\
& FTRL  $\eta_{25}$ & 0.003 & 0.042 \\
& FTRL  $\eta_{100}$ & 0.001 & 0.659 \\
& FTL  & 0.007 & $<$0.001 \\
\midrule
\multirow{4}{*}{\mybluehyperlink{sine}{Sine-trend}} & Trained & 0.085 & $<$0.001 \\
& FTRL  $\eta_{25}$ & 0.123 & $<$0.001 \\
& FTRL  $\eta_{100}$ & 0.076 & $<$0.001 \\
& FTL  & 1.092 & $<$0.001 \\
\midrule
\multirow{4}{*}{\mybluehyperlink{uniform}{Uniform}} & Trained & 0.995 & $<$0.001 \\
& FTRL  $\eta_{25}$ & 0.247 & $<$0.001 \\
& FTRL  $\eta_{100}$ & 0.514 & $<$0.001 \\
& FTL  & 0.006 & $<$0.001 \\
\bottomrule
\end{tabular}
  \end{minipage}
  \hspace{0.06\textwidth} %
  \begin{minipage}[t]{0.47\textwidth}
    \centering
    \caption{
    \textbf{Summary of $p_{\text{reg}}$ and $\hat{\beta}$ with \texttt{Horizon Generalization}[$T=25 \rightarrow T=100$] for the FOL environment} with $\Pi = B(\pmb{0}_d, 1, \norm{\cdot}_2)$. We provide results on both the training distribution and multiple \texttt{Reward Generalization}. 
    }
    \label{tab:l2ball_table}
    \begin{tabular}{llrr}
\toprule
Reward & Algorithm & $\hat{\beta}$ & $p_\text{reg}$ \\
\midrule
\multirow{4}{*}{\mybluehyperlink{adaptive}{Adaptive}} & Trained & 0.166 & $<$0.001 \\
& FTRL  $\eta_{25}$ & 0.170 & $<$0.001 \\
& FTRL  $\eta_{100}$ & 0.170 & $<$0.001 \\
& FTL & 0.170 & $<$0.001 \\
\midrule
\multirow{4}{*}{\mybluehyperlink{alternating}{Alternating}} & Trained & 0.084 & $<$0.001 \\
& FTRL  $\eta_{25}$ & 0.086 & $<$0.001 \\
& FTRL  $\eta_{100}$ & 0.086 & $<$0.001 \\
& FTL & 0.086 & $<$0.001 \\
\midrule
\multirow{4}{*}{\mybluehyperlink{bernoulli}{Bernoulli}} & Trained & 0.010 & $<$0.001 \\
& FTRL  $\eta_{25}$ & 0.010 & $<$0.001 \\
& FTRL  $\eta_{100}$ & 0.010 & $<$0.001 \\
& FTL & 0.013 & $<$0.001 \\
\midrule
\multirow{4}{*}{\mybluehyperlink{gaussianmu}{Gaussian}} & Trained & 0.007 & $<$0.001 \\
& FTRL  $\eta_{25}$ & 0.003 & $<$0.001 \\
& FTRL  $\eta_{100}$ & 0.003 & $<$0.001 \\
& FTL & 0.003 & $<$0.001 \\
\midrule
\multirow{4}{*}{\mybluehyperlink{noisy}{Noisy Alternating}} & Trained & 0.009 & $<$0.001 \\
& FTRL  $\eta_{25}$ & 0.009 & 0.001 \\
& FTRL  $\eta_{100}$ & 0.010 & 0.001 \\
& FTL & 0.009 & 0.001 \\
\midrule
\multirow{4}{*}{\mybluehyperlink{sine}{Sine-trend}} & Trained & 0.041 & $<$0.001 \\
& FTRL  $\eta_{25}$ & 0.042 & $<$0.001 \\
& FTRL  $\eta_{100}$ & 0.042 & $<$0.001 \\
& FTL & 0.042 & $<$0.001 \\
\midrule
\multirow{4}{*}{\mybluehyperlink{uniform}{Uniform}} & Trained & 0.006 & $<$0.001 \\
& FTRL  $\eta_{25}$ & 0.004 & $<$0.001 \\
& FTRL  $\eta_{100}$ & 0.005 & $<$0.001 \\
& FTL & 0.005 & $<$0.001 \\
\bottomrule
\end{tabular}
  \end{minipage}
\end{table*}

\clearpage

\subsection{Multi-Armed Bandits}
\label{appendix:ssec-mab-results-tf}
Here, we report the regret over time and the final regret distribution for the \mybluehyperlink{gamma}{Gamma} reward.
\begin{figure}[!h]
    \centering
    \includegraphics[width=0.9\linewidth]{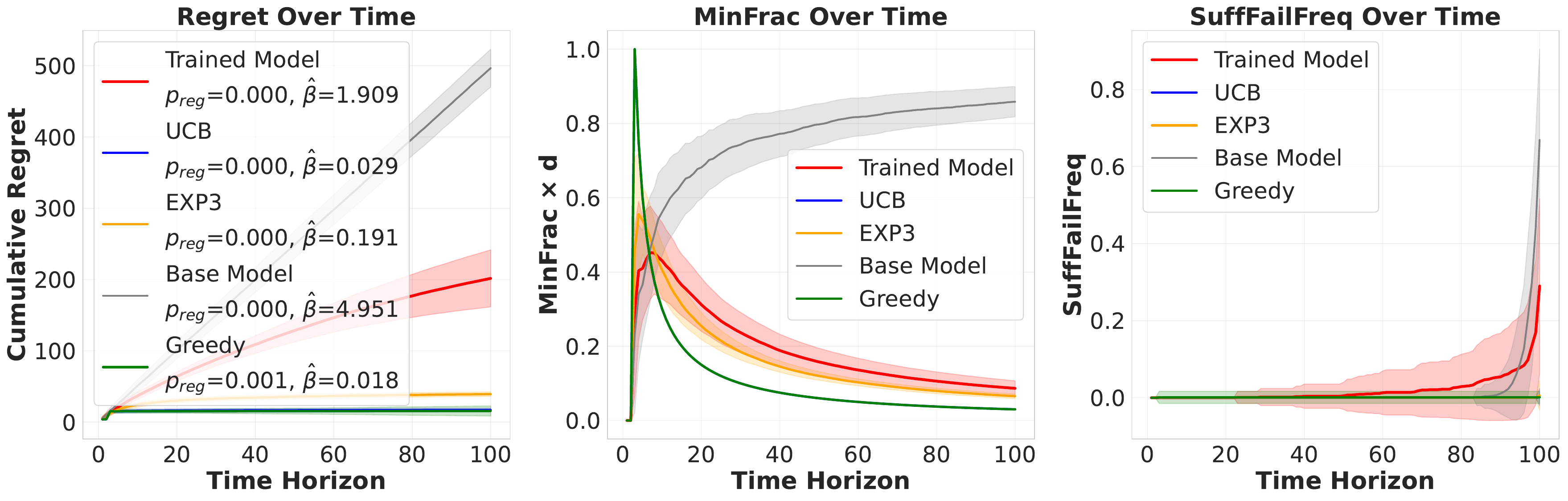}
    \caption{\textbf{The regret over time and the final regret distribution for the \protect\mybluehyperlink{gamma}{Gamma} reward under \texttt{Horizon Generalization}[$T=25 \rightarrow T=100$].} 
     $\texttt{MinFrac}(t)$ also exhibits a trend that reflects a proper \textit{E-E tradeoff}: the metric first increases (active exploration in early rounds) and later decreases (progressive exploitation of optimal actions). Meanwhile, $\texttt{SuffFailFreq}(t)$ maintains a consistently lower value than the Greedy and the Base Model near the end of the horizon, indicating convergence toward more optimal action choices. }
    \label{fig:tfgamma}
\end{figure}

\section{Deferred Experimental Results for \Cref{sec:open-weight-training}}
\label{appendix:sec5results}
\ifthenelse{\boolean{iclr}}
{

}{}

In this section, we provide more results for open-weight LLMs in \Cref{sec:open-weight-training}. We mainly consider two parts of the experimental results deferred from \Cref{sec:open-weight-training}: 1) the experimental results of \textit{distributional} output, where we identify a major problem for \textit{distributional} output for the open-weight LLM, Gemma-2-9b-it -- \textit{high-entropy simplex bias}; 2) the experimental results of a smaller open-weight LLM, Phi-3.5-mini-instruct.

\subsection{Experimental Results of \textit{Distributional} Output}
\label{appendix:distributional-results}

In \Cref{sec:open-weight-training}, we provide results for only \textit{action-based} output, while in this part of the Appendix, we provide results for \textit{distributional} output, and explore the reason why \textit{distributional} output cannot achieve a similar improvement as we have seen in the \textit{action-based} output in open-weight LLMs like Gemma-2-9b-it. We identify that for Gemma-2-9b-it, the simplex it generates usually concentrates around the uniform simplex, and cannot assign a very high probability mass to one action, and a very low probability mass to other actions. We refer to this problem as \textit{high-entropy simplex bias}, where \briefalgorithmexpand{} cannot mitigate this problem in Gemma-2-9b-it. In this appendix and \Cref{sec:training-gpt}, however, the experimental results for the stronger closed-weight LLM like GPT-4o mini demonstrate that \textit{high-entropy simplex bias} is not a universal problem for any LLMs. 

In this part, we will mainly focus on the FOL environment to understand which type of output can be used for \briefalgorithmexpand{}.

\subsubsection{Experimental Results of $d=3$ Actions}

\begin{figure}[!h]
   \centering
   \includegraphics[width=0.9\linewidth]{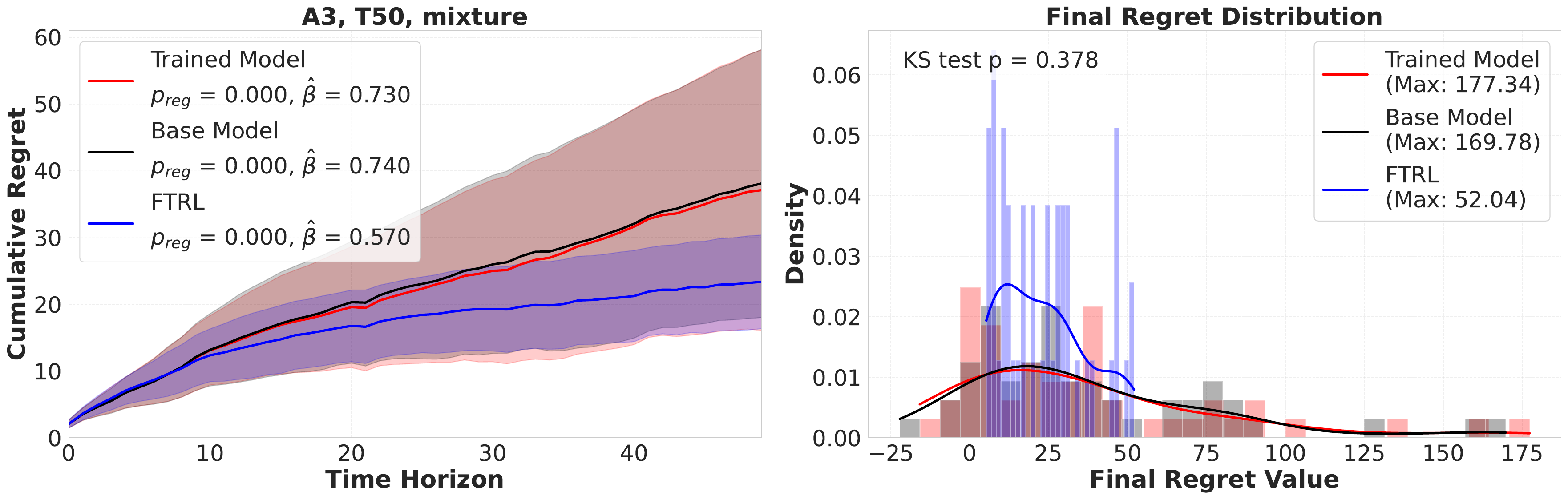}
   \caption{
  \textbf{The regret over time and the final regret distribution for the FOL environment with \texttt{Horizon Generalization}[$T=25 \rightarrow T=50$] on Gemma-2-9b-it}, trained and tested on a mixture of the \protect\mybluehyperlink{gaussianmu}{Gaussian}, \protect\mybluehyperlink{uniform}{Uniform} and \protect\mybluehyperlink{sine}{Sine-trend} rewards. Our trained model shows similar regret value and regret growth rate to the base model, indicating that \briefalgorithmexpand{} with \textit{distributional} outputs is ineffective on Gemma-2-9b-it. 
   }
   \label{fig:open-FOL-indist-appendix-A3}
\end{figure}

We first provide the results of $d=3$ actions using \textit{distributional} output. Similar to the experimental results in \Cref{ssec:exp-results-openweight}, the model is trained with $d=3$ actions and a time horizon of $T=25$ using a mixture of the \mybluehyperlink{gaussianmu}{Gaussian}, \mybluehyperlink{uniform}{Uniform}, and \mybluehyperlink{sine}{Sine-trend} rewards, with policy space $\Pi= \Delta(\cA)$. However, the experiments in this section use \textit{distributional} output and \textit{policy-with-reasoning} format. As in \Cref{ssec:exp-results-openweight}, we also evaluate it with time horizon $T=50$ and a mixture of the \mybluehyperlink{gaussianmu}{Gaussian}, \mybluehyperlink{uniform}{Uniform}, and \mybluehyperlink{sine}{Sine-trend} rewards. From \Cref{fig:open-FOL-indist-appendix-A3}, we can conclude that \briefalgorithmexpand{} cannot minimize the regret for $d=3$ actions. However, in \Cref{sec:training-gpt}, we consistently succeed in training with distributional output. We therefore hypothesize that this is due to the \textit{high-entropy simplex bias} of the open-weight (or weak) models, which we explain in the following subsection.

\subsubsection{Evidence of \textit{high-Entropy Simplex Bias}}
\label{appendix:ssec:distributional-output}
Assume that the LLM never observes information about numbers in the context of distributional output, and thus never encounters probabilities smaller than 0.1 (\textit{i.e.}, no extreme outputs). Since the LLM does not operate through a number but instead treats each digit as a token, the absence of such tokens in training means it cannot generate them. Therefore, if we expect the LLM to produce a versatile policy (or probability distribution), it must at least be exposed to those specific tokens during training. We hypothesize that weak open-weight LLMs are unable to produce high-entropy policies. We illustrate this with experiments by examining the distribution of the probabilistic simplex generated by Gemma-2-9b-it.

\begin{figure}[!h]
   \centering
   \includegraphics[width=0.9\linewidth]{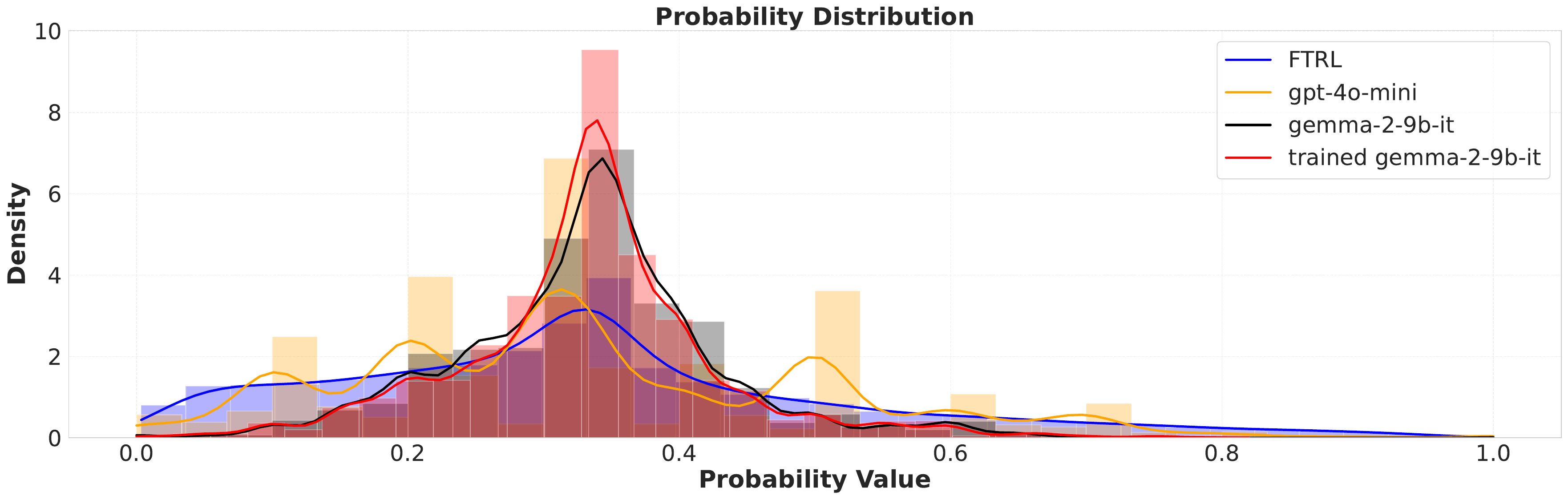}
   \caption{
   \textbf{Distribution of the probabilistic simplex with $d = 3$ actions.} We collect the probability values of multiple probabilistic simplices generated by Gemma-2-9b-it, and plot the distribution. The simplex of Gemma-2-9b-it has significantly higher entropy than baseline algorithms and the more capable closed-weight LLM, GPT-4o mini. The simplex of the trained model has a similar high entropy.
   }
   \label{fig:open-FOL-indist-appendix-dist}
\end{figure}

\Cref{fig:open-FOL-indist-appendix-dist} indicates that FTRL and the capable closed-weight LLM like OpenAI's GPT-4o mini, although they may have some specific preference on certain value (like $0.10$), generally include both simplices with high entropy (values concentrated near uniform policy) and low entropy (values which are relatively far from near uniform policy). However, the simplex of the open-weight LLMs like Gemma-2-9b-it concentrates near a uniform policy with a low variance. The trained model doesn't show an improvement for the \textit{high-entropy simplex bias} problem, which is a potential limitation for \briefalgorithmexpand{}, especially in small LLMs, where its original distribution for probabilistic simplices has certain preferences or limitations.

\subsubsection{Experimental Results of Simpler Probabilistic Simplices}

\begin{figure}[!h]
   \centering
   \includegraphics[width=0.9\linewidth]{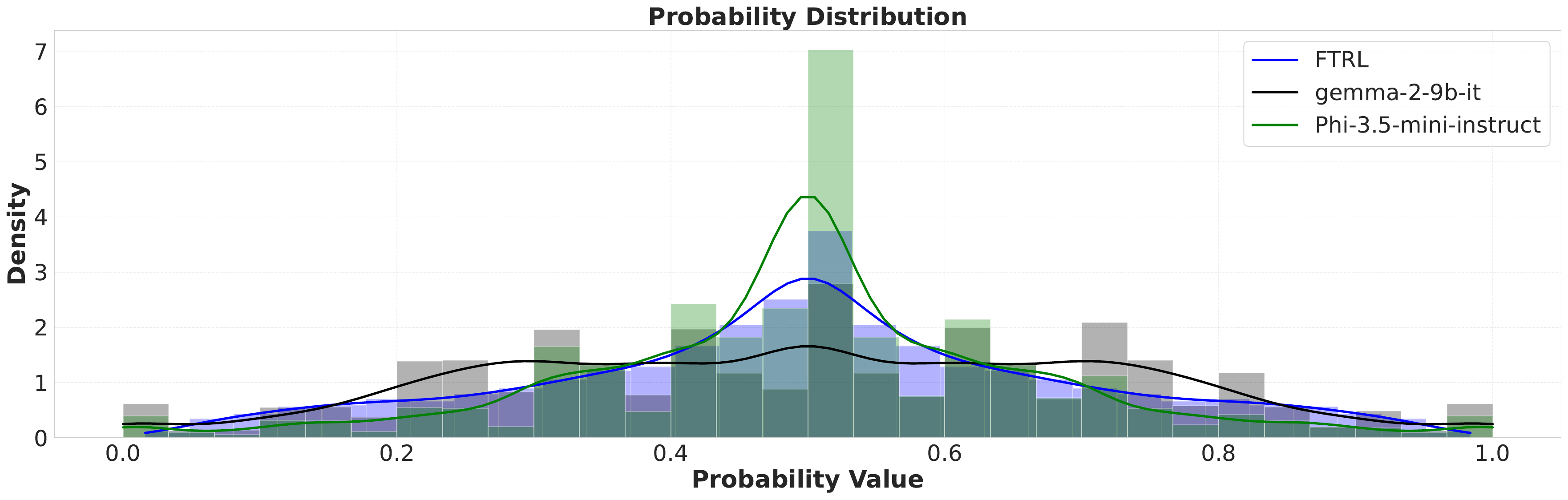}
   \caption{
   \textbf{Distribution of the probabilistic simplex with $d=2$ actions.} We gather the probability values from multiple probabilistic simplices generated by Gemma-2-9b-it and Phi-3.5-mini-instruct and visualize their distributions. The results indicate that the simplices produced by both models are not severely affected by the \textit{high-entropy simplex bias}.
   }
   \label{fig:open-FOL-indist-appendix-dist2}
\end{figure}
From the discussion above, we suspect that the problem for \textit{distributional} output is that the model faces a serious \textit{high-entropy simplex bias} problem. In this part, we examine simpler probabilistic simplices of action $d=2$ for the FOL environment. The experimental results show that if the probabilistic simplices of action $d=2$ are simple enough, Gemma-2-9b-it or even smaller open-weight LLMs like Phi-3.5-mini-instruct can have improved regret behavior when applying \briefalgorithmexpand{}.

We first show results of the distribution of the probabilistic simplex with action $d=2$ in \Cref{fig:open-FOL-indist-appendix-dist2}, which demonstrate for a simpler simplex (action $d=2$), the \textit{high-entropy simplex bias} is not that severe. 

\begin{figure}[!h]
   \centering
   \includegraphics[width=0.9\linewidth]{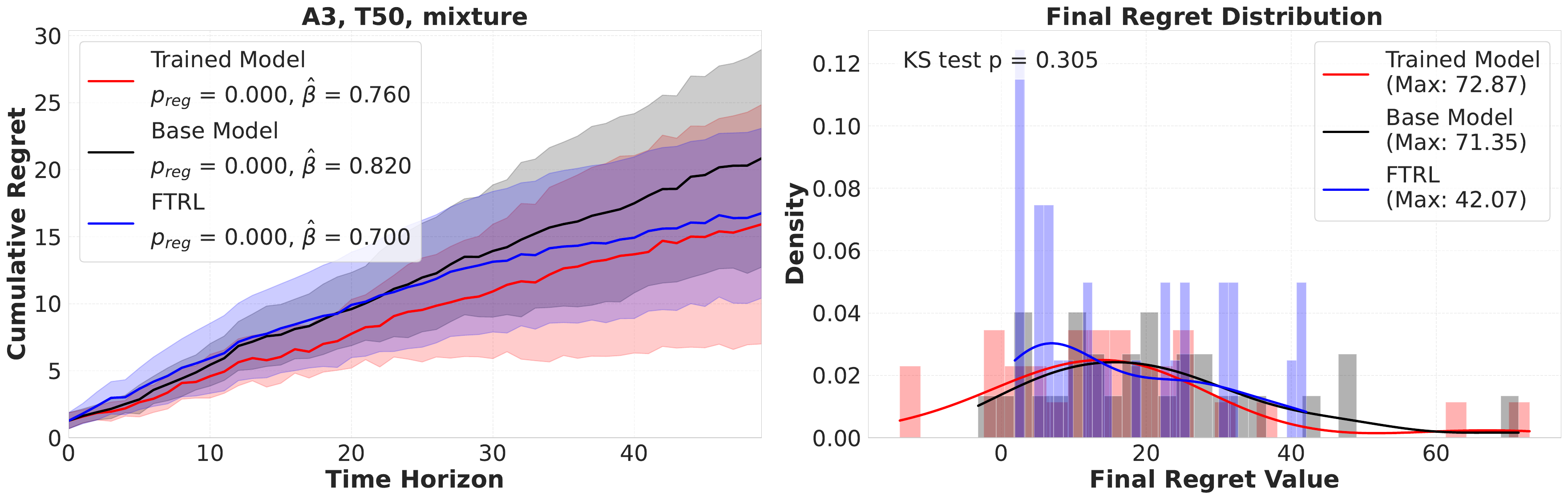}
   \caption{
   \textbf{The regret over time and the final regret distribution for the FOL environment under \texttt{Horizon Generalization}[$T=25 \rightarrow T=50$] on Gemma-2-9b-it,} trained and evaluated on a mixture of the \protect\mybluehyperlink{gaussianmu}{Gaussian}, \protect\mybluehyperlink{uniform}{Uniform}, and \protect\mybluehyperlink{sine}{Sine-trend} rewards with $d=2$ actions. The trained model attains \textbf{lower regret} than both the baseline algorithm FTRL and the base model Gemma-2-9b-it, demonstrating \textbf{sublinear regret behavior}. 
   }
   \label{fig:open-FOL-indist-appendix-A2}
\end{figure}

We also provide results on Gemma-2-9b-it. We train the model with $d=2$ actions and a time horizon of $T=25$ using a mixture of the \mybluehyperlink{gaussianmu}{Gaussian}, \mybluehyperlink{uniform}{Uniform}, and \mybluehyperlink{sine}{Sine-trend} rewards, and evaluate it with the same distribution and a longer time horizon $T=50$ in \Cref{tab:open-FOL-indist-appendix-A2} and \Cref{fig:open-FOL-indist-appendix-A2}. 

{
\begin{table}[h!]
\centering
\resizebox{\textwidth}{!}{%
\begin{tabular}{l|ccc||ccc||ccc}
\thickhline
\multicolumn{1}{c|}{} & \multicolumn{3}{c||}{\textbf{\mybluehyperlink{gaussianmu}{Gaussian}}} & \multicolumn{3}{c||}{\textbf{\mybluehyperlink{uniform}{Uniform}}} & \multicolumn{3}{c}{\textbf{\mybluehyperlink{sine}{Sine-trend}}} \\ \cline{2-10}
\multicolumn{1}{c|}{} & max(LR) & avg(LR) & $\hat{\beta}$ & max(LR) & avg(LR) & $\hat{\beta}$ & max(LR) & avg(LR) & $\hat{\beta}$ \\ \hline
FTRL & 36.60 & 20.22 & 0.87 & 43.25 & 29.44 & 0.73 & 9.72 & 4.53 & 0.2 \\
\cline{1-10}
Gemma-2-9b-it & \cellcolor{yellow!30}47.44 & 21.18 & \cellcolor{yellow!30}0.91 & 71.28 & 28.93 & 0.84 & 56.04 & 9.41 & 0.34 \\
Trained Gemma-2-9b-it & 51.28 & \cellcolor{yellow!30}21.14 & 0.92 & \cellcolor{yellow!30}59.85 & \cellcolor{yellow!30}25.80 & \cellcolor{yellow!30}0.82 & \cellcolor{yellow!30}52.88 & \cellcolor{yellow!30}8.61 & \cellcolor{yellow!30}0.24 \\ \thickhline
\end{tabular}
}
\caption{
\textbf{Summary of the regret value for the FOL environment with \texttt{Horizon Generalization}[$T=25\rightarrow T=50$] on Gemma-2-9b-it}, trained and tested on a mixture of the \protect\mybluehyperlink{gaussianmu}{Gaussian}, \protect\mybluehyperlink{uniform}{Uniform}, and \protect\mybluehyperlink{sine}{Sine-trend} rewards with $d=2$ actions. The trained model achieves lower regret value and sublinear regret growth than the base model on the \protect\mybluehyperlink{uniform}{Uniform} and \protect\mybluehyperlink{sine}{Sine-trend} rewards. }
\label{tab:open-FOL-indist-appendix-A2}
\end{table}
}

\begin{figure}[!h]
   \centering
   \includegraphics[width=0.9\linewidth]{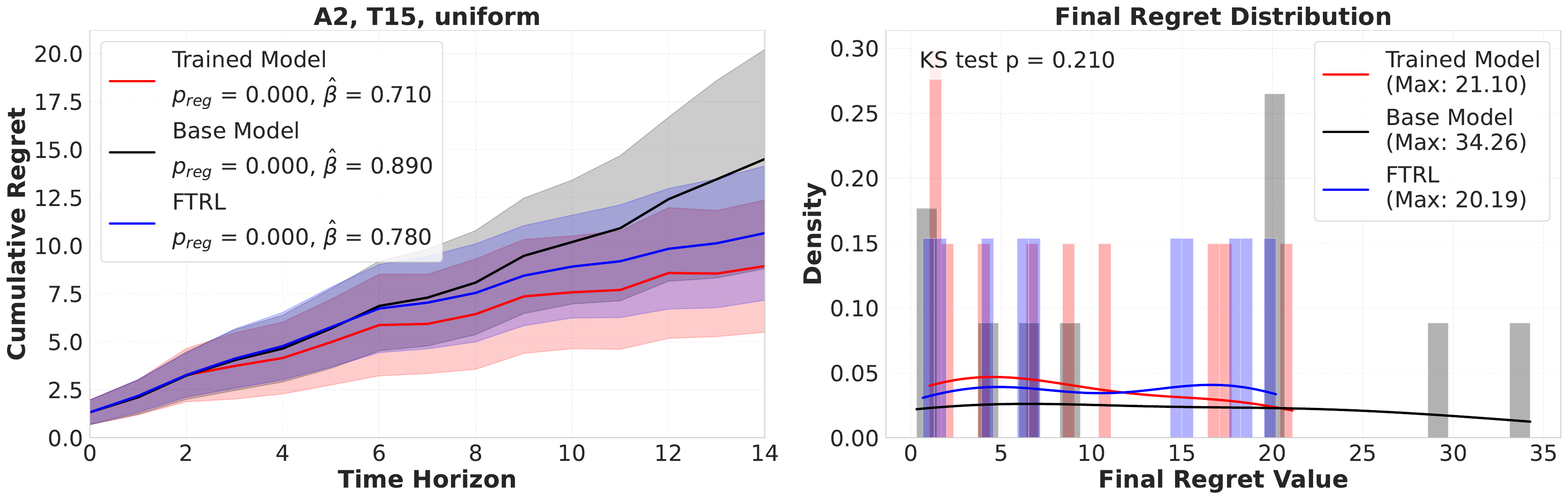}
   \caption{
   \textbf{The regret over time and the final regret distribution for the FOL environment under \texttt{In-Distribution} on Phi-3.5-mini-instruct}, trained and tested with the \protect\mybluehyperlink{uniform}{Uniform} reward ($T=15$, $d=2$). The trained model shows consistently lower cumulative regret and a slower rate of regret increase compared to both the baseline algorithm (FTRL) and the untrained base model, confirming sublinear regret performance.
   }
   \label{fig:phi-dialogue}
\end{figure}

We then provide results on Phi-3.5-mini-instruct. We train the model with $d=2$ actions and a time horizon of $T=15$ using the \mybluehyperlink{uniform}{Uniform} reward, and evaluate it with the same distribution and the same time horizon $T=15$ in \Cref{fig:phi-dialogue}. 

This result shows that for $d = 2$, \briefalgorithmexpand{} can significantly minimize the regret value, for Gemma-2-9b-it and even smaller open-weight LLMs like Phi-3.5-mini-instruct. Since $d=2$ requires an easier simplex than $d=3$, this result probably indicates that if Gemma-2-9b-it (or even Phi-3.5-mini-instruct) is capable enough to generate a "good" probabilistic simplex, \briefalgorithmexpand{} will still work for the \textit{distributional} output. 

\subsection{Comparing \textit{Dialogue-Type} and \textit{Summary-Type} Interactions in \briefalgorithmexpand{}}

\label{appendix:summary-results}

In this section, we compare \textit{dialogue-type} and \textit{summary-type} interactions in \briefalgorithmexpand{}. The advantages of \textit{dialogue-type} interactions are as follows: (1) it retains the complete history of the interaction directly in the prompt, whereas \textit{summary-type} interactions must extract information from previous outputs and integrate a manually or automatically generated summary into the current prompt—an approach that may be unnatural for real-world use cases; and (2) when applied to \briefalgorithmexpand{}, \textit{summary-type} interactions are prone to overfitting.

\begin{figure}[!h]
   \centering
   \includegraphics[width=0.9\linewidth]{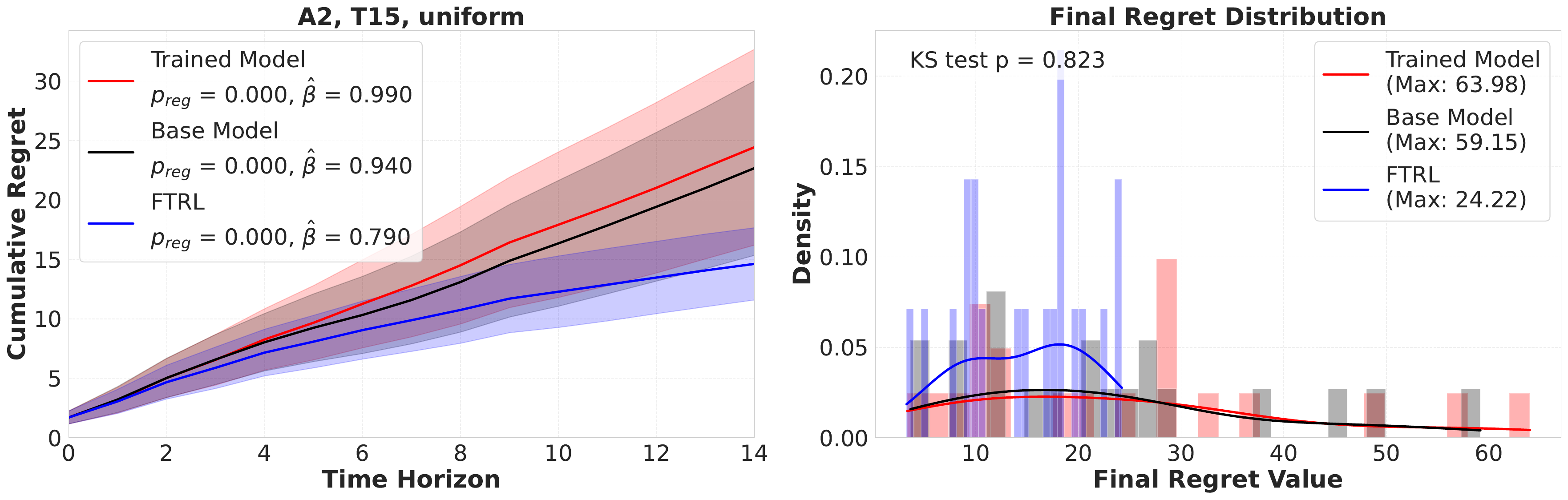}
   \caption{
   \textbf{The regret over time and the final regret distribution for the FOL environment under \texttt{In-Distribution} on Phi-3.5-mini-instruct}, trained and tested with the \protect\mybluehyperlink{uniform}{Uniform} reward ($T=15$, $d=2$). Unlike the other experiments, this one is based on \textbf{\textit{summary-type} interaction}. In this case, the trained model actually performs worse, showing even higher regret than the base model (Phi-3.5-mini-instruct). 
   }
   \label{fig:phi-summary}
\end{figure}

\begin{figure}[!h]
   \centering
   \includegraphics[width=0.9\linewidth]{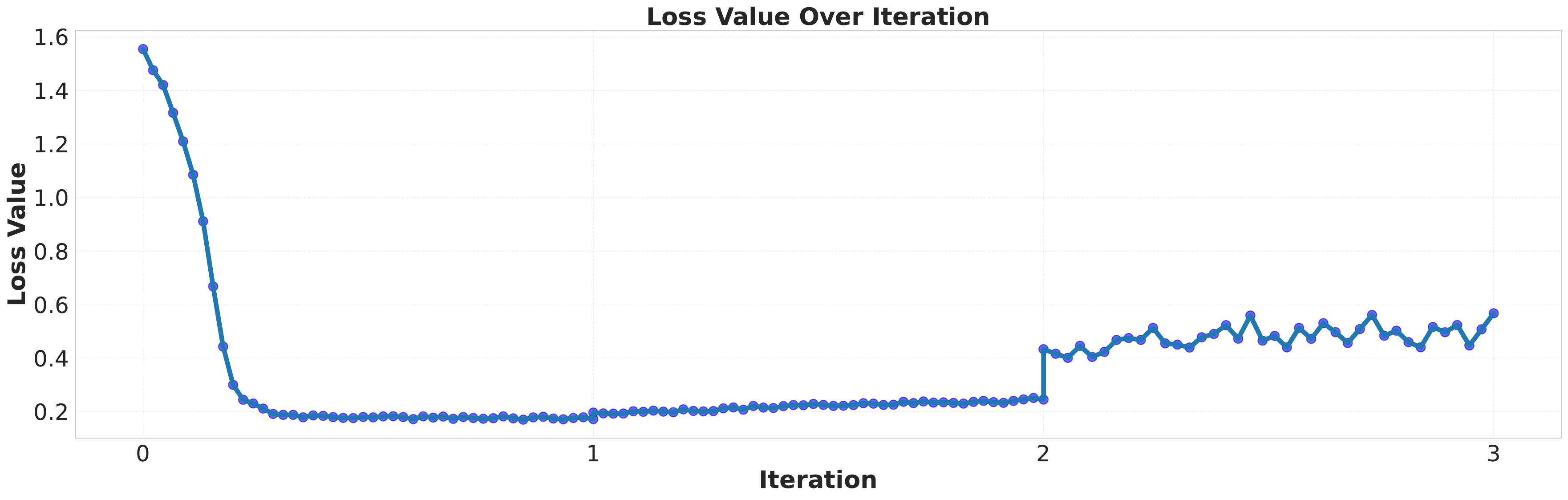}
   \caption{
   \textbf{Loss over iteration for the \protect\mybluehyperlink{uniform}{Uniform} reward ($T = 15$) with \textit{summary-type} ineraction.} Although the loss drop significantly during the iteration 1. The loss increase during the iteration 2 and iteration 3, showing that training is overfitting. 
   }
   \label{fig:phi-loss}
\end{figure}

We compare the results of \textit{summary-type} interaction with \textit{dialogue-type} interaction (see \Cref{fig:phi-dialogue} for \textit{dialogue-type} interaction) on Phi-3.5-mini-instruct. For \textit{summary-type} interaction, we also train the model with $d=2$ actions and a time horizon of $T=15$ using the \mybluehyperlink{uniform}{Uniform} reward, and evaluate it with the same distribution and the same time horizon $T=15$. \Cref{fig:phi-summary} show the regret behavior of \textit{summary-type} interaction, where we find the regret behavior of the trained model is worse than the base model. We identify this as a case of overfitting, possibly because the Phi-3.5-mini-instruct is not capable enough to do the training using \textit{summary-type} interaction. The loss plot over iterative training in \Cref{fig:phi-loss} further supports this claim. 

\clearpage
\subsection{Deferred Figures of the FOL Environment}

\begin{figure}[!h]
   \centering
   \includegraphics[width=0.9\linewidth]{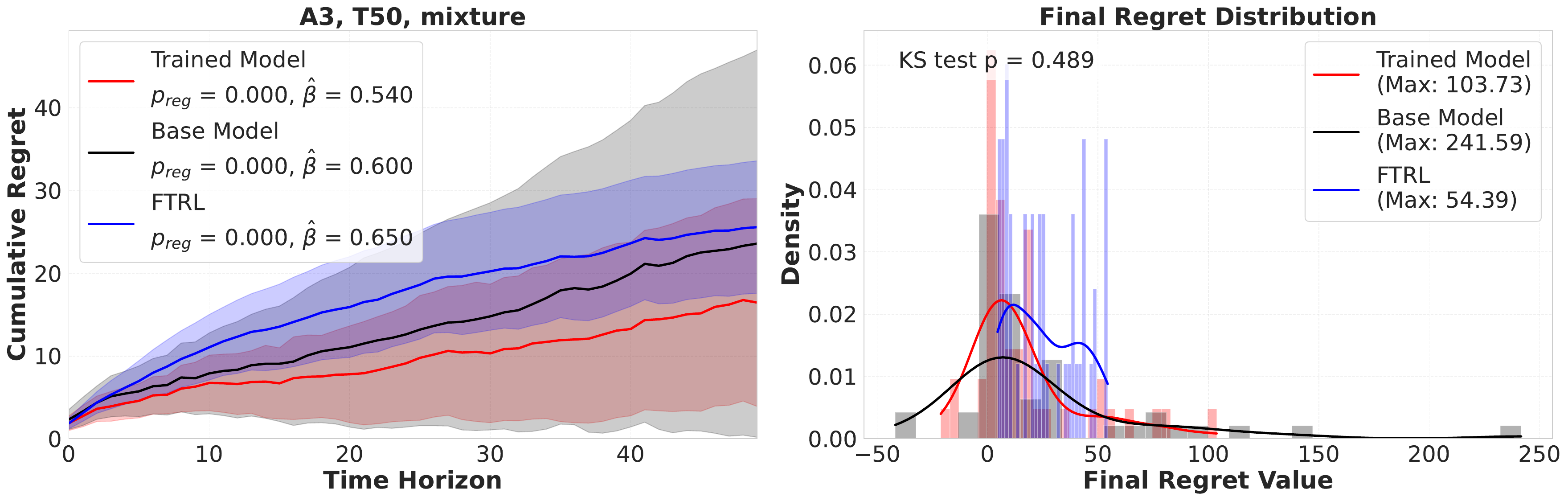}
   \caption{
   \textbf{The regret over time and final regret distribution for the FOL environment under \texttt{Horizon Generalization}[$T=25 \rightarrow T=50$] on Gemma-2-9b-it}, trained and evaluated with a mixture of the \protect\mybluehyperlink{gaussianmu}{Gaussian}, \protect\mybluehyperlink{uniform}{Uniform}, and \protect\mybluehyperlink{sine}{Sine-trend} rewards, which shows a lower regret value and sublinear regret growth after \briefalgorithmexpand{}. 
   }
   \label{fig:open-FOL-indist}
\end{figure}
\begin{figure}[!h]
   \centering
   \includegraphics[width=0.9\linewidth]{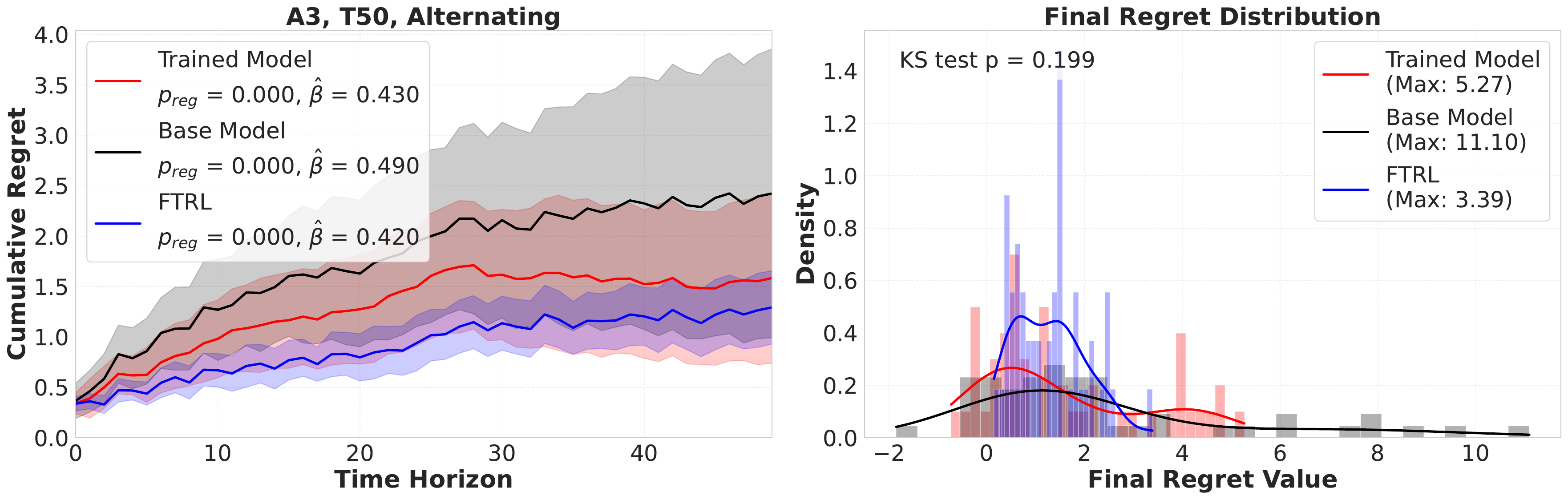}
   \caption{\textbf{The regret over time and the final regret distribution for the FOL environment under both \texttt{Horizon Generalization}[$T=25 \rightarrow T=50$] and \texttt{Reward Generalization}[a mixture of the \protect\mybluehyperlink{gaussianmu}{Gaussian}, \protect\mybluehyperlink{uniform}{Uniform}, and \protect\mybluehyperlink{sine}{Sine-trend} rewards $\rightarrow$ \protect\mybluehyperlink{alternating}{Alternating}] on Gemma-2-9b-it}, which shows a significant lower regret value and sublinear regret growth after \briefalgorithmexpand{}. 
   }
   \label{fig:open-FOL-alternating}
\end{figure}

\begin{figure}[!h]
   \centering
   \includegraphics[width=0.9\linewidth]{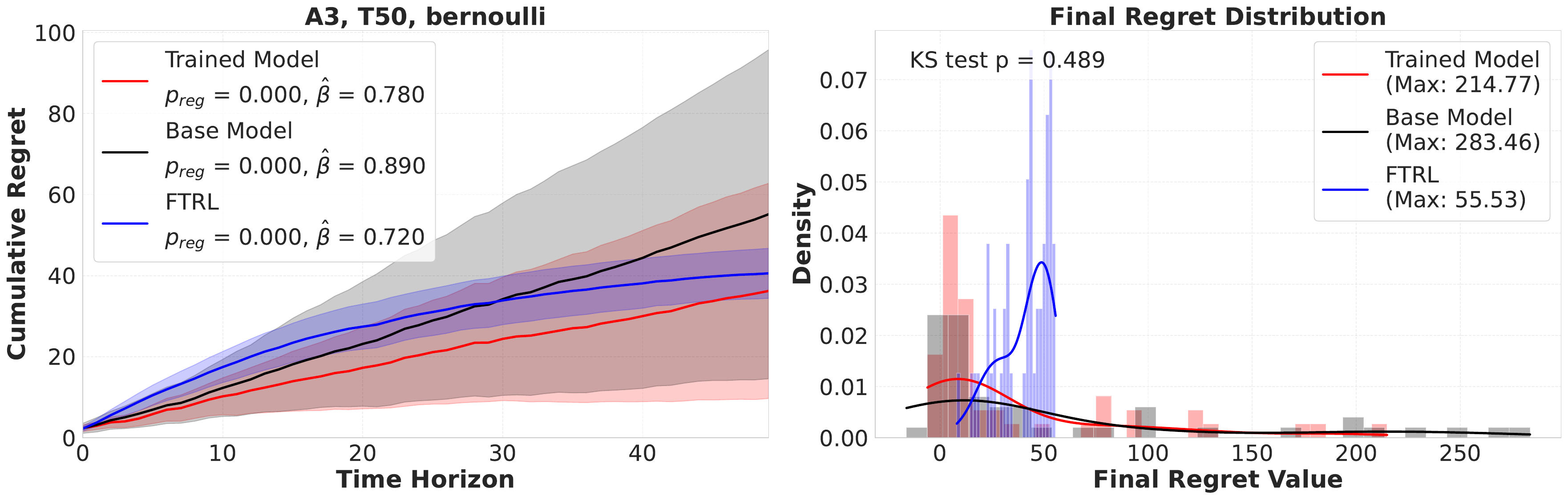}
   \caption{\textbf{The regret over time and the final regret distribution for the FOL environment under both \texttt{Horizon Generalization}[$T=25 \rightarrow T=50$] and \texttt{Reward Generalization}[a mixture of the \protect\mybluehyperlink{gaussianmu}{Gaussian}, \protect\mybluehyperlink{uniform}{Uniform}, and \protect\mybluehyperlink{sine}{Sine-trend} rewards $\rightarrow$ \protect\mybluehyperlink{bernoulli}{Bernoulli}] on Gemma-2-9b-it}, which shows a significant lower regret value and sublinear regret growth after \briefalgorithmexpand{}.
   }
   \label{fig:open-FOL-bernoulli}
\end{figure}

\begin{figure}[!h]
   \centering
   \includegraphics[width=0.9\linewidth]{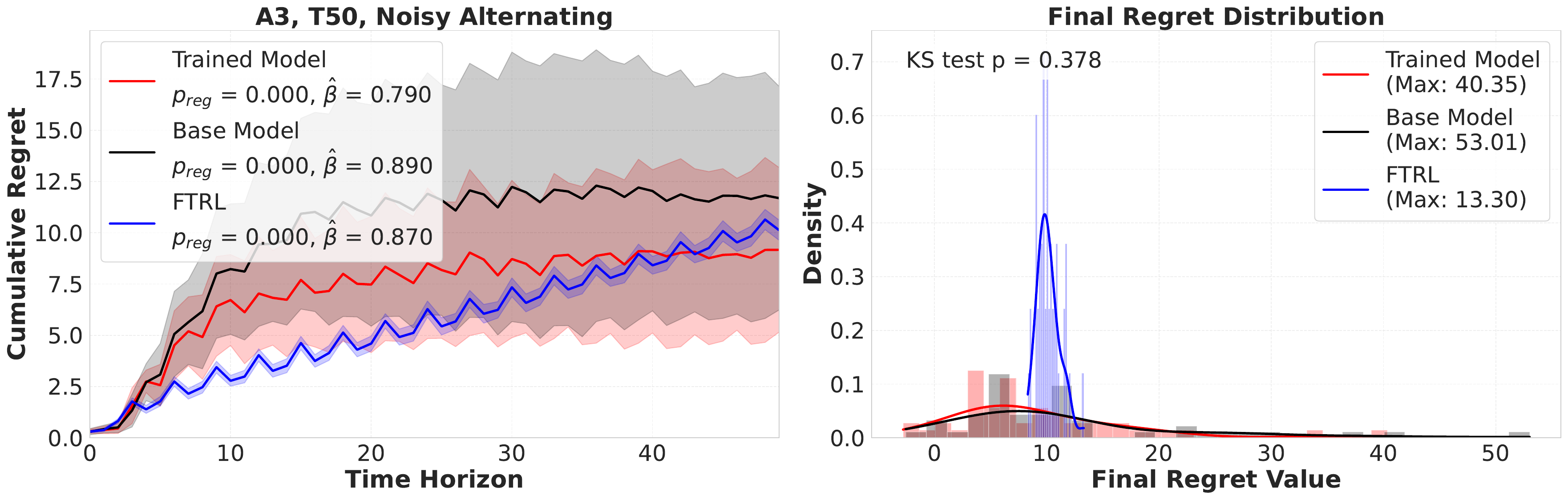}
   \caption{\textbf{The regret over time and the final regret distribution for the FOL environment under both \texttt{Horizon Generalization}[$T=25 \rightarrow =50$] and \texttt{Reward Generalization}[a mixture of the \protect\mybluehyperlink{gaussianmu}{Gaussian}, \protect\mybluehyperlink{uniform}{Uniform}, and \protect\mybluehyperlink{sine}{Sine-trend} rewards $\rightarrow$ \protect\mybluehyperlink{noisy}{Noisy Alternating}] on Gemma-2-9b-it}, which shows a significant lower regret value and a sublinear regret growth after \briefalgorithmexpand{}. 
   }
   \label{fig:open-FOL-noisy-alternating}
\end{figure}

\subsection{Deferred Figures of the MAB Environment}

\begin{figure}[H]
   \centering
   \includegraphics[width=0.9\linewidth]{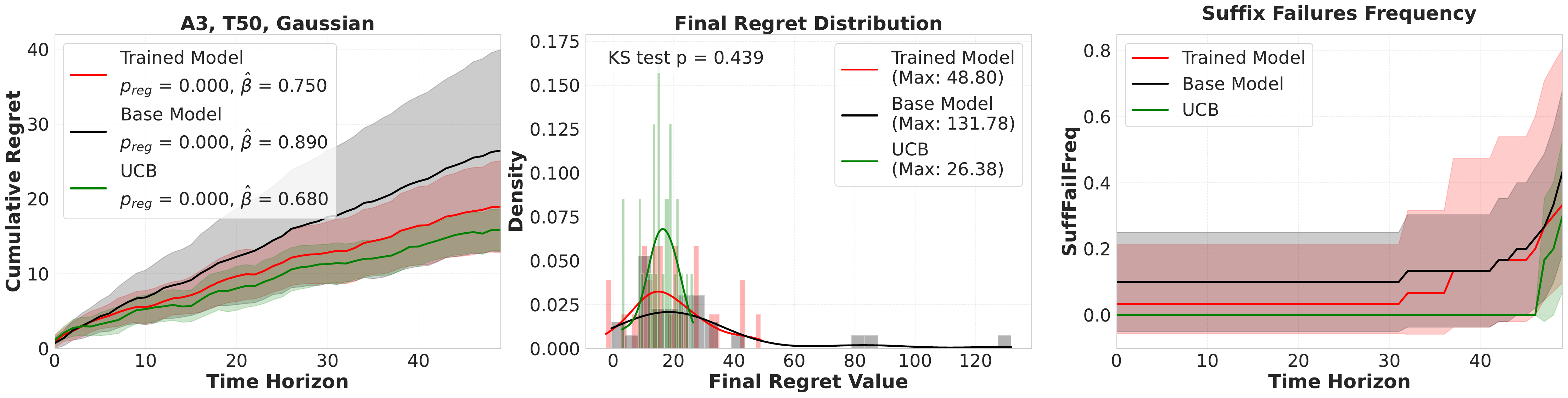}
   \caption{\textbf{The regret over time, the final regret distribution, and the exploration metric \texttt{SuffFailFreq$(t)$} for the MAB environment under \texttt{Horizon Generalization}[$T=25 \rightarrow T=50$] on Gemma-2-9b-it}, which demonstrates a lower regret value, sublinear growth rate, and improved exploration after \briefalgorithmexpand{}. }
   \label{fig:open-MAB-gaussian-T50-gemma}
\end{figure}

\begin{figure}[H]
   \centering
   \includegraphics[width=0.9\linewidth]{fig/qwen3_MAB/gaussian_T100.pdf}
   \caption{\textbf{The regret over time, the final regret distribution, and the exploration metric \texttt{SuffFailFreq$(t)$} for the MAB environment under \texttt{Horizon Generalization}[$T=25 \rightarrow T=100$] on Qwen3-8B}, which demonstrates a lower regret value, sublinear growth rate, and improved exploration after \briefalgorithmexpand{}. }
   \label{fig:open-MAB-gaussian-T50-qwen}
\end{figure}

\begin{figure}[H]
   \centering
   \includegraphics[width=0.9\linewidth]{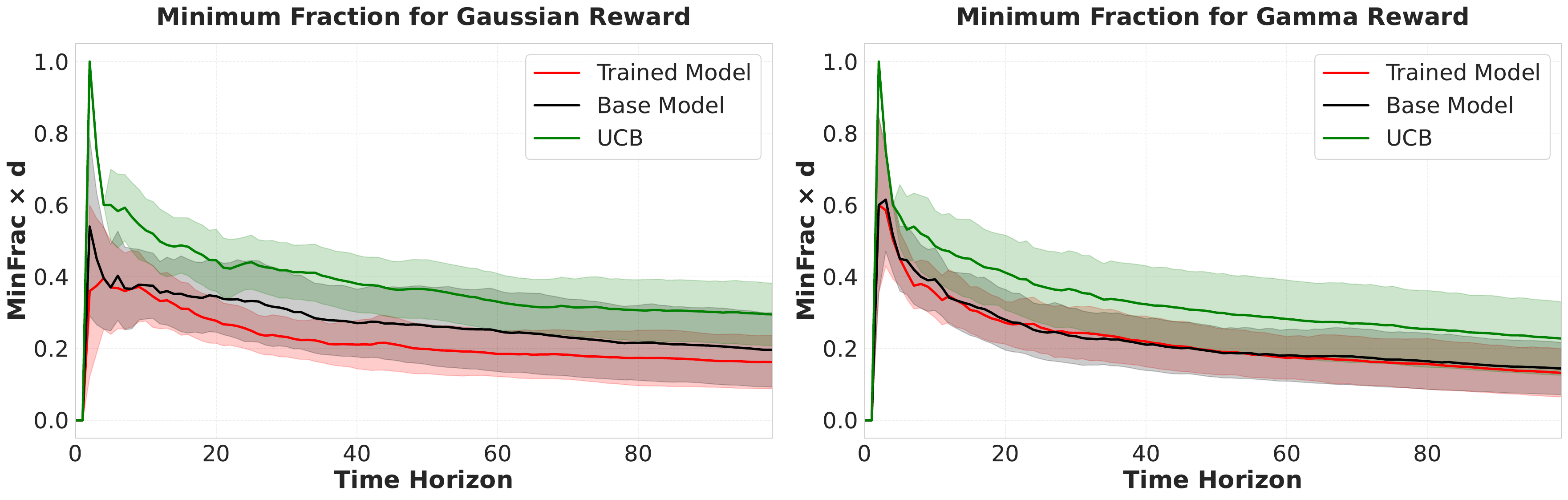}
   \caption{\textbf{The exploitation metric \texttt{MinFrac$(t)$} for the MAB environment under both \texttt{Horizon Generalization}[$T=25 \rightarrow T=100$] on Qwen3-8B} using both the \protect\mybluehyperlink{gaussianmu}{Gaussian} and \protect\mybluehyperlink{gamma}{Gamma} rewards. The result shows that the base model already exhibits strong exploitation behavior. }
   \label{fig:open-MAB-minfrac}
\end{figure}

\subsection{Quantitative Analysis: Assessing How \briefalgorithmexpand{} Enhances the Robustness to Reward Fluctuations on the Open-Weight LLM Qwen3-8B}

\label{appendix:reasoning}

\begin{figure}[H]
    \centering
    \includegraphics[width=0.6\linewidth]{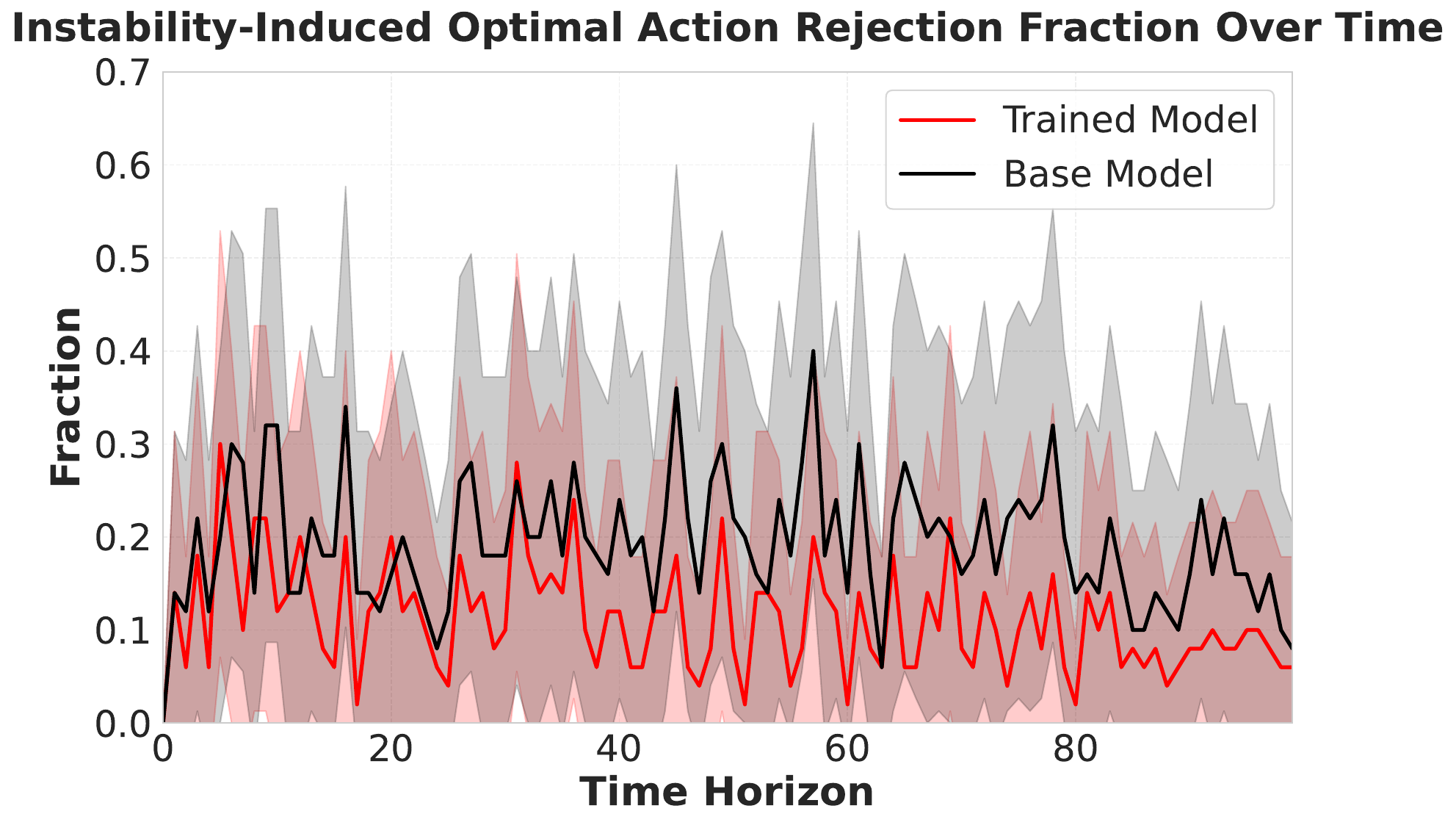}
    \caption{
    \textbf{The Instability-Induced Optimal Action Rejection Fraction Over Time.} The trained model has significantly lower fraction than the base model, indicating its robustness over the instable reward behavior. 
    }
    \label{fig:reasoning-variance}
\end{figure}

To gain deeper insight into how \briefalgorithmexpand{} enhances the decision-making capability of LLMs on Qwen3-8B, we perform a quantitative analysis on the robustness to reward fluctuations comparing the outputs of the base model and the trained model for the MAB environment inspired by a qualitative observation that the base model exhibits high sensitivity to reward fluctuations and struggles to handle optimal actions with both high mean rewards and high variance. 

Using GPT-4o mini as a zero-shot classifier, we categorize the model outputs into two groups depending on whether they explicitly mention that the optimal action is avoided due to unstable reward behavior and report the resulting fraction in \Cref{fig:reasoning-variance}. 

We observe that after \briefalgorithmexpand{}, the instability-induced optimal action rejection fraction significantly drops, indicating that the trained model exhibits an improvement in robustness to unstable reward behavior.

\section{Deferred Experimental Results for \Cref{sec:training-gpt}}
\label{appendix:sec6results}
We include here the tables that we deferred  from the main paper due to page constraints.

\subsection{Full-Information Online Learning}
\subsubsection{\texttt{In-Distribution} Evaluation}
In the \texttt{In-Distribution} evaluation, which is the same configuration as training, we observe improvements in both the max value and average value of the final regret and the empirical growth of regret in \Cref{tab:FOL-indist}. 

{
\begin{table}[H]
\centering
\resizebox{\textwidth}{!}{%
\begin{tabular}{l|ccc||ccc||ccc}
\thickhline
\multicolumn{1}{c|}{} & \multicolumn{3}{c||}{\textbf{\mybluehyperlink{gaussianmu}{Gaussian}}} & \multicolumn{3}{c||}{\textbf{\mybluehyperlink{uniform}{Uniform}}} & \multicolumn{3}{c}{\textbf{\mybluehyperlink{sine}{Sine-trend}}} \\ \cline{2-10}
\multicolumn{1}{c|}{} & max(LR) & avg(LR) & $\hat{\beta}$ & max(LR) & avg(LR) & $\hat{\beta}$ & max(LR) & avg(LR) & $\hat{\beta}$ \\ \hline
FTRL & 29.72 & 18.52 & 0.69 & 33.24 & 19.32 & 0.81 & 37.32 & 29.22 & 0.36 \\
GPT-4o mini & 40.16 & 16.35 & 0.64 & 36.75 & 15.69 & 0.74 & 37.24 & 8.98 & 0.43 \\
Trained GPT-4o mini & \cellcolor{yellow!30}29.89 & \cellcolor{yellow!30}15.03 & \cellcolor{yellow!30}0.61 & \cellcolor{yellow!30}25.76 & \cellcolor{yellow!30}13.12 & \cellcolor{yellow!30}0.66 & \cellcolor{yellow!30}25.39 & \cellcolor{yellow!30}8.35 & \cellcolor{yellow!30}0.39 \\ \thickhline
\end{tabular}
}
\caption{
\textbf{Summary of the regret value for the FOL environment under \texttt{In-Distribution} on GPT-4o mini,} trained and evaluated on a mixture of the \protect\mybluehyperlink{gaussianmu}{Gaussian}, \protect\mybluehyperlink{uniform}{Uniform}, and \protect\mybluehyperlink{sine}{Sine-trend} rewards with time horizon $T=15$. 
We report the maximum and average values of the cumulative regret at round $T$, denoted as {max(LR)} and {avg(LR)},  respectively.
The exponent $\hat{\beta}$ is estimated from the empirical growth of regret, where regret scales approximately as $\Theta(t^{\hat{\beta}})$ (see \Cref{ssec:regret}).
Highlighted (yellow) cells indicate the lower {regret} value between {those corresponding to} \textit{GPT-4o mini} and \textit{Trained GPT-4o mini}, emphasizing which performs better in terms of regret at round $T$ and growth rate exponent ($\hat{\beta}$) for each reward generation process. This model consistently shows improved regret over time for \texttt{In-Distribution} tasks.}
\label{tab:FOL-indist}
\end{table}
}

\subsubsection{Evaluation for Generalization}
\begin{table}[H]
\centering
\resizebox{\textwidth}{!}{%
\begin{tabular}{l|ccc||ccc||ccc}
\thickhline
\multicolumn{1}{c|}{} & \multicolumn{3}{c||}{\textbf{\mybluehyperlink{gaussianmu}{Gaussian}}} & \multicolumn{3}{c||}{\textbf{\mybluehyperlink{uniform}{Uniform}}} & \multicolumn{3}{c}{\textbf{\mybluehyperlink{sine}{Sine-trend}}} \\ \cline{2-10}
\multicolumn{1}{c|}{} & max(LR) & avg(LR) & $\hat{\beta}$ & max(LR) & avg(LR) & $\hat{\beta}$ & max(LR) & avg(LR) & $\hat{\beta}$ \\ \hline
FTRL & 30.36 & 18.44 & 0.63 & 29.30 & 17.55 & 0.76 & 25.54 & 10.65 & 0.47 \\
GPT-4o mini & \cellcolor{yellow!30}31.07 & 17.10 & 0.61 & 34.91 & 14.67 & 0.72 & 33.85 & 9.47 & 0.42 \\
Trained GPT-4o mini & 36.26 & \cellcolor{yellow!30}16.56 & \cellcolor{yellow!30}0.60 & \cellcolor{yellow!30}24.21 & \cellcolor{yellow!30}12.76 & \cellcolor{yellow!30}0.64 & \cellcolor{yellow!30}33.65 & \cellcolor{yellow!30}8.57 & \cellcolor{yellow!30}0.39 \\
\thickhline
\end{tabular}%
}
\caption{ \textbf{Summary of the regret value for the FOL environment under \texttt{Linguistic Context Generalization}[GPT-4o $\rightarrow$ Gemini-2.0-Flash] on GPT-4o mini,} trained and evaluated on a mixture of the \protect\mybluehyperlink{gaussianmu}{Gaussian}, \protect\mybluehyperlink{uniform}{Uniform}, and \protect\mybluehyperlink{sine}{Sine-trend} rewards with time horizon $T=15$. 
Highlighted (yellow) cells indicate the lower value between the base model and the trained model, highlighting which model performs better in terms of cumulative regret at round~$T$ and the growth rate exponent~$\hat{\beta}$ for each reward generation process. These results demonstrate consistent improvement of the trained model under \texttt{Linguistic Context Generalization}. }
\label{tab:FOL-ood-gemini}
\end{table}
\begin{table}[H]
\centering
\resizebox{0.75\textwidth}{!}{%
\begin{tabular}{l|ccc||ccc}
\thickhline
\multicolumn{1}{c|}{} & \multicolumn{3}{c||}{\textbf{\mybluehyperlink{alternating}{Alternating}}} & \multicolumn{3}{c}{\textbf{\mybluehyperlink{bernoulli}{Bernoulli}}} \\ \cline{2-7}
\multicolumn{1}{c|}{} & max(LR) & avg(LR) & $\hat{\beta}$ & max(LR) & avg(LR) & $\hat{\beta}$ \\ \hline
FTRL & 6.19 & 6.19 & 0.20 & 28.33 & 10.82 & 0.72 \\
GPT-4o mini & \cellcolor{yellow!30}35.70 & 28.21 & 0.71 & 32.14 & 10.10 & \cellcolor{yellow!30}0.71 \\
Trained GPT-4o mini & 37.30 & \cellcolor{yellow!30}28.20 & \cellcolor{yellow!30}0.70 & \cellcolor{yellow!30}30.85 & \cellcolor{yellow!30}9.66 & 0.73 \\
\thickhline
\end{tabular}%
}
\caption{
\textbf{Summary of the regret value for the FOL environment under \texttt{Reward Generalization}[a mixture of the \protect\mybluehyperlink{gaussianmu}{Gaussian}, \protect\mybluehyperlink{uniform}{Uniform}, and \protect\mybluehyperlink{sine}{Sine-trend} rewards $\rightarrow$ \protect\mybluehyperlink{alternating}{Alternating}, \protect\mybluehyperlink{bernoulli}{Bernoulli}] on GPT-4o mini,} trained on a mixture of the \protect\mybluehyperlink{gaussianmu}{Gaussian}, \protect\mybluehyperlink{uniform}{Uniform}, and \protect\mybluehyperlink{sine}{Sine-trend} rewards. 
Highlighted (yellow) cells indicate the lower value between the base model and the trained model, highlighting which model performs better in terms of cumulative regret at round~$T$ and the growth rate exponent~$\hat{\beta}$ for each reward generation process. The trained model achieves comparable or better performance under \texttt{Reward Generalization}, demonstrating that its regret behavior does not significantly degrade under distribution shift of the reward. 
}
\label{tab:FOL-ood-reward}
\end{table}

\begin{table}[H]
\centering
\resizebox{\textwidth}{!}{%
\begin{tabular}{l|ccc||ccc||ccc}
\thickhline
\multicolumn{1}{c|}{} & \multicolumn{3}{c||}{\textbf{\mybluehyperlink{gaussianmu}{Gaussian}}} & \multicolumn{3}{c||}{\textbf{\mybluehyperlink{uniform}{Uniform}}} & \multicolumn{3}{c}{\textbf{\mybluehyperlink{sine}{Sine-trend}}} \\ \cline{2-10}
\multicolumn{1}{c|}{} & max(LR) & avg(LR) & $\hat{\beta}$ & max(LR) & avg(LR) & $\hat{\beta}$ & max(LR) & avg(LR) & $\hat{\beta}$ \\ \hline
FTRL & 34.09 & 23.46 & 0.68 & 31.41 & 21.14 & 0.77 & 29.06 & 10.72 & 0.42 \\
GPT-4o mini & 49.54 & 22.49 & 0.65 & 42.92 & 18.39 & 0.72 & 28.60 & \cellcolor{yellow!30}11.56 & \cellcolor{yellow!30}0.46 \\
Trained GPT-4o mini & \cellcolor{yellow!30}38.86 & \cellcolor{yellow!30}20.93 & \cellcolor{yellow!30}0.62 & \cellcolor{yellow!30}39.68 & \cellcolor{yellow!30}16.90 & \cellcolor{yellow!30}0.69 & \cellcolor{yellow!30}28.42 & 12.45 & 0.47 \\
\thickhline
\end{tabular}%
}
\caption{
\textbf{Summary of the regret value for the FOL environment under \texttt{Action Space Size Generalization}[$d=3 \rightarrow d=4$] on GPT-4o mini}, trained and evaluated on a mixture of the \protect\mybluehyperlink{gaussianmu}{Gaussian}, \protect\mybluehyperlink{uniform}{Uniform}, and \protect\mybluehyperlink{sine}{Sine-trend} rewards. 
Highlighted (yellow) cells indicate the lower value between the base model and the trained model, highlighting which model performs better in terms of cumulative regret at round~$T$ and growth rate exponent~$\hat{\beta}$ for each reward generation process. The trained model demonstrates comparable or improved performance, indicating robustness to increased action dimensionality. 
}
\label{tab:FOL-ood-action}
\end{table}
\clearpage

\subsection{Multi-Armed Bandits}

\begin{figure}[H]
   \centering
   \includegraphics[width=\linewidth]{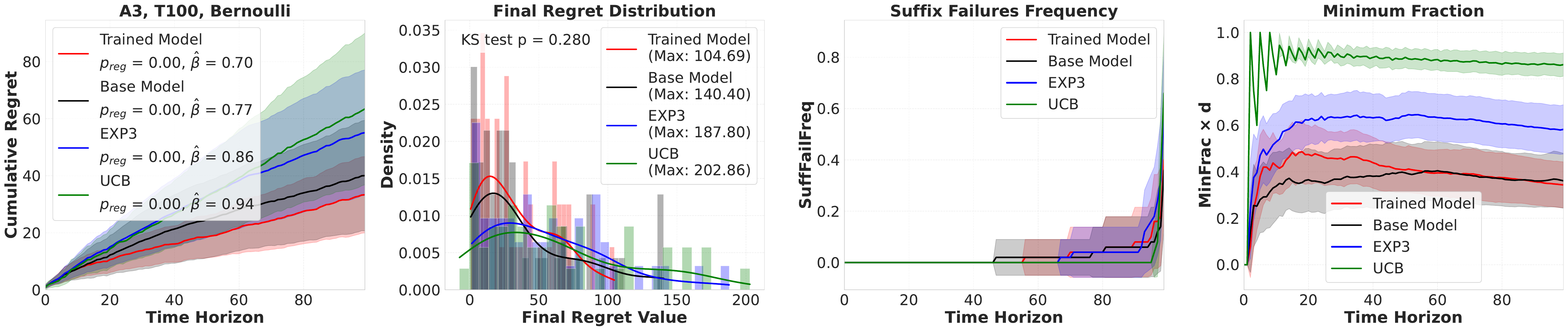}
   \caption{
   \textbf{The regret over time, the final regret distribution, and the exploration and exploitation metrics for the MAB environment under both \texttt{Horizon Generalization}[$T=25 \rightarrow T=100$] and \texttt{Reward Generalization}[\protect\mybluehyperlink{gaussianmu}{Gaussian} $\rightarrow$ \protect\mybluehyperlink{bernoulli}{Bernoulli}] on GPT-4o mini,} showing a lower regret and sublinear regret behavior for the trained model. The trained GPT-4o mini model sustains effective exploration and adapts its exploitation strategy over time.}
   \label{fig:Bandit-ood-bernoulli}
\end{figure}

\begin{figure}[H]
   \centering
   \includegraphics[width=\linewidth]{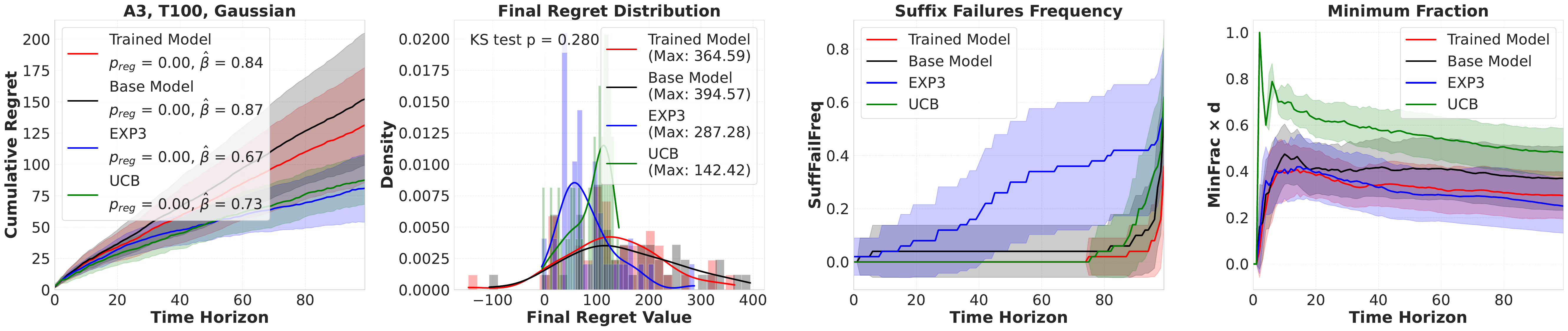}
   \caption{
   \textbf{The regret over time, the final regret distribution, and the exploration and exploitation metrics for the MAB environment under \texttt{Linguistic Context Generalization}[GPT-4o $\rightarrow$ Gemini-2.0-Flash] on GPT-4o mini using Gemini-2.0-Flash-generated contexts,} showing a lower regret and sublinear regret behavior for the trained model. The trained GPT-4o mini model sustains effective exploration and adapts its exploitation strategy over time.}
   \label{fig:Bandit-ood-gemini}
\end{figure}

\begin{figure}[H]
   \centering
   \includegraphics[width=\linewidth]{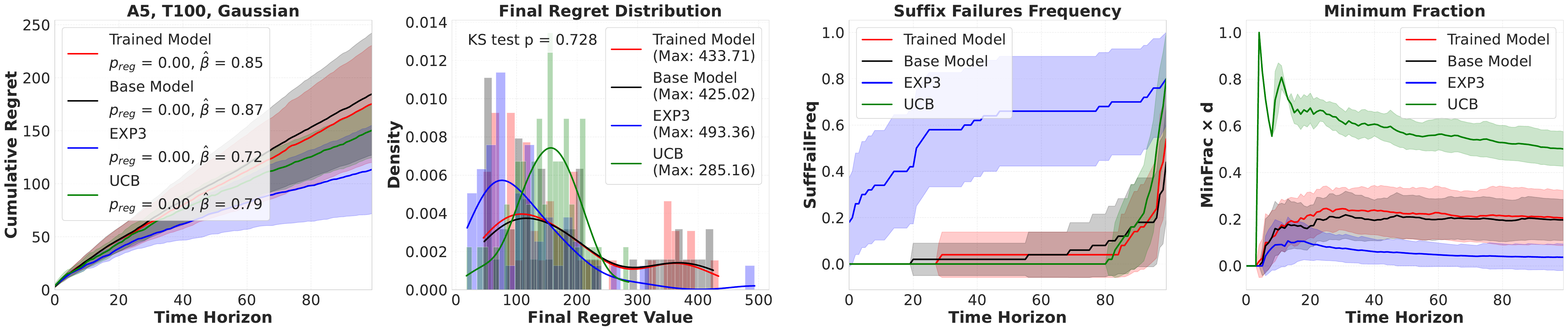}
   \caption{
   \textbf{The regret over time, the final regret distribution, and the exploration and exploitation metrics for the MAB environment under \texttt{Action Space Size Generalization}[$d=3 \rightarrow d=5$] on GPT-4o mini,} showing a slightly lower regret-growth exponent, but not lower maximum regret or final-regret distribution, for the trained model. The trained GPT-4o mini model sustains effective exploration and adapts its exploitation strategy over time.}
   \label{fig:Bandit-ood-action}
\end{figure}
\clearpage
\subsection{Non-Stationary Multi-Armed Bandits}
\begin{figure}[H]
   \centering
   \includegraphics[width=0.9\linewidth]{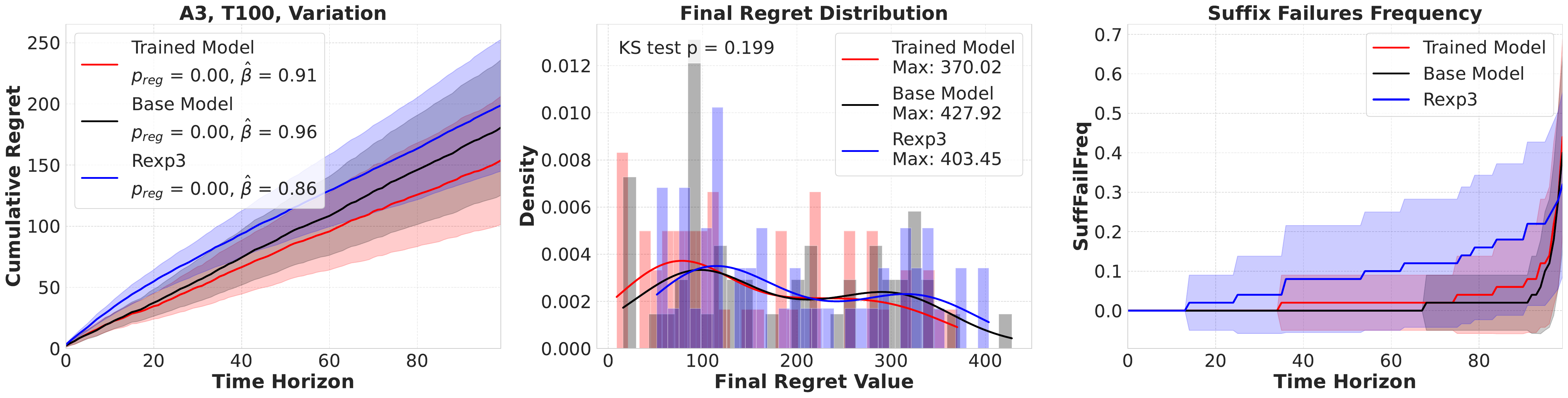}
   \caption{
   \textbf{The regret over time, the final regret distribution, and the exploration metric for the NS-MAB environment under \texttt{Horizon Generalization}[$T=25 \rightarrow T=100$]}, trained and evaluated with the \protect\mybluehyperlink{gradual}{Gradual Variation} reward, which shows a lower regret and sublinear regret behavior for the trained model.}
   \label{fig:nonstationary-ood-gradual}
\end{figure}
\begin{figure}[H]
   \centering
   \includegraphics[width=0.9\linewidth]{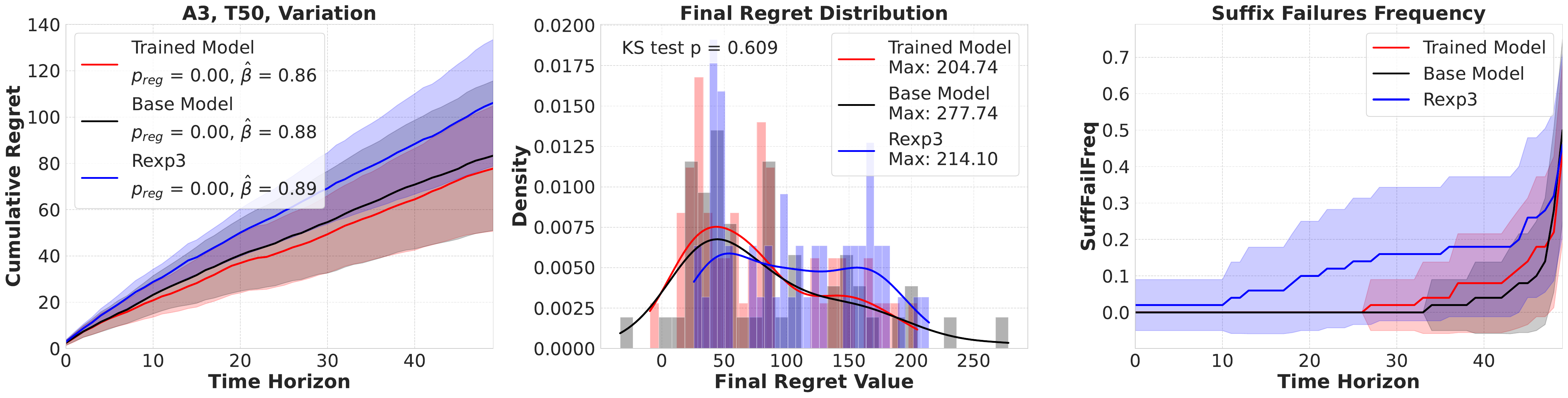}
   \caption{
   \textbf{The regret over time, the final regret distribution, and the exploration metric for the NS-MAB environment under \texttt{Horizon Generalization}[$T=25 \rightarrow T=50$] and \texttt{Linguistic Context Generalization}[GPT-4o mini $\to$ Gemini 2.0-Flash]}, trained and evaluated with the \protect\mybluehyperlink{gradual}{Gradual Variation} reward, which shows a lower regret and sublinear regret behavior for the trained model.}
   \label{fig:nonstationary-ood-gemini}
\end{figure}
\begin{figure}[H]
   \centering
   \includegraphics[width=0.9\linewidth]{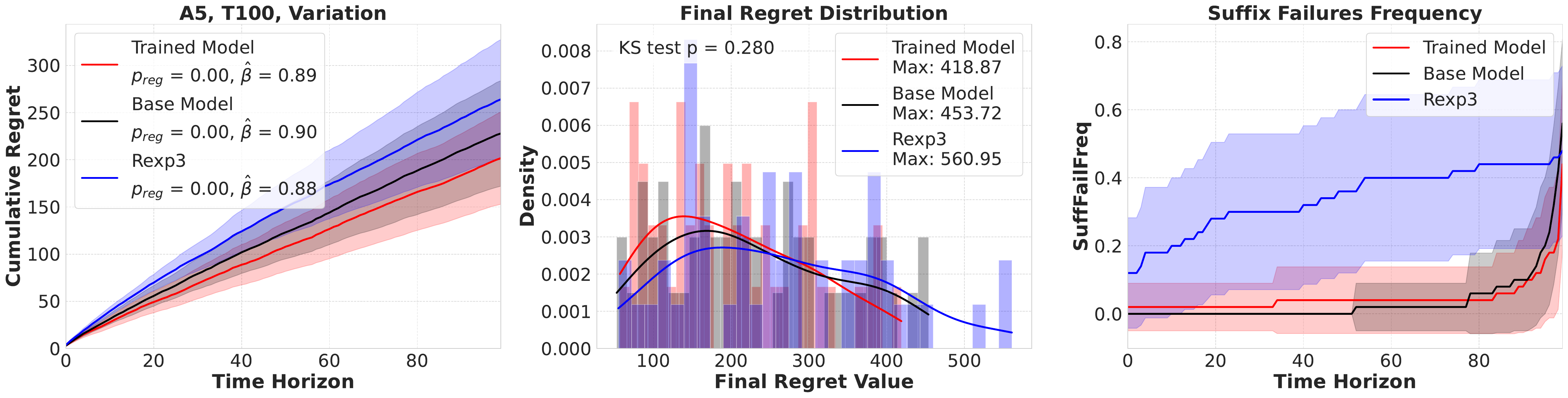}
   \caption{
   \textbf{The regret over time, the final regret distribution, and the exploration metric for the NS-MAB environment under \texttt{Horizon Generalization}[$T=25 \rightarrow T=100$] and \texttt{Action Space Size Generalization}[$d=3 \rightarrow d=5$]}, trained and evaluated with the \protect\mybluehyperlink{gradual}{Gradual Variation} reward, which shows a lower regret and sublinear regret behavior for the trained model.}
   \label{fig:nonstationary-ood-action}
\end{figure}
\clearpage

\section{Omitted Proof of \Cref{thm:single-layer-regret-minimizer}}
\label{appendix:thm1pf}
\linear*
\begin{proof}
   \begin{equation}
    \begin{aligned}
        g(&Z_t;V, K, Q, v_c, k_c, q_c) 
        \\
        &= \sum_{i=1}^t \left(VR_i R_i^\intercal (K^\intercal (Qc + q_c)) + \left(V k_c^ \intercal (Qc + q_c)+ v_c (Qc + q_c)^\intercal K\right)R_i + v_c k_c^\intercal(Qc + q_c)\right),
    \end{aligned} \label{eqn:original-param-proof}
\end{equation}

which can be expressed with a larger class 
\begin{align}
    g(Z_t, \mathbb{A}, \beta, \mathbb{C}, \delta):= \sum_{i=1}^t(\mathbb{A} R_i R_i^\intercal \beta + \mathbb{C} R_i + \delta), \label{eqn:reparam-ABCD}
\end{align}
where $\mathbb{A} \in \RR^{d \times d}$, $\beta, \mathbb{C}, \delta \in \RR^d$, and $Z_t = (R_1, \dots, R_t)$. Then, if a minimizer of 
\begin{align}
    f(\mathbb{A}, \beta, \mathbb{C}, \delta): &= \EE \left[\sum_{t=1}^{T} \left\| \sum_{i=1}^{t-1}(\mathbb{A} R_i R_i^\intercal \beta + \mathbb{C} R_i + \delta) - \pi^\star(R_1, ..., R_T) \right\|_2^2  \right] \nonumber
\end{align}
can be expressed as  $\mathbb{A} = V, \beta = K^\intercal (Qc + q_c), \mathbb{C} = V k_c^\intercal (Qc + q_c) + v_c(Qc+ q_c)^\intercal K, \beta = v_c k_c^\intercal (Qc + q_c)$, then we can conclude that the corresponding $V, Q, K, v_c, q_c, k_c$ are also a minimizer of  
\begin{align*}
    \EE \left[ \sum_{t = 1}^T \norm{ g(Z_{t-1}) - \pi^\star(R_1, ..., R_T) }^2_2)\right],
\end{align*}
since the corresponding $V, Q, K, v_c, q_c, k_c$ constitute a minimizer among a larger class. Now, since $\Pi = B(\pmb{0}_d, R_{\Pi}, \norm{\cdot})$, we can rewrite $f$ as 
\begin{align}
    f(\mathbb{A}, \beta, \mathbb{C}, \delta): &= \EE \left[\sum_{t=1}^{T} \left\| \sum_{i=1}^{t-1}(\mathbb{A} R_i R_i^\intercal \beta + \mathbb{C} R_i + \delta) - R_\Pi\left( \frac{\sum_{i=1}^T R_i}{\left\|\sum_{i=1}^T R_i\right\|_2} \right) \right\|_2^2  \right]\label{eqn:alternative-single-layer-transformer}
\end{align}

Here, Expectation is calculated with $R_i \sim \cN(\pmb{0}_d, I_{d \times d})$ and all $R_i$ are independent.

\safevspace{6pt}
\textbf{{\color{blue}Step 1}. Finding the best $\delta$}.
We have
\begin{align*}
&\EE \left[\sum_{t=1}^{T} \left\| \sum_{i=1}^{t-1}(\mathbb{A} R_i R_i^\intercal \beta + \mathbb{C} R_i + \delta) - R_\Pi\left( \frac{\sum_{i=1}^T R_i}{\left\|\sum_{i=1}^T R_i\right\|_2} \right) \right\|_2^2\right]
\\ 
&= \EE \left[\sum_{t=1}^{T} \left\| \sum_{i=1}^{t-1}(\mathbb{A} R_i R_i^\intercal \beta + \mathbb{C} R_i - A \beta) - R_\Pi\left( \frac{\sum_{i=1}^T R_i}{\left\|\sum_{i=1}^T R_i\right\|_2} \right) \right\|_2^2\right] + \frac{T(T-1)}{2} \norm{A  \beta + \delta}_2^2
\end{align*}
since $\EE[\norm{X}^2_2] = \EE[\norm{X - \EE[X]}^2_2] + \norm{\EE[X]}^2_2 $, $\delta = A  \beta$ holds for any random vector $X$. Therefore, $\delta = -A  \beta$ is the condition to minimize $f$. 

\textbf{{\color{blue}Step 2}. Plugging the optimality condition for $\delta$ into $f$ }.

Plugging $\delta = -A \beta$ to $f$ provides  
\begin{align*}
    f(\mathbb{A}, &\beta, \mathbb{C}, -\mathbb{A}  \beta) =\EE \left[\sum_{t=1}^{T} \left\| \sum_{i=1}^{t-1}(\mathbb{A} (R_i R_i^\intercal - I_{d \times d}) \beta + \mathbb{C} R_i) - R_\Pi\left( \frac{\sum_{i=1}^T R_i}{\left\|\sum_{i=1}^T R_i\right\|_2} \right) \right\|_2^2\right] 
    \\
    &= \sum_{t=1}^{T} \Biggl( \underbrace{\EE\left[ \norm{\sum_{i=1}^{t-1} \mathbb{A} (R_iR_i^\intercal- I_{d \times d}) \beta}_2^2 \right]}_{(i)} + \EE\left[ \norm{\sum_{i=1}^{t-1}\CC R_i}_2^2 \right] + \EE\left[\norm{R_\Pi\frac{\sum_{i=1}^T R_i}{\left\|\sum_{i=1}^T R_i\right\|_2}}_2^2\right] \Biggr) 
    \\
    &\quad + \underbrace{2\sum_{t=1}^T \EE\left[\left( \sum_{i=1}^{t-1} \mathbb{A} (R_iR_i^\intercal- I_{d \times d}) \beta \right)^\intercal  \left(\sum_{i=1}^{t-1} \CC R_i\right) \right]}_{(ii)}  
    \\
    &\qquad - \underbrace{2\sum_{t=1}^T \EE\left[\left( \sum_{i=1}^{t-1} \mathbb{A} (R_iR_i^\intercal- I_{d \times d}) \beta \right)^\intercal  \left(R_\Pi\frac{\sum_{i=1}^T R_i}{\left\|\sum_{i=1}^T R_i\right\|_2}\right) \right]}_{(iii)}  
    \\
    &\qquad \quad - {2\sum_{t=1}^T \EE\left[\left(\sum_{i=1}^{t-1} \CC R_i\right)^\intercal  \left(R_\Pi\frac{\sum_{i=1}^T R_i}{\left\|\sum_{i=1}^T R_i\right\|_2}\right) \right]}
\end{align*}
We can easily check that $(ii)$ and $(iii)$ are $0$ as they are polynomials of odd degrees and we have $Z \overset{d}{=}-Z$. For the part $(i)$, we have 
\begin{align}
    \sum_{t=1}^T \EE\left[ \norm{\sum_{i=1}^{t-1} \mathbb{A} (R_iR_i^\intercal- I_{d \times d}) \beta}_2^2 \right] &= \sum_{t=1}^T \EE \left[\sum_{i_1 = 1}^{t- 1} \sum_{i=1}^{t-1} \beta^\intercal (R_{i_1} R_{i_1}^\intercal- I_{d \times d})\mathbb{A}^\intercal \mathbb{A} (R_iR_i^\intercal- I_{d \times d}) \beta \right] \nonumber 
    \\
    &\underset{(1)}{=}\sum_{t=1}^T  \EE \left[ \sum_{i=1}^{t-1} \beta^\intercal (R_{i} R_{i}^\intercal- I_{d \times d})\mathbb{A}^\intercal \mathbb{A} (R_iR_i^\intercal- I_{d \times d}) \beta \right] \label{eqn:calculation-under-local-opt}
    \\
    &= \frac{(T-1)T}{2} \beta^\intercal \EE \left[ (A (R_iR_i^\intercal- I_{d \times d}))^\intercal (A (R_iR_i^\intercal- I_{d \times d})) \right] \beta. \nonumber
\end{align}
Here, $(1)$ holds because if $i_1 \neq i$, we can calculate $\EE (R_{i_1} R_{i_1}^\intercal- I_{d \times d}) = \bO_{d \times d}$. Note that  \Cref{eqn:calculation-under-local-opt} is minimized when $\PP( \mathbb{A} (R_iR_i^\intercal- I_{d \times d})\beta = \pmb{0}_d) = 1$. 

If $\mathbb{A} \neq \bO_{d \times d}$, suppose that the singular value decomposition of $A = U \Lambda V$ yields that $\Lambda$ is a diagonal matrix whose first diagonal element is non-zero, and $U, V$ are orthogonal matrices. Then, we want to find $\beta$ that $ U \Lambda V (R_i R_i^\intercal-I_{d \times d}) \beta = \pmb{0}_d$ for any $R_i$ such that $p(R_i) \neq 0$, where $p$ corresponds to the probability density function of reward vectors. Since $U$ is invertible, we only need to consider $\Lambda V (R_i R_i^\intercal - I_{d \times d}) \beta = \pmb{0}_d$. Since $\Lambda$'s first diagonal component is non-zero, we will consider equation $e_1^\intercal \Lambda V (R_i R_i^\intercal- I_{d \times d}) \beta = 0$. This is equivalent to $V_1 (R_i R_i^\intercal - I_{d \times d}) \beta = 0$,  where $V_1$ is the first row of $V$, and is a non-zero vector.

\begin{restatable}{claim}{positivevolume}
\label{claim:positivevolume}
Define $\cL = B(\pmb{0}_d,1, \norm{\cdot}_2$) and $\cS:=\left\{V_1\left(R_i R_i^{\intercal}-I_{d \times d}\right): R_i \in \mathcal{L}\right\} \subseteq  \RR^{d}.$ Then, $\cS$ has positive volume. As a consequence, this set is full $d$-dimensional and contains $d$ linearly independent vectors.     
\end{restatable}
Therefore, {we can find $d$ reward vectors $\{R_i\}_{i\in[d]}$ such that the vectors $\{V_1 (R_i R_i^\intercal- I_{d \times d})\}_{i\in[d]}$ are linearly independent}.
Hence, if we want to minimize  \Cref{eqn:calculation-under-local-opt}, either $A=\bO_{d \times d}$ or $\beta =\pmb{0}_d$ should hold. In both cases, \Cref{eqn:reparam-ABCD} can be re-written as 
\begin{align*}
    g(Z_t; \mathbb{A}, \beta, \mathbb{C}, \delta):= \sum_{i=1}^t \mathbb{C} R_i,
\end{align*}
and this is covered by the original parametrization (\Cref{eqn:original-param}) with $K^\intercal (Qc + q_c) = v_c = \pmb{0}_d$. 

\safevspace{6pt}
\textbf{{\color{blue}Step 3}. Finding the best $\CC$}.

Now, we optimize over $\mathbb{C}$, by minimizing the following objective: 
\begin{align*}
    f(\mathbb{C}):&= \EE \left[\sum_{t=1}^{T} \left\| \sum_{i=1}^{t-1}\mathbb{C} R_i - R_\Pi\left( \frac{\sum_{i=1}^T R_i}{\left\|\sum_{i=1}^T R_i\right\|_2} \right) \right\|_2^2  \right]
    \\
    &= \sum_{t=1}^T \EE \left[\norm{\sum_{i=1}^{t-1}\mathbb{C} R_i}_2^2\right] - 2R_\Pi \sum_{t=1}^T \EE \left[\left(\sum_{i=1}^{t-1} \CC R_i \right)^\intercal \left( \frac{\sum_{i=1}^T R_i}{\left\|\sum_{i=1}^T R_i\right\|_2} \right)\right] 
    \\
    &\quad +R_\Pi^2\sum_{t=1}^T \EE\left[\frac{(\sum_{i=1}^T R_i)^\intercal(\sum_{i=1}^T R_i) }{\left\|\sum_{i=1}^T R_i\right\|_2^2}\right] 
    \\
    &=   \frac{T(T-1)}{2} d \CC^\intercal \CC  - 2R_\Pi \CC^\intercal  \sum_{t=1}^T \EE \left[   \frac{\left(\sum_{i=1}^{t-1} R_i \right)^\intercal \left(\sum_{i=1}^T R_i\right)}{\left\|\sum_{i=1}^T R_i\right\|_2} \right] 
    \\
    &\quad +\sum_{t=1}^T \EE\left[\frac{(\sum_{i=1}^T R_i)^\intercal(\sum_{i=1}^T R_i) }{\left\|\sum_{i=1}^T R_i\right\|_2^2}\right], 
\end{align*}
since $\EE \left[\norm{\sum_{i=1}^{t-1}R_i}_2^2 \right]=  (t-1) d$.  

Therefore, the optimal $\CC$ can be calculated as $$\CC=\frac{2R_\Pi}{T(T-1)d} \sum_{t=1}^T \EE\left[ \frac{\bigl(\sum_{i=1}^{t-1}R_i\bigr)\bigl(\sum_{i=1}^T R_i\bigr)^{\intercal}} {\bigl\|\sum_{i=1}^T R_i\bigl\|_2}
\right],$$ or we can rewrite it as 

\begin{align*}
    \CC   =\frac{2R_\Pi}{T(T-1)d}
     \sum_{t=1}^{T}  \EE\left[      \frac{S_{t-1} S_T^{\intercal}}             {\|S_T\|_2}\right],
\end{align*}
 where $S_T := \sum_{i=1}^{T}R_i, 
S_{t-1} := \sum_{i=1}^{t-1}R_i$. 
Because \((S_{t-1},S_T)\) is jointly Gaussian, $\mathbb E[S_{t-1}\mid S_T]
   =\frac{t-1}{T} S_T$, so 
   \begin{align*}
\mathbb E\Bigl[\tfrac{S_{t-1}S_T^{\intercal}}{\|S_T\|_2}\Bigr]
   =\frac{t-1}{T}
     \mathbb E\Bigl[\tfrac{S_TS_T^{\intercal}}{\|S_T\|_2}\Bigr]
   \end{align*}
   and 
   \begin{align*}
    \CC   =\frac{2R_\Pi}{T(T-1)d} \frac{T-1}{2} \EE\left[      \frac{S_{T} S_T^{\intercal}}             {\|S_T\|_2}\right] = -\frac{R_\Pi}{Td}  \EE\left[      \frac{S_{T} S_T^{\intercal}}             {\|S_T\|_2}\right].
\end{align*}
\begin{restatable}{claim}{isotropic}
\label{claim:isotropic}
If $QX \overset{d}{=}X$ for all $Q \in \mathbb{O} (d)$, then 
\[
\mathbb E\Bigl[\tfrac{XX^{\intercal}}{\|X\|_2}\Bigr]
      =\frac{\mathbb E[\|X\|_2]}{d} I_{d \times d}.
\]
\end{restatable}
Therefore, we have 
\begin{align*}
    \CC =  \frac{R_\Pi}{Td}  \frac{\EE \norm{S_{T}}_2}{d} I_{d \times d}   \underset{(i)}{=}  \frac{R_\Pi}{Td}  \sqrt{2T} \frac{\Gamma\left(\tfrac{d+1}{2}\right)}{\Gamma\left(\tfrac{d}{2}\right)}I_{d} =  \frac{\sqrt{2}R_\Pi}{\sqrt{T}d}   \frac{\Gamma\left(\tfrac{d+1}{2}\right)}{\Gamma\left(\tfrac{d}{2}\right)}I_{d}.         
\end{align*}
Here, (i) holds since \( S_t\sim\mathcal N(\pmb{0}_d,tI)\). Actually, this can be calculated even in closed-form as 
\begin{align*}
    \CC = \frac{\sqrt{2} R_\Pi}{\sqrt{T}} \frac{(d-1)!!}{2^{d/2}} \frac{\sqrt{\pi}}{(\frac{d}{2}-1)!d} I_{d}, 
\end{align*}
if $d$ is even, and as 
\begin{align*}
    \CC = \frac{\sqrt{2}R_\Pi}{\sqrt{T}} \frac{\left(\frac{d-1}{2}-1\right)! 2^{\frac{d-1}{2}}}{(d-2)!!} \frac{1}{\sqrt{\pi}} I_{d}, 
\end{align*}
{if $d$ is odd.} 
If  $d$ goes to infinity, {we have}
\begin{align}
    \lim_{d \to \infty} \sqrt{d} \CC =  \lim_{d \to \infty}   \frac{\sqrt{2}R_\Pi}{\sqrt{T}d} \frac{d}{\sqrt{2}}I_{d} =   \frac{R_\Pi}{\sqrt{T}}  I_{d \times d}.
\end{align}
\end{proof}

\positivevolume*
\begin{proof}[Proof of Claim \ref{claim:positivevolume}]
Let \(v:=V_1^{\intercal}\in \mathbb{R}^d\), and define
\[
s_v(R):=\left(RR^{\intercal}-I_d\right)v ,
\qquad R\in \mathcal L .
\]
Since \(RR^{\intercal}-I_d\) is symmetric, the image of this map is the transpose
of the set
\[
\left\{V_1\left(RR^{\intercal}-I_d\right):R\in \mathcal L\right\}.
\]
Thus it suffices to show that \(s_v(\mathcal L)\subset \mathbb{R}^d\) has positive
\(d\)-dimensional volume.

The derivative of \(s_v\) with respect to \(R\) is
\[
D_R s_v(R)
=
(R^{\intercal}v)I_d+Rv^{\intercal}.
\]
Indeed, for any perturbation \(h\in \mathbb{R}^d\),
\[
D_R s_v(R)[h]
=
h(R^{\intercal}v)+R(h^{\intercal}v).
\]
Therefore the Jacobian matrix is
\[
J_v(R):=D_R s_v(R)=(R^{\intercal}v)I_d+Rv^{\intercal}.
\]

Using the matrix determinant lemma, whenever \(R^{\intercal}v\neq 0\),
\[
\det J_v(R)
=
\det\left((R^{\intercal}v)I_d+Rv^{\intercal}\right)
=
2(R^{\intercal}v)^d .
\]
Because \(v\neq 0\), we can choose \(R_0\in \mathcal L\) such that
\(R_0^{\intercal}v\neq 0\); for example,
\[
R_0=\frac{1}{2}\frac{v}{\|v\|_2}.
\]
Then
\[
\det J_v(R_0)=2(R_0^{\intercal}v)^d\neq 0.
\]
By continuity, there exists an open ball \(\mathcal B\subset \mathcal L\) around
\(R_0\) such that
\[
\det J_v(R)\neq 0
\qquad \text{for all } R\in \mathcal B .
\]
Hence \(s_v\) is locally nondegenerate on \(\mathcal B\). By the inverse function
theorem, \(s_v(\mathcal B)\) contains an open subset of \(\mathbb{R}^d\). Therefore
\(s_v(\mathcal L)\) has positive \(d\)-dimensional volume.

Since transposition preserves \(d\)-dimensional volume, the set
\[
\left\{V_1\left(RR^{\intercal}-I_d\right):R\in \mathcal L\right\}
\]
also has positive volume. Consequently, this set is full-dimensional and contains
\(d\) linearly independent vectors.
\end{proof}
\isotropic*
\begin{proof}[Proof of \Cref{claim:isotropic}]
By the rotational invariance of $X$, we have that for any $Q \in \mathbb{O}(d)$,
\[
\mathbb{E}\left[\frac{XX^\intercal}{\|X\|}\right] 
= \mathbb{E}\left[\frac{QX (QX)^\intercal}{\|QX\|}\right] 
= Q \mathbb{E}\left[\frac{XX^\intercal}{\|X\|}\right] Q^\intercal.
\]
This shows that $\mathbb{E}\left[\frac{XX^\intercal}{\|X\|}\right]$ is invariant under conjugation by orthogonal matrices, and hence must be a scalar multiple of the identity matrix:
\[
\mathbb{E}\left[\frac{XX^\intercal}{\|X\|}\right] = \lambda I_{d\times d},
\]
for some $\lambda \in \mathbb{R}$. To determine $\lambda$, we take the trace of both sides:
\[
\text{Tr}\left(\mathbb{E}\left[\frac{XX^\intercal}{\|X\|}\right]\right) = \mathbb{E}\left[\frac{\text{Tr}(XX^\intercal)}{\|X\|}\right] = \mathbb{E}\left[\frac{\|X\|^2}{\|X\|}\right] = \mathbb{E}[\|X\|],
\]
and
\[
\text{Tr}(\lambda I_d) = \lambda d.
\]
Equating both traces yields $\lambda = \frac{\mathbb{E}[\|X\|]}{d}$. Therefore, we have 
\[
\mathbb{E}\left[\frac{XX^\intercal}{\|X\|}\right] = \frac{\mathbb{E}[\|X\|]}{d} \cdot I_{d \times d},
\]
which completes the proof. 
\end{proof}

\end{document}